\documentclass[acmlarge]{acmart}
\usepackage{rotating}
\usepackage{graphicx}
\usepackage{caption}
\usepackage{multirow,array}
\usepackage{float}
\usepackage{amsmath}
\usepackage{ragged2e}
\usepackage{longtable}

\AtBeginDocument{%
  }

\setcopyright{acmcopyright}
\copyrightyear{2023}
\acmYear{2023}
\acmDOI{XXXXXXX.XXXXXXX}

\acmJournal{JACM}
\acmVolume{37}
\acmNumber{4}
\acmArticle{111}
\acmMonth{8}

\begin{document}

%\title{A comprehensive review on Explainable AI (XAI) in Financial domain}
\title{A Comprehensive Review on Financial Explainable AI}
%\title{A comprehensive review on Explainable AI (XAI) in Finance}

\author{Yeo Wei Jie}
\email{yeow0082@e.ntu.edu.sg}
\orcid{0000-0001-5254-8359}
\affiliation{%
  \institution{Nanyang Technological University (NTU)}
  \streetaddress{50 Nanyang Ave}
  \country{Singapore}
  \postcode{639798}
}
\author{van der Heever Wihan}
\affiliation{%
  \institution{Nanyang Technological University (NTU)}
  \streetaddress{50 Nanyang Ave}
  \country{Singapore}
  \postcode{639798}
}
\email{wihan001@e.ntu.edu.sg}

\author{Mao Rui}
\orcid{0000-0002-1082-8755}
\affiliation{%
  \institution{Nanyang Technological University (NTU)}
  \streetaddress{50 Nanyang Ave}
  \country{Singapore}
  \postcode{639798}
}
\email{rui.mao@ntu.edu.sg}

\author{Cambria Erik}
\orcid{0000-0002-3030-1280}
\affiliation{%
  \institution{Nanyang Technological University (NTU)}
  \streetaddress{50 Nanyang Ave}
  \country{Singapore}
  \postcode{639798}
}
\email{cambria@ntu.edu.sg}

\author{Satapathy Ranjan}
\orcid{0000-0002-0733-7381}
\affiliation{%
  \institution{(corresponding author) Institute of High Performance Computing (IHPC), Agency for Science, Technology and Research (A\textasteriskcentered STAR)}
  \streetaddress{Fusionopolis Way, \#16-16 Connexis, Singapore}
  \country{Republic of Singapore}
  \postcode{138632}
}
\email{satapathy_ranjan@ihpc.a-star.edu.sg}

\author{Mengaldo Gianmarco}
\orcid{0000-0002-0157-5477}
\affiliation{%
  \institution{National University of Singapore (NUS); Asian Institute of Digital Finance at NUS,  Singapore; Honorary Research Fellow, Imperial College London, United Kingdom}
  \streetaddress{9 Engineering Drive 1}
  \country{Singapore}
  \postcode{117575}
}
\email{mpegim@nus.edu.sg}

\renewcommand{\shortauthors}{Wei Jie et. al.}

\begin{abstract}
The success of artificial intelligence (AI), and deep learning models in particular, has led to their widespread adoption across various industries due to their ability to process huge amounts of data and learn complex patterns. However, due to their lack of explainability, there are significant concerns regarding their use in critical sectors, such as finance and healthcare, where decision-making transparency is of paramount importance.  In this paper, we  provide a comparative survey of methods that aim to improve the explainability of deep learning models within the context of finance. We categorize the collection of explainable AI methods according to their corresponding characteristics, and we review the concerns and challenges of adopting explainable AI methods, together with future directions we deemed appropriate and important.
\end{abstract}




%\received{20 February 2023}
%\received[revised]{12 March 2023}
%\received[accepted]{5 June 2023}

\maketitle

\section{Introduction}
\label{sec:introduction}

Finance is a constantly evolving sector that is deeply rooted in the development of human civilization. One of the main tasks of finance is the efficient allocation of resources, with a chief example being the handling of capital flows between various entities with different needs.
These entities can be divided into individuals, companies, and countries, and lead to the common categorization of personal, corporate, and government finance. The sector can be traced back to 5000 years ago, in the agrarian societies that had been established and developed for some thousand of years at the time. Indeed, one of the first examples of banking, a central institution within finance, can be attributed to the Babylonian empire. Since then, societal development and technological advances have pushed the field to undergo several changes. In the past two decades, these changes have been particularly marked, due to the accelerating pace of technological development, especially in the context of AI. The latter has started spreading across multiple segments of finance, from digital transactions to investment management, risk management, algorithmic trading, and more~\cite{FinOverview}. The use of novel AI- and non-AI technologies to automate and improve financial processes is now known as FinTech (Financial Technology), and its growth in the past two decades has been remarkable~\cite{Fintech}. In this review, we focus on AI-based technologies and machine learning for financial applications.s

Financial researchers and practitioners have been relying on supervised, unsupervised, and semi-supervised machine learning methods as well as reinforcement learning for tackling many different problems. Some examples include credit evaluation, fraud detection, algorithmic trading, and wealth management. In supervised-based machine learning methods it is common to use e.g., neural networks to identify complex relationships hidden in the available labeled data. The labels are usually provided by domain experts. For instance, one can think of building a stock-picking system, where a domain expert labels periods of positive and negative returns. The machine is then tasked to build the relationship between (possibly) high-dimensional data, and positive and negative returns of a given stock (or multiple stocks) and generalize to unseen data to e.g., predict the future stock's behavior. In unsupervised-based machine learning methods, the task is instead to identify data with similar characteristics that can therefore be clustered together, without domain-expert labeling. For example, one can think of identifying all stocks that have similar characteristics into clusters using some similarity metrics, such as valuation, profitability and risk. Semi-supervised learning is a middle ground between supervised and unsupervised learning, where only a portion of the data is labeled. Finally, reinforcement learning aims to maximize, through a set of actions, the cumulative reward specified by the practitioners. Reinforcement learning is used in finance for e.g., portfolio construction. Reinforcement learning is strictly related to Markov decision processes and substantially differs from both supervised and unsupervised learning. 

Among supervised, unsupervised, and reinforcement learning methods, there is vast heterogeneity in terms of complexity. Some methods are considered easier to understand, hence to interpret by practitioners (also referred to as white-box methods), while others are considered not interpretable (also referred to as black-box methods). To this end, neural networks and deep learning strategies, that underpin the majority (albeit not the entirety) of recent machine learning methods for financial applications, are considered black-box methods - i.e., the reason for a given prediction is not of easy access when available). This constitutes a critical issue, especially in risky and highly regulated sectors, such as healthcare and finance, where a wrong decision may lead to catastrophic loss of life (healthcare) or capital (finance). Hence, it was deemed important to understand the reasons (i.e., the data and patterns) the machine used to make a given decision. This aspect encompasses the broad field of AI transparency. The latter is composed of three pillars, (i) AI awareness, (ii) AI model explainability, and (iii) AI outcome explainability. The first is tasked to understand whether AI is involved in a given product. The second is responsible to provide a detailed explanation of the AI model, including its inputs and outputs. The third is responsible to provide a granular explanation of the inputs' contributions to the AI model's outcomes. To this last category, we find a vast array of post-hoc interpretability methods. In this review, we assume that AI awareness is achieved, i.e., we know that in a given financial process AI is involved, and focus on AI explainability, also referred to as eXplainable AI or simply XAI. A further distinction commonly made is between interpretability and explainability of an AI model. These two terms, frequently used interchangeably, have subtle differences. Interpretability refers to how and why a model works. Explainability refers to the ability of explaining the results in human terms. 

While deep learning methods are considered black-boxes, many other methods in finance are considered white-box methods. The trade-off between complexity and interpretability is perhaps one of the most debated aspects in the field of financial AI. On the one hand, white-box methods are highly interpretable but lack the ability to grasp complex relationships, frequently failing to meet the desired performance. On the other hand, black-box methods are not interpretable but usually (although not always) meet the desired performance. Therefore, it is not surprising that there are significant efforts being pushed forward in recent years to render black-box methods more interpretable, where the primary example is the field of deep learning. In this paper, we provide an extensive review of XAI methods for the financial field that we name FinXAI. Although there have been a number of surveys on XAI methods~\cite{cambria2023survey,sahakyan2021explainable,arrieta2020explainable,guidotti2018survey,molnar2020interpretable,mueller2019explanation,rojat2021explainable}, these papers are targeted towards general XAI, and are not specific to finance. Hence, we conduct a review on explainability techniques exclusively related to financial use cases.

To compile this review, we took into account 69 papers, focusing mainly, though not exclusively on the third pillar, i.e., the explainability of the inputs' contributions to the AI model's outcomes. To this end, we considered both post-hoc interpretability methods applied to black-box deep learning models, and  inherently transparent models that do not require further post-hoc interpretability. Despite the relatively small number of collected papers in the field of XAI, it is important to note that our main objective is to focus specifically on XAI techniques applicable to the financial industry. This targeted approach will provide valuable insights for researchers in related fields and will ultimately help drive innovation and progress in the financial industry. With the growing need for transparency and accountability of deep learning, the XAI community has seen increasing growth in the number of works published, we focus here instead only on works concerning financial use cases. Notably, FinXAI is but a small subset of the general field of XAI and, thus we take a holistic approach to assembling existing studies with the goal of keeping up to date with the current approaches. The papers were queried from both Google Scholar and Scopus where we searched using a set of keywords relating to works that have applied explainable AI techniques in financial use cases, the set of keywords include \textit{``XAI, explainable AI, finance, financial sector, financial field, explainable ML''}. We try to collect a diverse set of papers that covers each category sufficiently well, and summarized in tables~\ref{tab: XAI_CE},~\ref{tab: XAI_FP},~\ref{tab: XAI_FM}. We also noticed the majority of explanation types were limited to fact-based explanations, hence we explicitly search for techniques explaining in the form of counterfactuals. Counterfactual explanations are deemed as a desirable form of explanation as the receiver tends to prefer understanding why a certain prediction was made instead of the opposing.

The main \textit{contributions} of our work as such:
\begin{itemize}
    \item We provide an extensive study on consolidating XAI methods in the field of finance (FinXAI), for researchers interested in prioritizing transparency in their solutions.
    \item We frame the FinXAI process as a sequential flow of decision-making processes (see Fig~\ref{fig:framework}), where we place importance on aligning the XAI technique with the target audience. The objective of this framework is to produce explanations that are both goal-oriented and audience-centric.
    \item We review current FinXAI techniques, analyze their technical contributions to ethical goals, and list down a number of key challenges faced in implementing XAI as well as important directions to be improved for the future.
\end{itemize}

The remainder of the review paper is organized as follows: Section~\ref{sec:definition} describes the definitions, reasons, and brief overview of FinXAI. Subsequently, we explain the methodology of FinXAI, starting from numerical in Section~\ref{section: numerical}, textual in Section~\ref{section: textual}, hybrid analysis in Section~\ref{section: hybrid} and ending with transparent models in Section~\ref{section: transparent}. In Section~\ref{section: FinXAI and ethical goals}, we analyze how the reviewed FinXAI methods contribute to ethical goals. Section~\ref{section:challenges} discusses key challenges of adopting explainable models and future directions for research. Finally, Section~\ref{section:conclusion} offers concluding remarks.
%
\iffalse
Inferring from~\ref{fig:paper selection}, there is a clear trend of increasing papers being published from 2017 to 2022, with 2022 having substantially more papers than from the previous year. 
\begin{figure}[ht!]
  \centering
  \includegraphics[width=\linewidth]{images/img9.png}
  \caption{Selection number of reviewed papers from 2017 to 2022. An increasing trend of published works in financial XAI in recent years.}
  \label{fig:paper selection}
\end{figure}
\fi
%

%%%%%%%%%%%%%%%%%%%%%%%%%%%%%%%%%%%%%%%%%%%%%%%%
% SECTION 2
%%%%%%%%%%%%%%%%%%%%%%%%%%%%%%%%%%%%%%%%%%%%%%%%
\section{FinXAI: Definition, Reason, Approach}
\label{sec:definition}
%%%%%%%%%%%%%%%%%%%%%%%%%%%%%%%%%%%%%%%%%%%%%%%%

This section details the definition, purpose, and approaches that have been taken in improving the transparency of AI models. A collection of existing literature reviews are analyzed and collated to give the reader a better understanding of commonly used terminologies in the XAI field, as well as the interlink between AI, linguistics, and social sciences which is essential to provide a solid understanding of the subject.

\subsection{Definition of FinXAI}
FinXAI or explainable AI methods in finance is inherently linked to the broader concept of AI transparency. As mentioned in section~\ref{sec:introduction}, this term encompasses under a single framework  three key steps: AI awareness, AI model explainability, and AI outcome explainability. In this review, we focus on the latter two aspects, model and outcome explainability. Model explainability means that the inner workings of a given AI solution are interpretable, and therefore the results may be interpreted by humans. This is typically the case for models with reduced complexity (i.e., white-box models), such as linear and logistic regression, and decision trees. 

Outcome explainability means that the inner workings of a given AI solution are not fully interpretable, and therefore the results may not be fully understandable by humans, unless some interpretability tools are applied to explain the AI outcomes. This is the case for complex models (i.e., black-box models), such as deep neural networks. In these cases, it is common to apply model agnostic post-hoc (and other) interpretability tools to understand the results the AI provided in human terms.   

% ``Explainability'' and ``interpretability'' are two common terms in the context of XAI, and they are often used interchangeably. However, the two terms bear subtle differences from each other. \cite{arrieta2020explainable} considers \textit{``explainability''}, a term describing the actions undertaken by the model to clarify the outcome or behavior. In comparison, \textit{"interpretability"} is a passive property inherent to the underlying nature of the model to make itself understandable to the end-users. A third term, \textit{``transparency''} is also used interchangeably with the former two terms. In our work, we assume it to be identical to interpretability. 

Correspondingly, XAI models may be cast into two broad categories:  intrinsically explainable due to their highly interpretable nature (e.g., linear and logistic regression), and extrinsically explainable, hence requiring an external tool to make them interpretable. In turn, these two categories of models lead to different classes of model transparency: \textit{simulatability}, \textit{decomposability}, and \textit{algorithmic transparency}~\cite{arrieta2020explainable}. Each of these three classes inherits the preceding class' properties, that is, if a model is decomposable, it is also simulatable, and if a model is algorithmically transparent is also decomposable and simulatable. In simple terms, \textit{simulatability} refers to the model's ability to allow a human observer to simulate a thought process over the inner workings of the model. \textit{Decomposability} entails that interpretability is available at every segment of the model, including inputs, outputs as well as model inner workings and parameters.  \textit{Algorithmic transparency} largely deals with the human user being able to understand how the model reacts with varying inputs and more importantly the ability to reason about errors the model produces.

A transparent model is an interpretable model which exhibits the ability to provide human-understandable explanations. Areas of transparency not only lie within the region of the model but also the data and design process of the end-product~\cite{van2020xai}. The EU High-Level Expert Group on AI~\cite{euhighlevel} states that the data the model interacted with should be traceable by human users at any given time. In addition, the design process of the system must be clear and explainable in a manner comprehensible to related stakeholders. The list of information types, regarded as explainable can even be extended to include principles and guidelines in the development of the AI system, as well as personnel involved in the implementation and development process~\cite{kuiper2022exploring}. 

A key rationale for explainability is to gain the trust of affected stakeholders. Examples of such stakeholders include regulators, board members, auditors, end-users and developers~\cite{davos}. To this end, the format and degree of explanation vary among audiences. The key message is usually conveyed in reports customized to the suitability of the receiving audience. It is common knowledge that financial service providers are regularly audited by supervisory authorities to ensure adherence to regulations and to prevent potential fraud from taking place. The level of scrutiny expected of the authorities is much higher than what the service providers expect. A study conducted by~\cite{kuiper2022exploring} involves a preliminary investigation to identify the types of information that are deemed necessary, in the perspective of banks and supervisory authorities. The result was that supervisory authorities identify all forms of information types as relevant while banks only consider a subset of them. As such, there exists a gap between each organization's understanding of necessity, more often than not leading to the delay in approving the deployment of financial services. 

\begin{figure}[ht!]
  \centering
  \includegraphics[width=\linewidth]{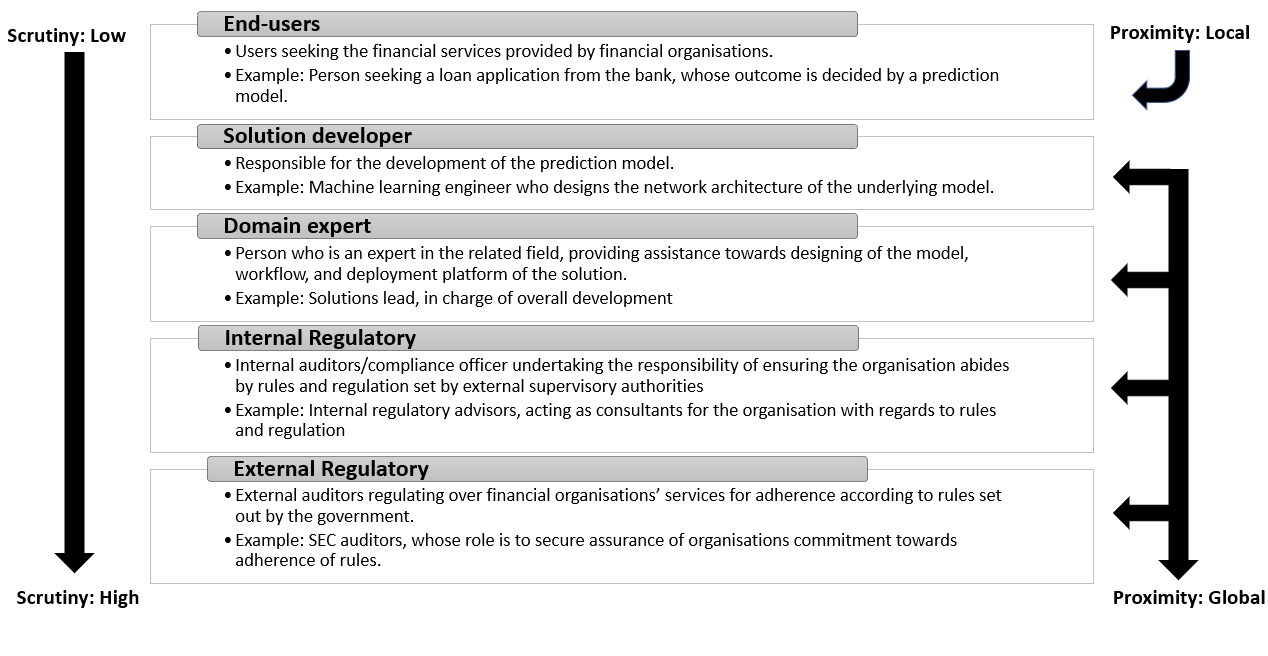}
  \caption{Levels of explanation requirements by different audiences, categorized by explanation proximity, and ordered by scrutiny level. Local proximity refers to explanations concerned about a specific outcome. Global proximity refers to the underlying reasoning and mechanics of an AI model). End-users typically require are satisfied with local-proximity explanations, and the level of scrutiny is low. Developers, domain experts and regulatory authorities require global-proximity explanations instead, and the level of scrutiny is much higher.}
  \label{fig:explanation types}
\end{figure}

As mentioned, the constitution of a good explanation is largely subjective. The amount of required information usually increases in a hierarchical manner starting from the audience to the regulatory authorities, as depicted in~Figure~\ref{fig:explanation types}. Here, scrutiny refers to the amount of information regarded as essential. On the one hand, end-users typically require the least amount of explanations (cause of outcome, data security) since they are usually only interested in resolving their practical concerns. On the other hand, external regulators require explanations about the end-product from head-to-toe (overall guideline in design process, accountable and involved personnel, deployment process, training structure of organization), including the end-users requirements. 

Proximity refers to the region of explanation provided by the XAI technique and can be classified under local (reasons about a particular outcome) and global (view of the underlying reasoning and mechanics of the AI model). End-users tend to be concerned with how the outcome affecting them is provided (local proximity). For example, a person whose credit card application was rejected would want to know the underlying reason behind it. In contrast, the solution providers and regulators tend to focus on the internal operations and design workflow of the product, for reasons related to performance enhancement, fairness in the model's sense of judgment, and identification of biases in the prediction (global proximity).

\subsection{Reasons for FinXAI}
\label{section: Reasons for FinXAI}
As mentioned in the previous section, various stakeholders lean towards different forms of explanation, naturally leading to different sets of goals the explanation can provide. A paramount reason for adopting explainable models is to ensure that financial solutions adhere to ethical standards set forth in the financial sector. The Monetary Authority of Singapore (MAS) stipulates that AI solutions should be developed in accordance with the Fairness, Ethics, Accountability and Transparency (FEAT) principles~\cite{mfa}. EU's General Data Protection Regulation (GDPR)~\cite{goodman2017european} in 2018 announced a law referred as ``right to explanation'', dictating that individuals affected by automated decision-making solutions have a right to ask for an explanation of the outcome made for them. 

The rising call for explainable models is mainly influenced by the rapid advancement of AI solutions and the increasing complexity surrounding them. \iffalse A notable example is the ChatGPT text generator rolled out by OpenAI. ChatGPT has amassed over 100M users over a short span of 2 months, making it the fastest-growing consumer product~\cite{chatgptuser}. Despite the promising capabilities of ChatGPT, such as defining complicated financial concepts~\cite{wenzlaff2022smarter,yue2023democratizing}, the reckless use of such an application can lead to an overreliance on its answers and raise ethical concerns~\cite{zaremba2023chatgpt}.\fi More importantly, public cases of AI model displaying biases in their prediction magnifies the urge for explainable solutions. A famous example is Google's image recognition software, that accidentally labels dark-skin humans as gorillas~\cite{googlecase}. Such biases can damage the company's reputation, and lead to profit losses.

The financial sector has its own set of ethics that should be upheld along with the desirable principles of AI ethics. These set of financial ethics often overlap with AI principles. An experiment involving 8 financial experts to investigate the relationship between the aforementioned sets was carried out in~\cite{rizinski2022ethically}. The results show that financial ethics (integrity, objectivity, competence, fairness, confidentiality, professionalism, diligence) has significant similarities with AI ethics (growth and sustainable development, human-centered values and fairness, transparency and explainability, safety and accountability). The strength of the links between each element was assessed, with integrity and fairness having the strongest relationship with AI ethics. Indeed, this is understandable given that AI solutions should naturally embody these qualities, regardless of the industry taken into account.
\begin{figure}[ht!]
  \centering
  \includegraphics[width=\linewidth]{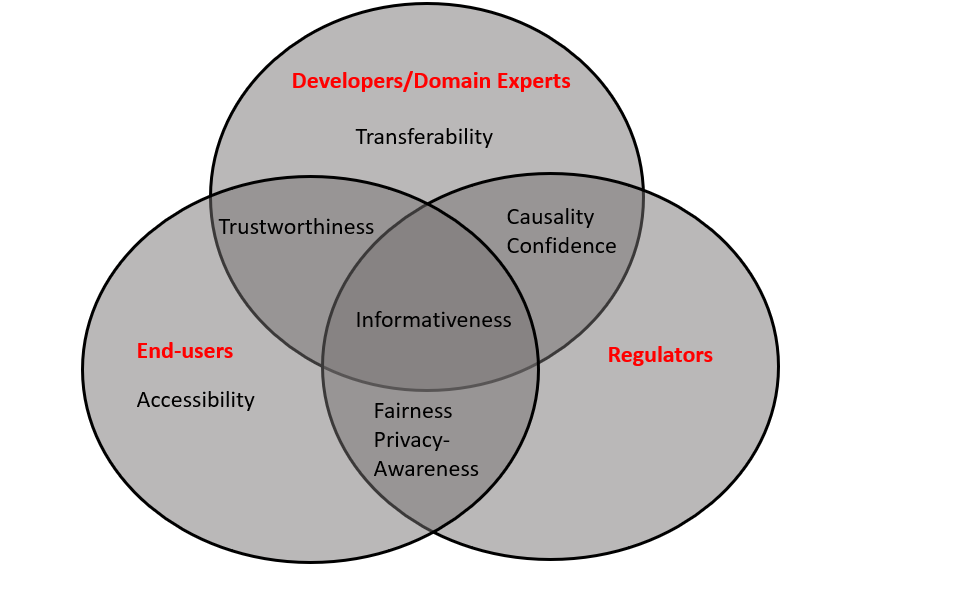}
  \caption{Ethical goals are classified under three broad audiences: end-users, developers/domain experts, and internal/external regulatory authorities. Some ethical goals are shared by the three different audiences considered, such as informativeness.~\cite{arrieta2020explainable}}
  \label{fig:XAI goals}
\end{figure}
As mentioned, the ethical goals set forth by XAI solutions differ among audiences, similar to the explanation types desired. Each audience is likely to be more affected by one than the other~\cite{arrieta2020explainable,mohseni2021multidisciplinary}. Figure~\ref{fig:XAI goals} list the ethical goals supported towards each set of audiences, and shows that there are some overlapping ethical goals across audiences. Referring to ~\cite{arrieta2020explainable}, we provide a brief explanation of each ethical goal reported in Figure~\ref{fig:XAI goals}, taking a financial perspective.
\begin{itemize}
    \item \textit{Trustworthiness}: Defined as instilling trust into users affected by the decisions of the AI model. Trustworthiness can be achieved when there is a high level of confidence in the model to constantly behave in the intended manner~\cite{ribeiro2016should}. Trust is also sustained if the services provided are transparent and enables affected user to maintain their faith in the service providers.  However, trust is a highly subjective quality and hence difficult to quantify. Judging if an explanation instills trust is mostly subjected to the affected users' opinions.
    \item \textit{Fairness}: Refers to delivering AI solutions and explanations to every user and stakeholder equally, removing possible biases. Indeed, bias mitigation is a key constituent of fairness. The transparency of the AI model allows for a fair and socially ethical analysis, where any form of biases existing in the product chain are eliminated. In the financial markets, users tend to use the services provided by a firm if they are assured of fair and unbiased treatment.
    \item \textit{Informativeness}: One important  objective of AI models is to provide assistance to human counterparts in making decisions. Therefore, it is vital that the problem statement is made clear at all times. By providing explanations, the model benefits both from a social perspective as well as a performance standpoint, since knowing what is being done opens up opportunities for further refinement. Most of the papers in the literature dealing with this aspect aim at identifying relevant features which equates to highlighting parts of the input data the model is paying attention to. This can assist in debugging and allows pruning of unnecessary features which may cause overfitting.
    \item \textit{Accessibility}: The main personnel interacting with algorithms are usually restricted to AI developers or domain experts, providing accessibility could allow for non-experts to get involved. This can be seen as an important stepping stone for making AI prevalent and well-accepted by the general society. Likewise, complicated algorithms deter financial companies from adopting such solutions, since extensive training is required while having to fear potential repercussions in the case of any unintended wrongdoings. If a model is able to relate its mechanisms in easily understandable terms, it can ease the fear of users and encourage more organizations to adopt such practices.
    \item \textit{Privacy Awareness}: Not knowing the full limits of accessibility in the data can result in a breach of privacy. Likewise, such an issue triggers concerns within the overall design workflow. Accountable personnel in the designing process should ensure third parties are only allowed restricted access to the end-users data and prevent any misuse which can disrupt data integrity. Privacy awareness is especially important in the financial sector due to the amount and sensitivity of the information being captured.
    \item \textit{Confidence}: The AI model should provide not only an outcome but also the confidence it has in the decision-making process, allowing domain experts to identify uncertainty in both model's results as well as the region of data captured. Stability in the prediction can be used to access a model's confidence while explanations provided by the model should only be trusted if it produces results that are consistent across different data inputs.
    \item \textit{Causality}: It is usually in the interest of developers or experts to understand the causality between data features. However, proving it is a difficult task that requires extensive experimenting. Correlation can be involved in assessing causality, though it is frequently not representative of causality. Since AI models only discover correlations among the data they learn from, domain experts are usually required to perform a deeper analysis of causal relationships.
    \item \textit{Transferability}: Allowing for the distillation of knowledge learned from AI models is an extensive area of research, a notable benefit is that it allows for the reusability of different models and averts endless hours of re-training. However, the complexity of the algorithms limits experts from deploying trained models in different domains. For example, a model trained to forecast future stock prices can likely be used to predict other financial variables such as bond price, market volatility, or creditworthiness, if the model behavior in these circumstances is known. Delivering an intuition of the inner workings can ease the burden of experts to facilitate adapting the knowledge learned, reducing the effort required for fine-tuning. Transferability is arguably one of the essential properties for the improvement of future AI models.
\end{itemize}

\subsection{Approach of XAI}
\label{section: How XAI}
\begin{figure}[htbp]
  \centering
  \includegraphics[width=\linewidth]{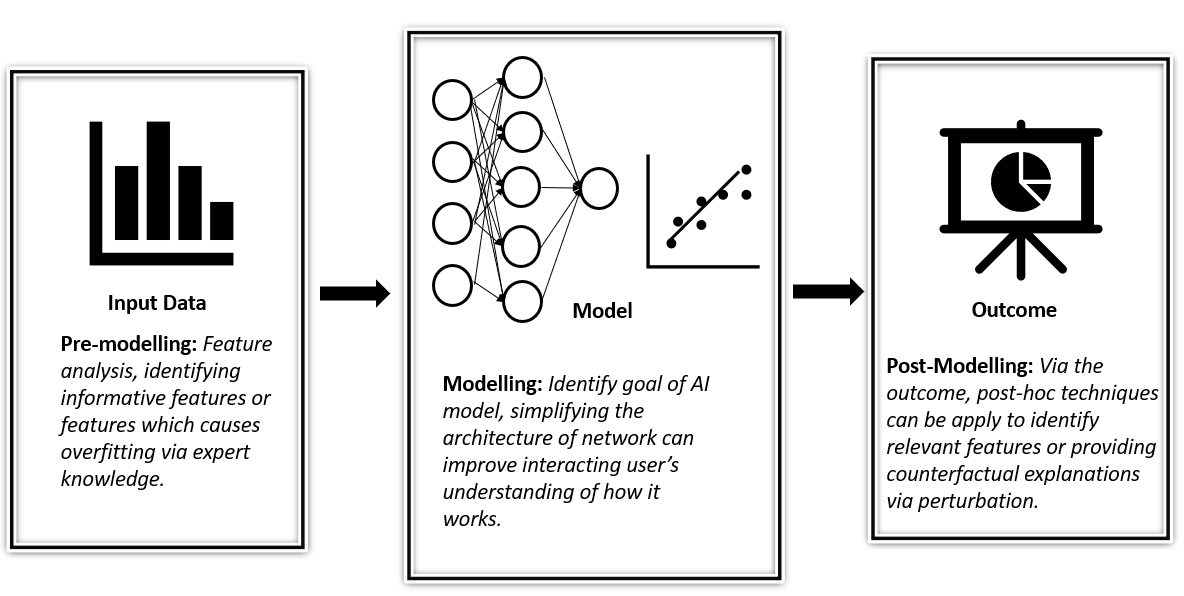}
  \caption{Different stages where interpretability can be injected into the design workflow.~\cite{xaimodelling}}
  \label{fig:XAI modelling}
\end{figure}
The review provided in this paper aims to give the readers an overall view of the XAI methodologies developed thus far in the financial industry. We note that explainability can be injected across different stages of the development cycle. These stages include: \textit{pre-modeling, modeling, and post-modeling}~\cite{xaimodelling}. Pre-modeling stage refers to the process chain before the designing stage of the AI model, this can include preliminary procedures which focus on identifying salient features by accessing readily available domain knowledge~\cite{islam2019infusing}. The modeling phase includes any adjustment to the model's architecture or optimization objective. As a start, simpler transparent models should be preferred over complex black-box models if the problem at hand is not too complicated. Most of the papers in the review focus on the post-modeling stage, mainly due to the flexibility and ease of designing explainability techniques. Since the outcome is provided, it provides developers with more information to design an appropriate explanation method towards the form of data interacted (See~Figure~\ref{fig:XAI modelling}). Most XAI techniques tend to focus on one stage of the modeling process, though it is possible to do so in two or more.\\
%\begin{table}[htbp]
%    \centering
    \begin{longtable}{|p{3.5em}|p{1em}|p{1em}|p{1em}|p{1em}|p{1em}|p{1em}|p{1em}|p{1em}|p{1em}|p{1em}|p{1em}|p{1em}|p{1em}|p{1em}|p{1em}|p{1em}|p{1em}|p{1em}|p{3.5em}|}
        \caption{Classification of papers relating to credit evaluation. The papers reviewed are split by task category and subsequently categorized by entailed properties. Missing options are either not stated or non-applicable.} \label{tab: XAI_CE} \\    %%%%<===
    \hline
        \multicolumn{1}{|p{3.5em}|}{Paper} & \multicolumn{2}{p{3.5em}|}{Trans -parency} &  \multicolumn{2}{c|}{Proximity}  & \multicolumn{5}{c|}{Explanation Procedure} & \multicolumn{4}{c|}{Audience} & \multicolumn{2}{p{3.5em}|}{Data Analysis} & \multicolumn{2}{p{3.5em}|}{Expla -nation Type} & \multicolumn{1}{p{3em}|}{Expl Eval} \\ \hline
         & \begin{sideways} Intrinsic \end{sideways} & \begin{sideways} Post-hoc \end{sideways} & \begin{sideways} Local \end{sideways} & \begin{sideways} Global \end{sideways} & \begin{sideways} Textual \end{sideways} & \begin{sideways} Visual \end{sideways} & \begin{sideways} By example \end{sideways} & \begin{sideways} Simplification \end{sideways} & \begin{sideways} Feature relevance \end{sideways} & \begin{sideways} End-user \end{sideways} & \begin{sideways} Developer \end{sideways} & \begin{sideways} Domain expert \end{sideways} & \begin{sideways} Regulatory \end{sideways} & \begin{sideways} Numerical \end{sideways} & \begin{sideways} Text \end{sideways} & \begin{sideways} Factual \end{sideways} & \begin{sideways} Counterfactual \end{sideways} & \\ \hline
         ~\cite{dikmen2022effects} &  & \checkmark & \checkmark & & & & & & \checkmark & \checkmark & & & & \checkmark & & \checkmark & &  \\ \hline
         ~\cite{gramespacher2021employing,chen2022generalized} & \checkmark & & & \checkmark & & & & & & & & & \checkmark & \checkmark & \checkmark & \checkmark & & \checkmark \\ \hline
         ~\cite{misheva2021explainable,serengil2022comparative} &  & \checkmark & \checkmark & \checkmark & & & & \checkmark & \checkmark & & \checkmark & & & \checkmark & & \checkmark &  & \\ \hline
         ~\cite{bussmann2021explainable,bussmann2020explainable} &  & \checkmark & \checkmark & \checkmark & & & & & \checkmark &\checkmark &\checkmark & & & \checkmark & & \checkmark &  &  \\ \hline
         ~\cite{rizinski2022ethically} &  & \checkmark & \checkmark & \checkmark & & & & & \checkmark & &\checkmark & & & \checkmark & & & \checkmark & \checkmark \\ \hline
         ~\cite{muller2022reshape} &  & \checkmark & \checkmark & \checkmark & & & & & \checkmark & & & & \checkmark & \checkmark & & \checkmark & & \\ \hline
         ~\cite{cazhang2022explainable} &  & \checkmark & \checkmark & \checkmark & & \checkmark & & & \checkmark & & & & \checkmark & \checkmark & & \checkmark & \checkmark & \\ \hline
         ~\cite{biecek2021enabling}  &  & \checkmark & \checkmark & \checkmark & & \checkmark & & & \checkmark & \checkmark & & & \checkmark & \checkmark & & \checkmark &  & \\ \hline
         ~\cite{crosato2021look} &  & \checkmark & \checkmark & \checkmark & & \checkmark & & & \checkmark & \checkmark & & & & \checkmark & & \checkmark &  & \\ \hline
         ~\cite{davis2022explainable} &  & \checkmark & \checkmark & \checkmark & & & \checkmark & \checkmark & \checkmark & & \checkmark & & \checkmark & \checkmark & & \checkmark & & \\ \hline
         ~\cite{sudjianto2021designing,dumitrescu2022machine} & \checkmark & & \checkmark & \checkmark & & & & & & & & & \checkmark & \checkmark & & \checkmark & & \\ \hline
         ~\cite{bueff2022machine} &  & \checkmark & & \checkmark & & & & & \checkmark & & & & \checkmark & \checkmark & & \checkmark & \checkmark & \\ \hline
         ~\cite{fritz2022financial,tran2022explainable} &  & \checkmark & & \checkmark & & & & & \checkmark & & \checkmark & \checkmark & & \checkmark & & \checkmark & &  \\ \hline
         ~\cite{srinivasan2019generating} & \checkmark  & & \checkmark & & \checkmark & & & & & \checkmark & & \checkmark & & & \checkmark & \checkmark & & \\ \hline
         ~\cite{grath2018interpretable} &  & \checkmark & \checkmark & & & & & & \checkmark & \checkmark & & & & \checkmark & & & \checkmark & \checkmark \\ \hline
         ~\cite{adams2020type} & \checkmark  & & \checkmark & \checkmark & & & & & & & & & \checkmark & \checkmark & \checkmark & \checkmark & & \\ \hline
         ~\cite{demajo2020explainable} &  & \checkmark & \checkmark & \checkmark & & & \checkmark & & \checkmark & \checkmark & & \checkmark & \checkmark & \checkmark & & \checkmark & & \checkmark \\ \hline
         ~\cite{luo2018beyond} &  & \checkmark & \checkmark & & & \checkmark & & & & & & \checkmark & & & \checkmark & \checkmark & & \\ \hline
         ~\cite{zhang2022interpretable} &  & \checkmark & \checkmark & & & \checkmark & & & &\checkmark & &\checkmark & & \checkmark & & & \checkmark & \checkmark \\
    \hline
\end{longtable}

The focused regions of finance can be broadly categorized under three sections~\cite{bahrammirzaee2010comparative}: \textit{credit evaluation} (peer-to-peer lending, credit assessment, credit risk management, credit scoring, accounting anomalies), \textit{financial prediction} (Asset allocation, stock index prediction, market condition forecasting, volatility forecasting, algorithmic trading, financial growth rate, economic crisis forecast, bankruptcy prediction, fraud detection, mortgage default) and \textit{financial analytics} (financial text classification, spending behavior, financial corporate social responsibility (CSR), customer satisfaction). 
Following the task classification, we further differentiate the studies based on the underlying characteristics of the XAI technique as shown in Table~\ref{tab: XAI_CE},~\ref{tab: XAI_FP},~\ref{tab: XAI_FM}. Specifically, we seek to answer questions such as ``\textit{What form of explanation is provided?}'' (explanation procedure), ``\textit{Who is the explanation intended for?}'' (audience), ``\textit{What kind of explanation is provided?}'' (proximity, explanation type).
% \newpage
%\begin{table}[htbp]
 %   \centering
    \begin{longtable}{|p{3.5em}|p{1em}|p{1em}|p{1em}|p{1em}|p{1em}|p{1em}|p{1em}|p{1em}|p{1em}|p{1em}|p{1em}|p{1em}|p{1em}|p{1em}|p{1em}|p{1em}|p{1em}|p{1em}|p{3.5em}|}
    \caption{Classification of papers relating to financial prediction. The papers reviewed are split by task category and subsequently categorized by entailed properties. Missing options are either not stated or non-applicable.} \label{tab: XAI_FP} \\    %%%%<===
    \hline
        \multicolumn{1}{|p{3.5em}|}{Paper} & \multicolumn{2}{p{3.5em}|}{Trans -parency} &  \multicolumn{2}{c|}{Proximity}  & \multicolumn{5}{c|}{Explanation Procedure} & \multicolumn{4}{c|}{Audience} & \multicolumn{2}{p{3.5em}|}{Data Analysis} & \multicolumn{2}{p{3.5em}|}{Expla -nation Type} & \multicolumn{1}{p{3em}|}{Expl Eval} \\ \hline
         & \begin{sideways} Intrinsic \end{sideways} & \begin{sideways} Post-hoc \end{sideways} & \begin{sideways} Local \end{sideways} & \begin{sideways} Global \end{sideways} & \begin{sideways} Textual \end{sideways} & \begin{sideways} Visual \end{sideways} & \begin{sideways} By example \end{sideways} & \begin{sideways} Simplification \end{sideways} & \begin{sideways} Feature relevance \end{sideways} & \begin{sideways} End-user \end{sideways} & \begin{sideways} Developer \end{sideways} & \begin{sideways} Domain expert \end{sideways} & \begin{sideways} Regulatory \end{sideways} & \begin{sideways} Numerical \end{sideways} & \begin{sideways} Text \end{sideways} & \begin{sideways} Factual \end{sideways} & \begin{sideways} Counterfactual \end{sideways} & \\ \hline
         ~\cite{zhang2020explainable} &  & \checkmark & & \checkmark & & \checkmark & & & & & &\checkmark & & \checkmark & & \checkmark & & \\ \hline
         ~\cite{yang2018explainable} &  & \checkmark & & \checkmark & & \checkmark & & & & &\checkmark &\checkmark & & &\checkmark & \checkmark & & \\ \hline
         ~\cite{deng2019knowledge} &  & \checkmark & \checkmark & \checkmark & & \checkmark & & & & &\checkmark & & & &\checkmark & \checkmark & & \\ \hline
          ~\cite{ghosh2021introspecting} &  & \checkmark & \checkmark & \checkmark & & & & \checkmark & \checkmark &\checkmark &\checkmark &\checkmark & & \checkmark & \checkmark & \checkmark & & \\ \hline
          ~\cite{collaris2018instance} &  & \checkmark & \checkmark & \checkmark & & & & \checkmark & \checkmark & & & \checkmark & & \checkmark & & \checkmark & & \checkmark \\ \hline
          ~\cite{benhamou2021explainable,fior2022leveraging} &  & \checkmark & \checkmark & \checkmark & & & & & \checkmark & & &\checkmark & & \checkmark & & \checkmark & & \\ \hline
          ~\cite{bracke2019machine} &  & \checkmark  & \checkmark  & \checkmark  & & & & & \checkmark  & \checkmark  & \checkmark  &  \checkmark & \checkmark  & \checkmark  & & \checkmark  & & \\ \hline
          ~\cite{nazemi2022interpretable} & \checkmark   & \checkmark  & & \checkmark  & & & & & & & &\checkmark & & \checkmark  & & \checkmark  & & \\ \hline
          ~\cite{carta2021explainable} & \checkmark   & & & \checkmark  & & & & & & &\checkmark &\checkmark & & \checkmark  & \checkmark  & \checkmark  & & \\ \hline
          ~\cite{cong2021alphaportfolio} &  & \checkmark  & & \checkmark  & & & & \checkmark  & \checkmark  & & &\checkmark & & \checkmark  & \checkmark  & \checkmark  & & \\ \hline
          ~\cite{yasodhara2021trustworthiness} &  & \checkmark  & & \checkmark  & & & & & \checkmark  & &\checkmark & & & \checkmark  & & \checkmark  & & \checkmark \\ \hline
          ~\cite{park2022interpretable,islam2019infusing,weng2022analysis,wand2022identifying,vivek2022explainable} &  & \checkmark  & & \checkmark  & & & & & \checkmark  & & & \checkmark & & \checkmark  & & \checkmark  & &\\ \hline
          ~\cite{babaei2022explainable} &  & \checkmark  & & \checkmark  & & & & & \checkmark  & & &  & \checkmark & \checkmark  & & \checkmark  & &\\ \hline
          ~\cite{lin2021xrr} & \checkmark & & \checkmark & & & \checkmark & & & & & & \checkmark & & & \checkmark & \checkmark & & \\ \hline
          ~\cite{gite2021explainable,bandi2021integrated} &  & \checkmark & \checkmark & & & & & \checkmark & & & \checkmark & & & \checkmark & \checkmark & \checkmark & & \\ \hline
          ~\cite{yan2019new} &  & \checkmark & \checkmark & & & & & \checkmark & & &  & & \checkmark & \checkmark & & \checkmark & & \\ \hline
          ~\cite{carta2022explainable} &  & \checkmark & \checkmark & & & & &  & \checkmark & &  &\checkmark & & \checkmark & & \checkmark & & \checkmark \\ \hline
          ~\cite{kumar2022explainable} &  & \checkmark & \checkmark & & & & &  & \checkmark & &  & \checkmark &  & \checkmark & & \checkmark & &  \\ \hline
          ~\cite{cho2023feature} &  & \checkmark & \checkmark & & & & &  & \checkmark & \checkmark &  & & & \checkmark & & \checkmark & \checkmark & \\ \hline
          ~\cite{shi2021xpm,kumar2017opening,chen2020explainable} &  & \checkmark & \checkmark & & & \checkmark & & & & \checkmark & \checkmark & & & \checkmark & & \checkmark & & \\ \hline
          ~\cite{achituve2019interpretable}  &  & \checkmark & \checkmark & & & \checkmark & & & & &  &\checkmark & & \checkmark & & \checkmark & & \\ \hline
          ~\cite{farzad2019determinants} &  & \checkmark &  & \checkmark & & \checkmark & & & \checkmark &\checkmark &\checkmark  &\checkmark &\checkmark & \checkmark & & \checkmark & & \checkmark \\ \hline
          ~\cite{ong2023finxabsaexplainable} & \checkmark  & & & \checkmark & & & & & \checkmark & & &\checkmark & & \checkmark & \checkmark & \checkmark & & \\ \hline
          ~\cite{yuan2020connecting} &  & \checkmark & \checkmark & & \checkmark & & & & &\checkmark & & & & & \checkmark & \checkmark & & \checkmark \\ 
    \hline
\end{longtable}

\newpage
%\begin{table}[htbp]
%    \centering
    \begin{longtable}{|p{3.5em}|p{1em}|p{1em}|p{1em}|p{1em}|p{1em}|p{1em}|p{1em}|p{1em}|p{1em}|p{1em}|p{1em}|p{1em}|p{1em}|p{1em}|p{1em}|p{1em}|p{1em}|p{1em}|p{3.5em}|}
    
    \caption{Classification of papers relating to financial analytics. The papers reviewed are split by task category and subsequently categorized by entailed properties. Missing options are either not stated or non-applicable. There were no evaluation metrics present for these papers.} \label{tab: XAI_FM} \\    %%%%<===
    \hline
        \multicolumn{1}{|p{3.5em}|}{Paper} & \multicolumn{2}{p{3.5em}|}{Trans -parency} &  \multicolumn{2}{c|}{Proximity}  & \multicolumn{5}{c|}{Explanation Procedure} & \multicolumn{4}{c|}{Audience} & \multicolumn{2}{p{3.5em}|}{Data Analysis} & \multicolumn{2}{p{4.5em}|}{Explanation Type}  & \multicolumn{1}{p{3em}|}{Expl Eval} \\ \hline
         & \begin{sideways} Intrinsic \end{sideways} & \begin{sideways} Post-hoc \end{sideways} & \begin{sideways} Local \end{sideways} & \begin{sideways} Global \end{sideways} & \begin{sideways} Textual \end{sideways} & \begin{sideways} Visual \end{sideways} & \begin{sideways} By example \end{sideways} & \begin{sideways} Simplification \end{sideways} & \begin{sideways} Feature relevance \end{sideways} & \begin{sideways} End-user \end{sideways} & \begin{sideways} Developer \end{sideways} & \begin{sideways} Domain expert \end{sideways} & \begin{sideways} Regulatory \end{sideways} & \begin{sideways} Numerical \end{sideways} & \begin{sideways} Text \end{sideways} & \begin{sideways} Factual \end{sideways} & \begin{sideways} Counterfactual \end{sideways} & \\ \hline
         ~\cite{rallis2022interpretation} &  & \checkmark & \checkmark & \checkmark & & & & & \checkmark & & & \checkmark & & \checkmark & & \checkmark & & \\ \hline
         ~\cite{zhang2022explainable} &  & \checkmark & \checkmark & \checkmark & & \checkmark & & \checkmark & \checkmark & & & & \checkmark & \checkmark & & \checkmark & \checkmark & \\ \hline
         ~\cite{maree2020towards} & \checkmark & \checkmark & & \checkmark & & & & & \checkmark & &\checkmark & & & \checkmark & \checkmark & \checkmark & & \\ \hline
         ~\cite{maree2022can} &  & \checkmark & & \checkmark & & & & & & \checkmark & & & \checkmark & \checkmark & & \checkmark & & \\ \hline
         ~\cite{maree2022understanding} &  & \checkmark & & \checkmark & & & & \checkmark & &  & &\checkmark & & \checkmark & & \checkmark & & \\ \hline
         ~\cite{lachuer2022explainable} &  & \checkmark & & \checkmark & & & &  & \checkmark &  & &\checkmark & & \checkmark & & \checkmark & & \\ \hline
         ~\cite{gramegna2020buy} &  & \checkmark & & \checkmark & & \checkmark & &  & \checkmark &  &\checkmark &\checkmark & & \checkmark & & \checkmark & & \\ \hline
         ~\cite{liu2020predicting} &  & \checkmark & & \checkmark & & \checkmark & &  &  &  & & & \checkmark & \checkmark & \checkmark & \checkmark & & \\ \hline
         ~\cite{yang2020generating} &  & \checkmark & \checkmark & & & & & & \checkmark & & & & \checkmark & & \checkmark & & \checkmark & \\ \hline
         ~\cite{ito2020ginn} &  & \checkmark & \checkmark & & & \checkmark & & & &\checkmark & &\checkmark & & & \checkmark & \checkmark & & \\ \hline
         ~\cite{newszhang2020explainable} &  & \checkmark & \checkmark & & \checkmark & & & & \checkmark & &\checkmark & &\checkmark & \checkmark & \checkmark & \checkmark & & \\
    \hline
\end{longtable}
\begin{itemize}
    \item \textit{Transparency}: As mentioned in Section~\ref{section: How XAI}, interpretability of the model is either derived via interpreting the internal mechanisms of the AI model or through external techniques aimed at delivering some form of visualization or intuition of how the model works. Most of the reviewed papers focus on post-hoc explainability techniques, which we believe are preferred for a number of reasons. Intrinsic models usually under-perform complex networks and as such, producing explanations for an inaccurate prediction is pointless. We additionally note that the method of conveying explanations for intrinsic models is by definition model-specific. This means the same method cannot be reused for a different model. While post-hoc techniques can be agnostic or specific towards any single model.
    \item \textit{Proximity}: The explanations provided by XAI tools can seek to explain either the derivation of an outcome, known as local explanation, or how the model outputs on a global scale, referred to as global explanation. Global explanations tend to provide information on how the model makes decisions globally based on the learned weights, data features, and structure of the network. Producing an acceptable global explanation tends to be difficult in most cases~\cite{molnar2020interpretable} as opposed to just a region of the input data. On the other hand, local explanations focus on a specific region of the dataset and seek to assist the receiver in understanding how a particular prediction is made. Local explanation is more accurate for unique cases where the dependency on input features is rarely captured by the AI model, which can cause global explanations to ignore such dependency. End-users tend to prefer local explanations as their concern lies with the explanation surrounding their outcome. Regulators and financial experts, on the other hand, prefer global explanations in order to have a complete understanding of the model.
    \item \textit{Explanation Procedure}: According to~\cite{arrieta2020explainable}, the various forms of post-hoc XAI techniques can be divided into several sections: text explanation (TE), visual explanation (VE), explanation by example (EE), explanation by simplification (ES) and feature relevance (FR). TE provides an explanation via text generation. Natural language tends to be easily understood by non-experts and is a common source of information in human society. VE enables visual understanding of the model's behavior, which may be preferable for image features~\cite{selvaraju2017grad}, such methods comprise graphical plots for both local and global explainability. EE captures a smaller subset of examples which represents the correlations modeled by the black-box model at a high level. ES techniques build a simpler surrogate model to approximate the underlying black-box model with high fidelity yet being interpretable. FR techniques aim to identify features deemed relevant for the model's prediction, by computing a relevance score for each feature. FR can account for explainability at both local and global levels and constitutes the largest share among the reviewed papers in our literature.
    \item \textit{Audience}: Since the quality of explanations is subjective, it is very difficult to derive a one-fit-all explanation and hence, explanations should be customized towards one's needs. The examples of audiences are referenced from Fig~\ref{fig:explanation types}, while we further merge internal and external regulators together. We highlight that aligning the objective of the explanation to the audience receiving it is important~\cite{tomsett2018interpretable}. Determining if an explanation is considered meaningful, is dependent on the target goals respective of each audience. Financial regulators, for example, would not be very concerned with understanding what sort of AI model or ML technique is used, but rather on the aspect of data privacy, model biases, or unfair treatment between affected end-users. It is uncommon for a single explanation to be deemed acceptable to audiences holding different positions in a financial company. An example is that the explanation produced for the developer tends to require additional customization before submitting to the immediate superior and the same applies to the proceeding higher-ups and external end-users.
    \item \textit{Data Type}: The most commonly used forms of input data among the reviewed papers consists of text, images, and numerical values. In terms of frequency among the forms of available data, numerical features are the most common source of information used in the financial industry. Images represent the least utilized source, as they tend to be storage intensive and contain a large amount of redundant information or are not applicable for most use cases. We only found a single work using image features.~\cite{chen2020explainable} performs classification of eight different candlestick patterns and the explanation is delivered through monitoring changes in prediction after applying adversarial attacks. Surprisingly, textual information is not used as frequently as expected, albeit being a valuable source of information for deriving market sentiment or understanding consumers' emotions towards certain aspects of the business product. It is also possible to unify multiple sources of information, otherwise known as multi-model data. A boost in performance can be achieved, for instance by combining the patterns learned from time-series features and sentiment from textual features. 
    \item \textit{Explanation Type}: A single explanation can be conveyed in various forms, including factual, contrastive, and counterfactual explanations~\cite{miller2019explanation}. Factual delivers straightforward explanations that seek to answer the question \textit{``Why does X lead to Y''} as opposed to contrastive \textit{``Why does X lead to Y instead of Z''}. Counterfactual instead reasons how the consequent can be changed with respect to the antecedent, answering the question \textit{``how to  achieve Z by changing X''}. Humans tend to prefer contrastive rather than factual explanations since the latter can have multiple answers and referring to~\cite{miller2019explanation}, explanations are selective. As humans tend to ignore a large portion of the explanations except for the important ones due to cognitive bias. For example, if Person A's loan application was rejected, there could be numerous reasons for this, such as \textit{``Person A's income was too low for the past 6 months}'', \textit{``Person A's only have 1 existing credit card''}, \textit{``Person A has had a credit default 3 months ago''} and so on. Whereas a contrastive explanation can instead involve comparing against another applicant whose outcome contrasts the target applicant's and an explanation can be made, highlighting the most significant factor. As argued by~\cite{lipton1990contrastive}, contrastive explanations are easier to deliver as one does not have to investigate the entire region of causes but rather a subset of it. Counterfactual explanations then seek to provide solutions for the contrastive explanation, commonly done by identifying the smallest changes to the input features, such that the outcome can be altered towards the alternative.
    \item \textit{Explanation Evaluation}: Despite the extensive studies carried out to investigate what defines a good explanation, it is difficult to qualitatively compare among interpretations. The quality of an explanation is mostly subjective as a single explanation can be perceived with varying opinions among audiences. Nonetheless, there exist a number of studies that provides a quantitative approach to evaluating explanations. These measurements can be derived from human experts~\cite{yang2020generating}, referencing financial ethical goals~\cite{adams2020type} or through statistical methods~\cite{muller2022reshape}. \cite{islam2019infusing} conducted a comparison between feature importance techniques in time series data and proposed a multivariate dataset that deals with the inability of techniques that identify salient time-series features. A vast majority of the reviewed papers focused only on evaluating the performance of the prediction model and consider it as a proxy for the quality of the explanation. We argue that such evaluation does not fully represent the quality of the explanation and even if so, it may not be suitable for every form of explanation procedure.
\end{itemize}
\textit{Selection Procedure}: We design a framework shown in Figure~\ref{fig:framework}, framing the designing of the XAI solution as a sequential decision-making process. The selection categories can be referenced from Tables~\ref{tab: XAI_CE},~\ref{tab: XAI_FP},~\ref{tab: XAI_FM}. The sequential structure of the framework ensures the explanation provided is tailored to the audience's needs while achieving the goal set out with respect to the target audience. We note that certain properties of the XAI technique have inner dependencies with each other, such as the relationship between explanation proximity and target audience. The quality of the explanation is evaluated and serves as feedback for any necessary adjustment, resulting in an audience-centric explanation.
\begin{figure}[ht!]
  \centering
  \includegraphics[width=\linewidth]{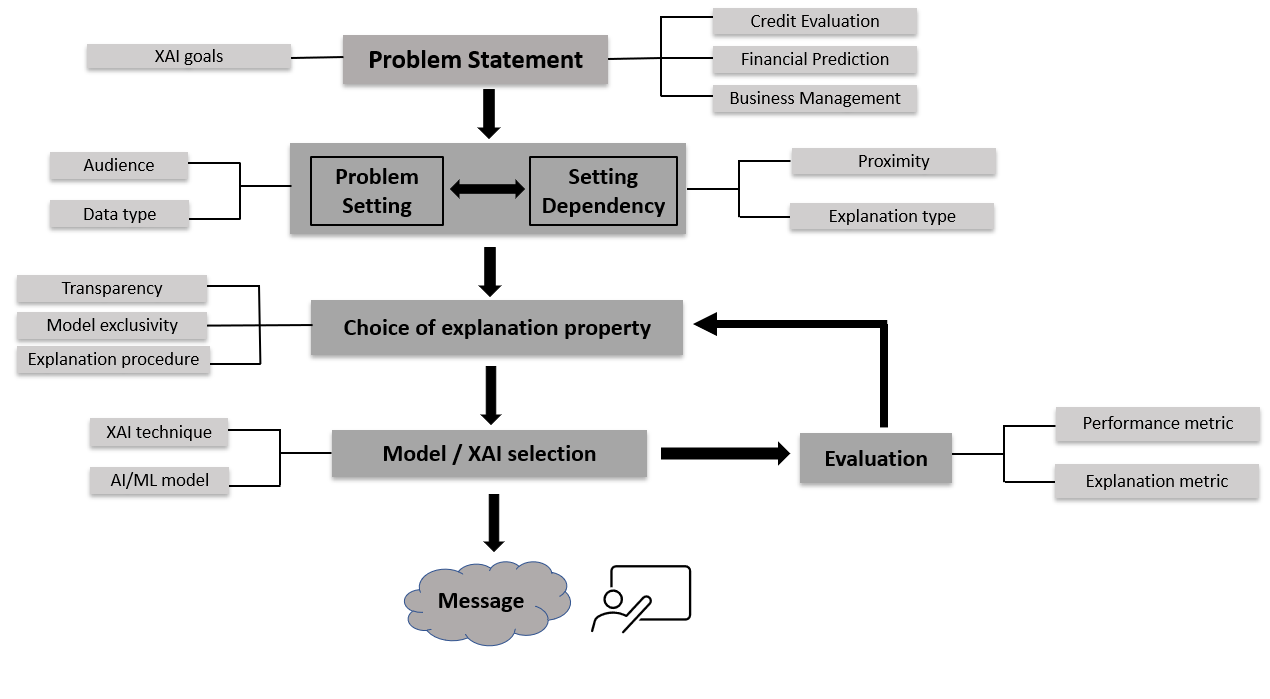}
  \caption{XAI framework depicting a sequential flow of decision-making events. The proximity (local/global) and explanation type (factual/counterfactual) should be chosen in accordance with the target audience and data type available. The choice of explanation property is assessed by an iterative evaluation under the appropriate metric for both performance and explanation conveyed.}
  \label{fig:framework}
\end{figure}
%

% \iffalse
% %%%%%%%%%%%%%%%%%%%%%%%%%%%%%%%%%%%%%%%%%%%%%%%%
% % SECTION ?
% %%%%%%%%%%%%%%%%%%%%%%%%%%%%%%%%%%%%%%%%%%%%%%%%
% \section{FinTech XAI methodology}
% In this section, we will review the collection of explainability approaches, such approaches either demonstrates the interpretability of transparent models or additionally supplement an additional layer of explainability to interpret the black-box models. Table~\ref{table: XAI overview} shows an overview of the reviewed papers, categorized accordingly to the properties in Section~\ref{section: How XAI}. This section will be broken into subsections, with Section~\ref{section: technical} highlighting technical analysis of input features, section~\ref{section: textual} on textual information and section~\ref{section: hybrid} will be mentioning models working with a hybrid of both.
% \fi

%%%%%%%%%%%%%%%%%%%%%%%%%%%%%%%%%%%%%%%%%%%%%%%%
% SECTION 3
%%%%%%%%%%%%%%%%%%%%%%%%%%%%%%%%%%%%%%%%%%%%%%%%
\section{XAI with numerical features}
\label{section: numerical}
% \iffalse
% Technical analysis (TA) is a term, commonly referred to as making financial predictions in the stock market, based on historical stock prices. The idea behind it lies with assuming the future price movements can be derived purely through historical information by means of statistic analysis~\cite{nti2020systematic}. Technical analysts typically ground their predictions on technical indicators, measuring different forms of market conditions including volatility, and growth/decline rates at different intervals. The belief that TA is able to generate substantially higher returns than the average benchmark return, otherwise known as alpha, is by contradicting the Efficient Market Hypothesis~\cite{fama1970efficient}, that there exists a lag in the stock market processing information~\cite{wong2003rewarding}. We however do note that in this paper, TA is not limited strictly to stock market activities and comprises of other financial tasks working with numerical features as well.
% \fi
Numerical features are a common source of information across all aspects of data-driven methodologies. Financial tasks such as credit scoring of individuals/firms and financial market forecasts commonly use a collection of historical numerical features, such as stock price, trade volume, and volatility, and apply various forms of data-driven models to make predictions. These data-driven models may include supervised learning approaches, e.g., classification and regression tasks, and unsupervised learning approaches, e.g., clustering tasks. The use of numerical features within the context of finance is well established, hence it is not surprising that the majority of reviewed studies focus on this area. In the following, we outline the main approaches used for explainability in this context, namely visual explanation, explanation by simplification, feature relevance, and explanation by example, and conclude with a brief summary.

\subsection{Visual Explanation} 
\begin{figure}[ht!]
  \centering
  \includegraphics[width=\linewidth]{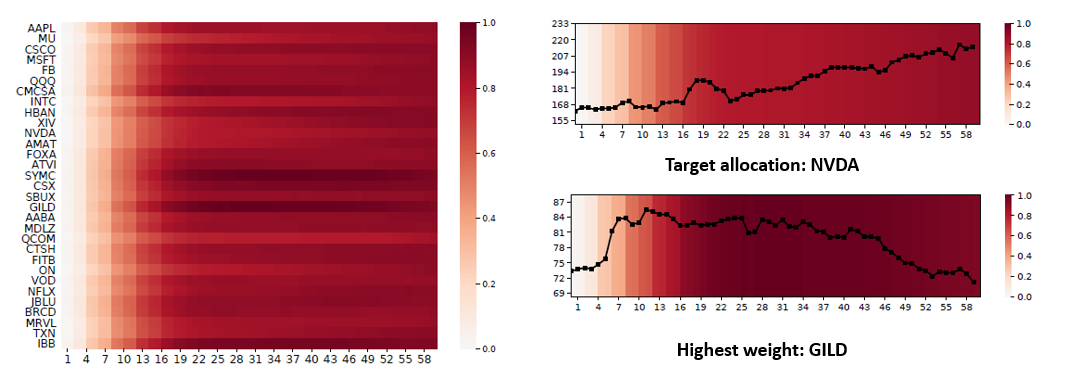}
  \caption{[left] shows a heatmap denoting the global attentiveness of individual stocks in the overall portfolio. [right] correspondingly presents a heatmap of individual assets. The agent chooses to allocate the most weight of the portfolio to NVDA, while surprisingly focusing most on a declining stock, GILD. The agent reduces the weightage of GILD and allocates to NVDA. Adapted from~\cite{shi2021xpm}. }
  \label{fig:grad-cam}
\end{figure}

Visual explanatory (VE) techniques generate explanations of the underlying model in the form of visuals. VE techniques can be both model-specific and model-agnostic. The model-specific techniques reviewed are mainly constructed to interpret image-based networks such as convolutional neural networks (CNN).~\cite{kumar2017opening} proposes to perform a deconvolution on the last layer preceding the output to extract a visual attentive map. The approach named CLass Enhanced Attentive Response (CLEAR) generates a graphical plot denoting the timeframe to which the stock-picking agent pays the most attention, along with a separate plot corresponding to the sentiment class of the stock.~\cite{chen2020explainable} implements a CNN network to identify 8 common candlestick patterns which are widely used for technical analysis in stock market trading. The authors then perform an adversarial attack on regions of the feature space, to demonstrate that the model is focusing on regions similar to how a human would process the candlesticks. \cite{shi2021xpm} employs a reinforcement learning (RL) agent to optimize a portfolio of equities while using a temporal CNN as a feature extractor. The dynamic asset allocation is interpreted with Gradient-weighted Class Activation Mapping (Grad-CAM)~\cite{selvaraju2017grad}, improving over simple deconvolutions by producing class-discriminative explanations, and is applicable to any deep neural networks. The proposed technique outputs a localization map using gradients corresponding to the target label. The right plot in Figure~\ref{fig:grad-cam} depicts a global map highlighting each asset's importance across the trading period. Interestingly in the left plot, the agent focuses on the worst-performing stock, GLID the most, rather than the high-performing stocks. Here, the agent predicts the stock decline and reduces the allocation proportion, and indirectly increases the weights of high-performing stocks which in this case is the target stock, NVDA.~\cite{achituve2019interpretable} proposes to use an attention mechanism~\cite{vaswani2017attention} to compute similarity scores of possibly fraudulent transactions on both feature and temporal levels and in return, allows for visualization at the top contributing features accounting for the model's prediction.

\begin{figure}[ht!]
    \centering
    \includegraphics[width=\linewidth]{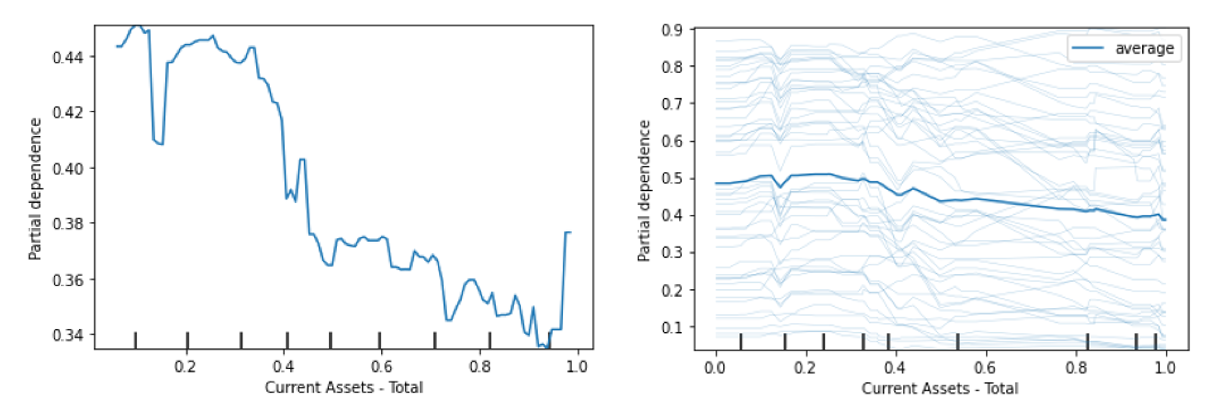}
    \caption{[left] shows a PDP on averaged marginal effects of total assets on the probability of statement restatement and [right] displays ICE, which considers instance-level relationship. Both show a negative relationship. \cite{zhang2022explainable}.}
  \label{fig:pdp}
\end{figure}

Model-agnostic VE techniques can be integrated with any form of model architecture and bear a similar resemblance with feature relevance techniques. Both investigate the effects on the model's output by adjusting the input features. \cite{cazhang2022explainable,biecek2021enabling,zhang2022explainable,farzad2019determinants} employ Partial Dependence Plots (PDP) to visualize the marginal effects of features relating to corporate distress, credit scoring, and detecting mortgage loans defaults. The generated plots can enable a way of inferring if the underlying input-output relationship is linear or complex. However, PDP has often been criticized for its assumption of independence between features, evaluating unrealistic inputs, and also conceals any heterogeneous effects of the input features. Accumulated Local Effects (ALE)~\cite{apley2016visualizing} address the concerns of feature correlation by considering the conditional distribution rather than the marginal one. In particular, it accumulates differences between intervals within the feature set to account for individual feature effects. \cite{crosato2021look} employs ALE on top of a tree ensemble model, XGBoost~\cite{chen2016xgboost}, as well as with global Shapley values~\cite{shapley1953value} for better scrutability. This work deduces that the increase in profit margin and solvency ratio leads to lower debt default rates of small enterprises. \cite{zhang2022explainable} evaluates across an arsenal of XAI techniques, encompassing the aforementioned, and also includes Individual Conditional Expectation (ICE) for financial auditing purposes. ICE differs subtly from PDP in that it considers instance-based effects rather than averaging across all instances, making it a local approach (see Figure~\ref{fig:pdp}). \cite{zhang2022interpretable} generates counterfactual explanations on credit loan applications by coupling unsupervised VAE with a supervised probit regression. The combined model yields a discriminative latent state, corresponding to class labels of either delinquency or non-delinquency. The counterfactual is subsequently produced by a stepwise manipulation function towards the opposite class label. The authors evaluate the generated counterfactuals quantitatively using maximum mean discrepancy (MMD)~\cite{zhang2022understanding}, which measures the number of successfully flipped class labels as well as minimal feature changes.

\subsection{Explanation by Simplification}
\begin{figure}[ht!]
    \centering
    \includegraphics[width=\linewidth]{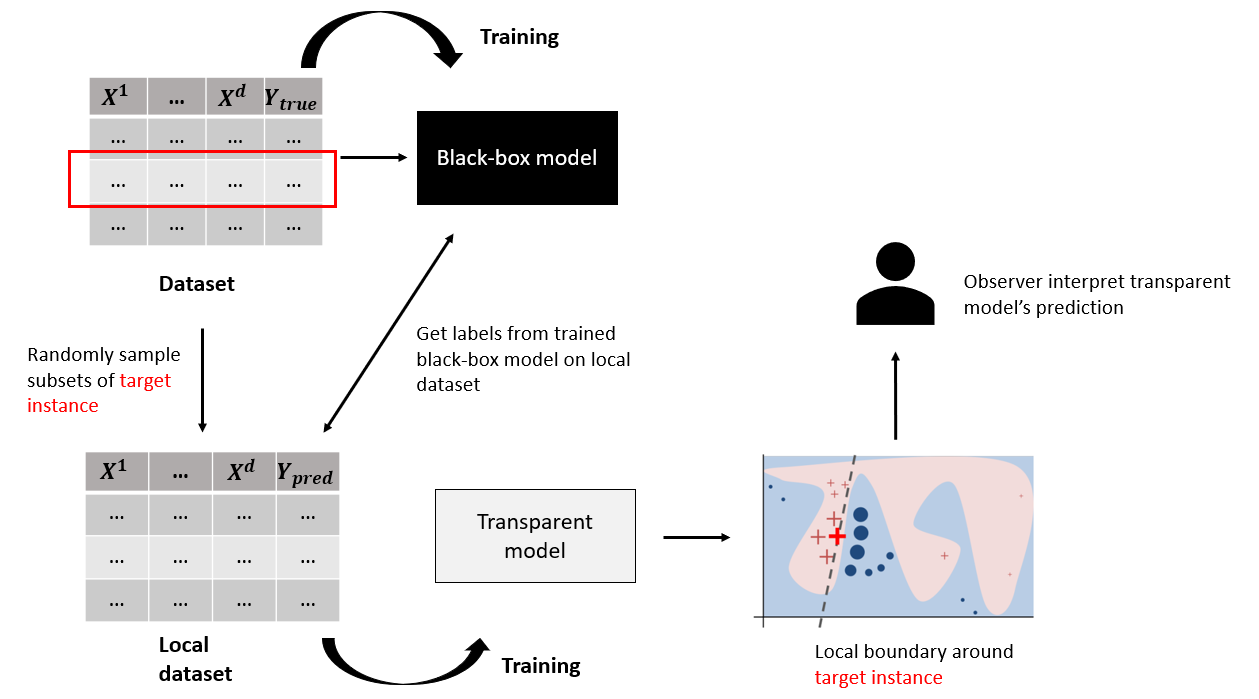}
    \caption{LIME process: Predictions of black-box model are uninterpretable. The local instance in the red box is the target to be explained. Subsets of nonzero elements of the target instance are uniformly drawn to form a local dataset on which the surrogate transparent model is trained on. The prediction from the linear transparent model can then be interpreted by the user.~\cite{ribeiro2016should}.}
  \label{fig:lime}
\end{figure}
The idea of Explanation by Simplification (ES) techniques is to introduce a surrogate model performing uncomplicated operations. The purpose is to allow the machine learning developer to formulate a mental model of the AI model's behavior. The surrogate model has to be interpretable and more importantly capture the performance of the black-box model with high fidelity. The latter property should be given a higher priority since there is little use for interpreting a low-fidelity solution. ML techniques which apply linear operations and rule extraction are applicable as surrogate models in place of uninterpretable neural networks. These include decisions tree (DT) with limited depth, linear/logistic regression, K-Nearest Neighbors (KNN), and generalized linear models (GLM).

\textit{Local Interpretable Model-agnostic Explanations (LIME)}~\cite{ribeiro2016should}: is perhaps one of the most popular explanation techniques across various use cases, including finance. LIME is a model-agnostic method that is used to provide insight as to why a certain prediction was made and can be constituted as an outcome explanation technique. Since LIME is a local-based technique, it only has to approximate the data points within a defined neighborhood, achieving a much more realistic goal instead of capturing an interpretable representation of the entire dataset. On a high level, LIME can be implemented as follows (see Fig~\ref{fig:lime}):
\begin{enumerate}
    \item The target instance to be explained is denoted as $x \in \mathbb{R}^d$. Uniformly sample $n$ random subsets of nonzero elements of $x$ to form local training points, $z \in {z_1, z_2, ..., z_n}$, where $z_i \in \mathbb{R}^d$ for $1 \leq i \leq n$. 
    \item Derive labels $f(z_i)$ for each point using the black-box model $f$. The surrogate model, $g$ is then trained on the derived dataset, $\{z,f(z)\} \in Z^n$.
    \item Choose a transparent surrogate model, $g$ and train it on the dataset, $Z^n$ via Equation~\ref{eq:lime}.
    \item Interpret the outputs of the transparent model on the target instance, $g(x)$.
\end{enumerate}
LIME minimizes the following loss function to optimize for both fidelity of the local model as well as minimal complexity.
\begin{equation}
    \label{eq:lime}
    \mathrm{explanation}(x) = \underset{g\in G}{\mathrm{argmin}}[\:L(f,g,\pi_x) + \Omega(g)]
\end{equation}
$L$ represents the loss function of the surrogate model $g$ on the labels $f$, weighted by proximity $\pi_x$. $\Omega$ represents the complexity or number of features in the surrogate model. $G$ is the set of all locally fitted models, where each explanation is produced by an individual local model. The authors additionally propose a sparse selection of features, named Submodular Pick LIME (SP-LIME), to present the observer with a global view, based on an allocated budget of maximal features to focus on. The method delivers diverse representation by omitting redundancy. \cite{misheva2021explainable,serengil2022comparative} use LIME on top of tree ensembles to identify the contributions of individual features pushing towards predicting a specific borrower as defaulting or successfully paying off the loan. Such explanations can be useful in preventing social bias by discovering any socially discriminative features on which the model may be focused, thereby instilling trust in the model's usability. \cite{yan2019new} extends LIME towards financial regulators requiring commercial banks to adhere to a set of financial factors, where they propose a method named LIMER (R stands for Regtech). The authors of LIMER argue that high acceptance of financial solutions can be achieved if such factors are integrated into the explainability design of the AI model.~\cite{collaris2018instance} implements model simplification by extracting logical rules from a random forest and selecting top most relevant rules. The decision rules are extracted from a local dataset, derived similarly to LIME without weighting the proximity of each drawn sample.~\cite{maree2022understanding} trains a recurrent neural network (RNN) to classify customer spending into five categories and an interpretable linear regression model was subsequently trained to predict the nodes formed by the RNN model. The authors then perform inverse regression which provides a mapping from output space to state space where the features responsible for categorizing customer spending can be identified.

\subsection{Feature Relevance}
\begin{figure}[ht!]
  \centering
  \includegraphics[width=\linewidth]{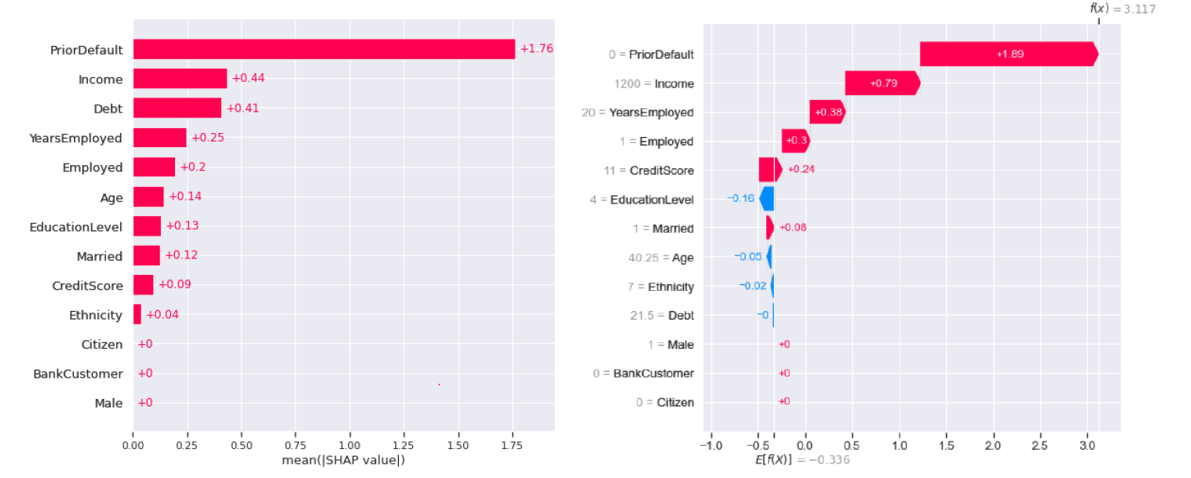}
  \caption{[left] Values of feature importance at a global level for the ML model's decision in credit card approval. [right] An example of instance-level, $E(f(X))$ represents the model's base prediction if no features were considered, and $f(x)$ represents the final prediction after summing the contributing features ($\phi_i$)~\cite{rizinski2022ethically}.} 
  \label{fig:shap}
\end{figure}

Feature relevance (FR) techniques account for the majority of the proposed explanation methodology we reviewed. FR techniques revolve around computing a relevance score for each feature, highlighting the respective contribution of the target feature either at a global or local scale. \textit{SHapley Additive exPlanations (SHAP)}~\cite{lundberg2017unified}, motivated by the fair distribution among players from game theory~\cite{shapley1953value} is a highly popular FR technique, which seeks to estimate the fair value of each feature in contributing towards the outcome, $f(x)$. The fair value, otherwise known as shapley values are determined, based on estimating the difference between the black-box function over feature subset $S$ with and without the target feature, $f(x_{S\cup\{i\}})$ and $f(x_S)$ respectively. The difference is then averaged across all possible coalitions within the feature set $F$.
\begin{equation}
    \label{eq:shap}
    f(x) = g(x') = \phi_0 + \sum^M_{i=1} \phi_i x'_i
\end{equation}
The final outcome is intuitively derived as an aggregate over all non-zero shapely values. SHAP's popularity stems from three attractive properties: guaranteeing a complete approximation of the original model $f(x)$ through additive feature attribution (see Equation~\ref{eq:shap}), ensuring non-contributing features have no impact on model output and consistency of feature values tracking the outcome contribution. We notice a large subset of papers reviewed has utilized SHAP as an explanation approach, likely given its flexibility towards explaining the model at both local and global scales (see Figure~\ref{fig:shap}). \cite{dikmen2022effects} incorporates SHAP with additional credit knowledge for the layperson to assess the logic of XGBoost's decision in a peer-to-peer lending scenario. \cite{muller2022reshape} introduces RESHAPE, designed for unsupervised deep learning networks, which provide explanations at the attribute level. Such explanations can assist auditors in understanding why an accounting statement is flagged as anomalous. The authors evaluated RESHAPE against other variants of SHAP, based on metrics measuring fidelity, stability, and robustness. Attributing to the recent frenzy in cryptocurrency which has led to a number of studies attempting to predict movements in the cryptocurrency market,~\cite{fior2022leveraging} proposes an interactive dashboard providing multiple graphical tools using SHAP for financial experts. \cite{babaei2022explainable} applies SHAP to explain predictions, generated by the popular mean-variance Markowitz model~\cite{10.2307/2975974} which is an optimization model for establishing the optimal portfolio balancing between returns and risk. The generated explanation provides regulators a means of asserting compliance of algorithmic automized traders, otherwise known as robot-advisors, with established rules and regulations. \cite{demajo2020explainable} incorporates Global Interpretation via Recursive Partitioning (GIRP) with SHAP as a global interpretability technique. GIRP uses the importance values generated by SHAP to further extract meaning insights from tree models, and the method is compared against a boolean rule technique in a credit scoring use case. \cite{bussmann2021explainable} constructs a tree-like visual explanation with TreeSHAP~\cite{lundberg2018consistent}, specifically designed for ensemble trees with an improvement in computational efficiency. The produced structure allows users to visualize clusters of similar outcomes describing company default risk.~\cite{yasodhara2021trustworthiness} compares TreeSHAP against impurity metrics using information gain, on ensemble tree models for investment quality prediction.  ~\cite{gramegna2020buy} identifies relevant features leading to consumers' decision on purchasing insurance and further clusters them into least to most likely groups with Shapley values. ~\cite{bussmann2020explainable} similarly implements SHAP to explain XGBoost's classification of credit risk, while comparing it against an interpretable logistic regression model. Other studies include discovering the relationship between corporate social responsibility and financial performance~\cite{lachuer2022explainable}, customer satisfaction~\cite{rallis2022interpretation}, GDP growth rates~\cite{park2022interpretable}, stock trading~\cite{benhamou2021explainable,kumar2022explainable}, financial distress~\cite{tran2022explainable}, market volatility forecast~\cite{weng2022analysis} and credit evaluation~\cite{rizinski2022ethically,bueff2022machine,fritz2022financial}.
\cite{wand2022identifying} performs K-means clustering on historical S\&P 500 stock information to identify dominant sector correlations which describe the state of the market. This work applies Layer-wise Relevance Propagation (LRP)~\cite{bach2015pixel}, after transforming the clustering classifier into a neural network since LRP is designed to work specifically with neural network architectures. \cite{carta2022explainable} prunes unimportant technical indicators using different configurations of a permutation importance technique, before implementing decision tree techniques for stock market forecasting. The proposed technique was compared with LIME and demonstrated better reliability.~\cite{bracke2019machine} introduces a set of feature relevance techniques, Quantitative Input Influence (QII)~\cite{datta2016algorithmic} to compute interaction effects between influential features and additionally for each multi-class label. The authors additionally evaluated the ability of the XAI technique with five questions relating to each individual audience class. All of the XAI methods shown thus far are implemented in the post-modeling stage, while ~\cite{islam2019infusing} is an example pertaining to pre-modeling where the identification of relevant features takes place before constructing the black-box model. This work explores the set of features relating to mortgage bankruptcy and performs feature mapping against a set of widely-used credit concepts. The utility of such an approach is confirmed through empirical evaluations.\\

As pointed out before, contrastive explanations are usually preferred. End-users subjected to an unfavorable AI model's decision would prefer a solution to the problem rather than a fact-based explanation which may present multiple possible reasons, giving little use to the explanation receiver. An explanation providing changes to be made such that the outcome can be reversed towards the favorable is referred to as a counterfactual explanation.  Counterfactuals are derived by computing small changes to the input features continuously until the outcome is altered to the target class. \cite{cho2023feature} first identifies significant features, attributing to bankruptcy through SHAP, and subsequently generates an optimal set of counterfactuals using Genetic Algorithm (GA). The loss optimized by GA composes of objectives describing desirable properties of a good counterfactual outcome, including minimizing the size of altered features and maximizing the feasibility of the outcome. \cite{grath2018interpretable} additionally provides positive counterfactual explanations, describing the required changes to the current inputs that would instead reverse the loan approval to rejection. Such explanations can provide some form of safety margin for the user to be mindful of.~\cite{vivek2022explainable} used various technique from DiCE~\cite{de2009ad} to generate counterfactuals under five different experimental conditions. The experiment aims to identify and study the effects of the causal variables in the fraud detection of ATM transactions.

\subsection{Explanation by Example}
Apart from techniques that identify feature relevance on varying scales or approximate with a surrogate model, another form of explanation exists by selecting representative samples to illustrate the model's behavior. Such techniques can be classified as Explanation by Example (EE). One such technique includes prototype-based explanations. Prototypes can be regarded as representatives of the entire dataset, chosen based on similarity and importance in the overall decision-making of the model. \cite{demajo2020explainable} implements protodash~\cite{gurumoorthy2019efficient}, a gradient-based algorithm in a credit loan application to select top $m$ prototypes, of which the top two are selected, with $m$ being 6. The resulting outcome is a number of representative prototypes and each instance can be represented by either generated prototype in the clusters. In this case, the proportions of allocated instances were balanced between both prototypes. The number of prototypes is a hyperparameter to be fine-tuned. A higher value of $m$ is frequently used where the complexity of the problem is a concern, albeit raises the risk of overfitting, while a lower value is used in simpler scenarios but incurs the risk of underfitting. For example, in the credit loan dataset, the selection of two prototypes was considered too little by domain experts who instead prefer 3-4 as being sufficiently representative of the evaluated dataset. \cite{davis2022explainable} similarly extracts representative instances using KNN and generates insights on out-of-sample instances by looking for similarities with the representative points. Additionally, the data points for computing the distance are instead replaced with Shapley values, taking into account the importance of input features. Representative samples are generally suitable if the user is interested in determining the types of patterns or behavior found in the dataset while being relatively fast and straightforward to implement.

\subsection{Summary of numerical features}
The above-mentioned approaches should be chosen according to the task at hand and the target audience. VE techniques, such as deconvolution and Grad-CAM, are less commonly used in the financial industry due to their limited applicability to networks other than CNNs. However, ALE, PDP, and ICE can be suitable approaches for financial analysts who might want to study the relationship between individual features and the model's outcome. ES is a straightforward approach that delegates the interpretability problem to a less complex surrogate model, though it incurs the additional cost of ensuring the faithfulness of the surrogate model. FR techniques allow users to observe each feature's contribution to the black-box prediction. Both global and local explanations serve different purposes for individual audiences. However, in situations where each feature equally contributes to the model outcome, such explanations might not be very helpful depending on the objective of the explanation. For example, a declined credit loan approval may have multiple features such as prior default, debt-to-income, and household capital contributing equivalently to the outcome. Such an explanation does not offer an obvious course of action for the applicant. EE techniques are particularly useful when the user wants a small set of representative samples to explain the model's outcome. This can provide a fast and straightforward explanation, but it has limited usefulness. 

%%%%%%%%%%%%%%%%%%%%%%%%%%%%%%%%%%%%%%%%%%%%%%%%
% SECTION 4
%%%%%%%%%%%%%%%%%%%%%%%%%%%%%%%%%%%%%%%%%%%%%%%%
\section{XAI for Textual information}
\label{section: textual}
%%%%%%%%%%%%%%%%%%%%%%%%%%%%%%%%%%%%%%%%%%%%%%%%

In this section, we review models operating with textual information. We note that the papers pertaining to this area represent the minority in the overall literature. When it comes to data preparation, textual data generally require additional pre-processing works such as stop word removal, stemming, lemmatizing, and tokenization. In terms of feature extraction, the semantics, and syntax structure around text is important and have to be learned to fully capture the information conveyed, unlike numerical features which are readily usable. Models suitable for training from textual data are also limited to a smaller subset of available techniques. Nevertheless, unstructured data such as text are in abundance. If properly processed, textual data can be used to derive informative signals such as market sentiment and emerging trends~\cite{ma2023multisource}. Textual data are commonly classified under alternative information, which comes in a wide variety of sources including social media, online reviews, blog posts, and news headlines~\cite{markok}, in contrast to non-alternative information which refers to data commonly utilized for financial analysis. Fortunately, a wide variety of explanation techniques exist which are compatible with textual information. Conveniently, textual information is applicable for XAI techniques delivering explanation via text generation, which can be preferable for the layperson as natural language provide an easier form of interpretation as compared to statistical graphs.

\subsection{Text Explanation}
\begin{figure}[ht!]
  \centering
  \includegraphics[width=\linewidth]{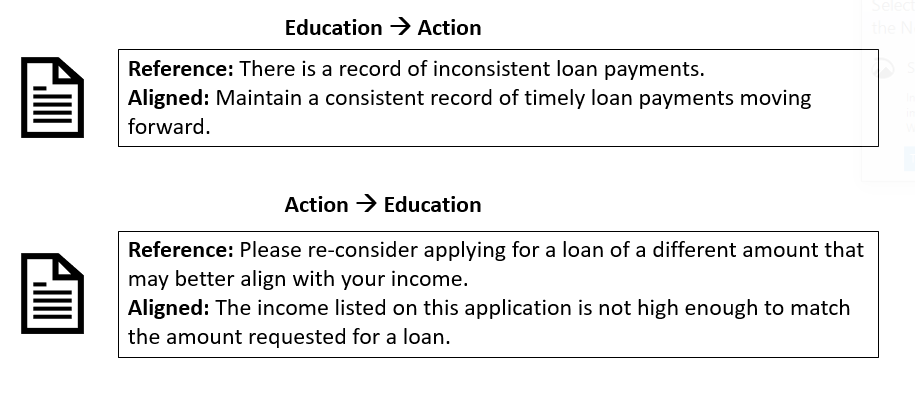}
  \caption{[Top] Transfer of educating statement to actionable statements advising applicant on actions to take such that the subsequent loan application can be approved. [Bottom] Transfer of original statement highlighting actions to educating statement conveying the reason for rejected loan application. \cite{srinivasan2019generating}. }
  \label{fig:GAN text}
\end{figure}

Text explanation techniques provide clarity in the form of generating informative textual statements to assist in the understanding of the model's behavior. The generated text can either be re-generated text statements by using some form of generative model or replacing selected words in the original sentence. \cite{srinivasan2019generating} utilizes Generative Adversarial Networks (GAN) to produce text statements that seek to align with user-defined inputs. Specifically, the explanation can take two different objectives, either converting actionable text to educational text or vice versa. The actionable text briefs audiences on optimal actions to consider, based on real-world responses from human responses on multiple loan application scenarios while the latter seeks to educate the audience on reasons attributing to the response. The transfer of objectives from one to another is analogous to the implementation of style transfer on images, a popular application of GANs that translates the style of an instance to the target image while retaining the content. Figure~\ref{fig:GAN text} shows a snippet of an example, the proposed method can identify the semantics behind the statement and relay the relationship between consistency and time, while knowing if the current income is above or below the required threshold.
\cite{yang2020generating} generates plausible counterfactual text sentences with a transformer architecture, trained under contextual decomposition. The explanation technique, derived from Sampling and Contextual Decomposition (SCD)~\cite{jin2019towards}, performs different actions including inserting, removing, or replacing words that are representative of the context in the statement, based on the target objective. The high-level idea of counterfactual generation involves identifying the most relevant word and replacing it with an antonym from a reference dictionary and continues until the outcome is reversed. The proposed transformer outperforms even human experts in classifying financial articles on merger \& acquisition event outcomes. \cite{yuan2020connecting} generates text explanations using a state-of-the-art natural language generation transformer decoder, GPT-2~\cite{radford2019language}, while fulfilling soft constraints of including keywords. The proposed technique, soft-constrained dynamic beam allocation (SC-DBA) extracts keywords corresponding to various levels of predicted market volatility using a separate network on harvested news titles. The quantitative measurement is evaluated based on the fluency and utility of the explanation produced.

\subsection{Visual Explanation for Text}
\begin{figure}[ht!]
  \centering
  \includegraphics[width=\linewidth]{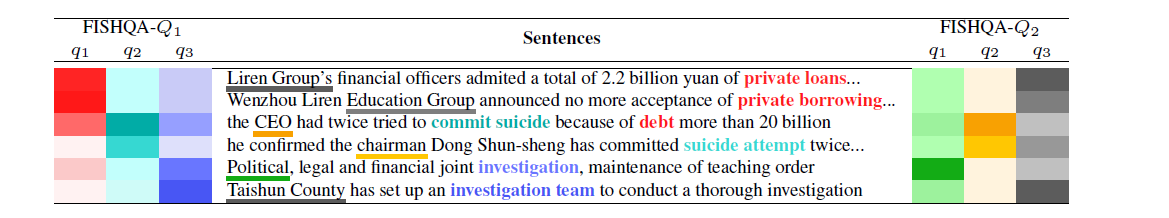}
  \caption{FISQHA: hierarchical attention model, different colors relating to different financial concepts, grey - company, light brown - executives, red - financing, blue - litigation, teal - personnel. \cite{luo2018beyond}. }
  \label{fig:FISHQA}
\end{figure}

Besides interpreting through text, users can understand through the form of visuals, which makes the use of attention a particularly attractive option. Attention was first introduced when it was used to consider correlations between words in a sentence in a parallel fashion and is a primary component in the transformer architecture~\cite{vaswani2017attention}. Transformers are notably suitable for processing long sequences of text and through the use of attention. They are computationally efficient compared to RNN-based models. It so happens that, computing attention scores of each word serves as a natural form of interpretation, by allowing users to visualize how the network is capturing information from the input text~\cite{han2022hierarchical}. Representative works in this area employ attention to highlight regions of text sentences that are deemed relevant for the output. \cite{yang2018explainable} utilizes dual-level attention with Gated-Recurrent Units (GRU)~\cite{chung2014empirical}, processing both inter-day and intra-day embedding of news titles relating to S\&P 500 companies. The attention module assigns a relevance score to each news article and the authors additionally construct a knowledge graph conducting concept mapping between relevant entities as a visual explanation. Corresponding to dual-level attention,~\cite{luo2018beyond,lin2021xrr} propose a hierarchical attention model at both the word and sentence level and produced explanations in the form of a heatmap, highlighting relevant text. The proposed method, FISHQA was trained to detect loan arrears from financial text statements, similar to the compared baselines. The uniqueness of the proposed method lies in providing FISHQA with additional user queries. The model was able to highlight regions of the statement corresponding to the set of expert-defined concepts. This form of explanation allows users to verify if the model is focusing on the correct terms relating to the concept at hand (refer to Figure~\ref{fig:FISHQA}). 
Along the lines of hierarchical attention,~\cite{lin2021xrr} introduces a quantitative measure to evaluate the precision and recall of captured against various lexicon dictionaries and expert annotated lists. The approach, analogous to the former study can be seen as an extrinsic process of ensuring the correctness of concept identification, by capturing words associated with financial risk.~\cite{deng2019knowledge} implements knowledge graphs to provide a visual linkage between event entities extracted from stock news articles. The approach offers users a visual understanding between the feature's relationship and the corresponding prediction. ~\cite{ito2020ginn} introduces GINN, an interpretable neural network, the network is designed in a way that each layer represents different entities such as words and concepts at the node level. The approach identifies words attributing to the predicted sentiment labels, as well as the concepts it belongs to.

\subsection{Summary of textual information}
TE techniques aim to augment existing input text or generate new text based on given inputs. Such explanations are commonly preferred since natural language is easily understood by humans if the explanation is concise and accurate. However, textual explanations may fail to capture the nuanced relationships between input features and model decisions. This can be especially detrimental and counterproductive to domain experts whose goal is to discover further improvements based on the provided explanations. Textual explanations might also require additional processing work to ensure fluency, coherence, and unambiguity. VE techniques in these works leverage the utility of attention to provide a glimpse into how the model is representing the input text, and the use of hierarchical attention allows for a more refined analysis. However, since attention captures the relationship between each word or sentence, such explanations might be overwhelming if the set of explainable features is too large. Audiences who are not well-versed in heatmaps or attention scores may have difficulty understanding the provided visuals.

%%%%%%%%%%%%%%%%%%%%%%%%%%%%%%%%%%%%%%%%%%%%%%%%
% SECTION 5
%%%%%%%%%%%%%%%%%%%%%%%%%%%%%%%%%%%%%%%%%%%%%%%%
\section{XAI for Hybrid information}
\label{section: hybrid}
%%%%%%%%%%%%%%%%%%%%%%%%%%%%%%%%%%%%%%%%%%%%%%%%

The remaining studies implementing post-hoc explanation techniques utilize a combination of both textual and numerical/technical features. With respect to instance-level explanations,~\cite{bandi2021integrated} combines the sentimental analysis of text with technical analysis of historical stock prices to train a random forest stock forecasting model, explained through LIME. The resulting explanation computes a set of relevant feature values and news wording corresponding to the respective outcome. Similarly,~\cite{gite2021explainable} implements LIME with LSTM-CNN and accurately identifies attentive words in consonant with the target sentiment. \cite{liu2020predicting} predicts the possibility of litigation on financial firms from examining 10-K financial reports and numerical indicators concerning the firm's accounting knowledge. The author additionally carry out an ablation study on the utility of hybrid information as opposed to individual and validated the initial approach. Correspondingly, the explanation served to regulators is framed as the identification of text leading to the suspicion of insider trading, with the help of an attention mechanism. \cite{newszhang2020explainable} adopted the practice of shapley values and further integrate external knowledge regarding truth factors, namely Truth Default Theory (TDT)~\cite{levine2014truth} to detect information fraud. The explanation module incorporates both shapley values and TDT to generate a report highlighting numerical contributions of features as well as a \textit{text explanation}. A union of \textit{explanation by simplification} and \textit{feature relevance} was proposed by~\cite{cong2021alphaportfolio,ghosh2021introspecting}.~\cite{ghosh2021introspecting} implements both LIME and SHAP, offering a global and local explanation of market fear prediction in the Indian financial market. ~\cite{maree2020towards} uses SHAP and identifies textual information to be more important for classifying financial transactions and further performs clustering to identify top contributing keywords. ~\cite{cong2021alphaportfolio} interprets an RL-trained agent's behavior in algorithmic trading. The resulting explanation enables experts to focus on time-dependent variables alongside consideration of non-linearity effects, which are reduced to a small subset of initial variables. The learned policy is simplified via policy distillation, onto the space of linear regressions such that an interpretable Lasso regression model can be used as an interpretable approximation. Subsequently, $k$-degree polynomial analysis is conducted to select salient features, with $k$ acting as an additional flexibility for the developer to decide. \cite{ong2023finxabsaexplainable} utilizes aspect-based sentiment analysis to study the relationship between stock price movement and top relevant aspects detected in tweets. The polarity of each aspect is derived from a SenticNet-based graph convolutional network (GCN)~\cite{liang2022aspect}. The proposed method can be seen as analogous to the feature relevance technique, aimed at deriving top contributing aspects with polarity values. The proposed work focuses on the relationship between financial variables instead of making financial predictions. Such information can allow for further analysis, leveraging the relationship between the price movement of individual stocks and individual sentiment of popular terms detected in tweets.\\

Hybrid information combines the utility of both numerical and textual information, which can lead to better performance and an increase in the number of compatible explanation techniques. For example, \textit{text generation} techniques can be used to generate natural language explanations for non-technical audiences, facilitating ease of understanding, while feature relevance approaches can be utilized to identify top contributing factors in the feature domain for technical experts. Models working with both numerical and textual information can also benefit from a performance point of view if such information can be processed without the risk of overfitting. However, it may be difficult for models to seamlessly perform with hybrid information, as it ultimately depends on the task at hand and may require complex feature engineering. For instance, the utility of text information largely depends on the source and often requires a significant amount of preprocessing before the data can be useful. The combination of both text and numerical features may increase the complexity of the explanation and end up being counterproductive. Such issues limit the inclusion of textual information in use cases such as stock trading or market index predictions. Nonetheless, we note that leveraging hybrid information to provide explanations can be a promising approach if the aforementioned issues are addressed.

%%%%%%%%%%%%%%%%%%%%%%%%%%%%%%%%%%%%%%%%%%%%%%%%
% SECTION 6
%%%%%%%%%%%%%%%%%%%%%%%%%%%%%%%%%%%%%%%%%%%%%%%%
\section{Explainability in Transparent models}
\label{section: transparent}
%%%%%%%%%%%%%%%%%%%%%%%%%%%%%%%%%%%%%%%%%%%%%%%%

The remaining studies look at instigating explainability from inherently transparent models. These models are typically restricted to ML models performing linear operations or rule extraction. The medium of explanation in transparent models is by nature model-specific, in the sense that the same mode of explaining how a model functions to an audience likely cannot be reused by a different model. The usability of transparent models in complicated tasks is largely restricted due to their poor predictive strength. Nevertheless, transparent models still remain an attractive option if sufficient performance can be guaranteed.

\textit{Linear/Logistic Regression}: Linear regression model is among the earliest ML models to be used for quantitative analysis. The prediction outcome can be easily derived as a weighted aggregate of input features. As such, the outcome can naturally be interpreted by inferring from the coefficients, $W_i$ which serves as a quantitative measure of feature importance for the outcome. Attributing to the linearity assumption, the output $y$ can be derived as such:  
\begin{equation}
    y = W_0 + W_1x_1 + W_2x_2 + ... W_nx_n + \epsilon
\end{equation}
One can easily interpret the outcome as \textit{``By increasing feature $x_i$ by one unit, the output increases by $W_i$''}. On the other hand, the logistic regression model is interpreted in a slightly different manner, since the output is bounded between [0,1], a logistic function is used. Logistic regression looks at the probability ratio between both outputs: \textit{``Increasing one unit of $x_i$ is equivalent to increasing \(\frac{P(y=1)}{P(y=0)}\) by $exp(W_i)$''}~\cite{molnar2020interpretable}. \cite{dumitrescu2022machine} addresses the trade-off between accuracy and interpretability by incorporating decision trees with logistic regression acting as the main operational backbone. The technique is coined as Penalised Logistic Tree Regression (PLTR). PLTR extracts binary variables from short-depth DTs, all the while establishing sparsity through a penalized lasso operation. The proposed model is able to account for non-linear effects in a credit-scoring dataset while retaining interpretability by observing the top-selected rules.

\textit{Decision Trees}: Decision trees are one of the most commonly used techniques in machine learning problems due to their simplicity and easily understandable structure. Unlike linear/logistic regression, DT can approximate nonlinear relationships and yet remain interpretable via simple \textit{if-else} logic. However, the transparency of tree models diminishes with increasing depth, and popular ensemble tree models such as XGBoost or gradient-boosting tree models completely eliminate any form of interpretability. The user can interpret decision trees by traversing through the root node and upon arrival at each leaf node. The outcome can simply be explained as \textit{``if $x_1$ is >/< $threshold_1$ AND $x_2$ is >/< $threshold_2$, $\cdots$ , outputs $Y$''}. ~\cite{gramespacher2021employing} employs a single DT and frames the loan approval task as one which maximizes profit for the lender firm. \cite{carta2021explainable} builds a lexicon dictionary associated with stock price variation, extracted from a dataset comprising both news and historical stock prices. The combined effort provides users with two forms of explanation, observed in a sequential rule-based manner as well as words correlated with the predicted market direction.

\textit{Others}:~\cite{adams2020type} constructs an interactive platform, Temenos XAI using fuzzy logic to make financial predictions. The authors demonstrated the efficacy and explainability in various downstream banking and trading scenarios. The usage of fuzzy-logic accounts for uncertainty, which is prevalent in the financial environment, and is especially useful for modeling imprecise information. The platform allows users to interpret the model on a global scale as well as at an instance-level, via observing the top contributing rules. \cite{chen2022generalized} builds on top of neural additive models (NAM)~\cite{agarwal2021neural} and introduces a generalized form of NAM, GGNAMS which focuses on sparse nonlinear interactions. GGNAMS can be regarded as an intermediate between fully connected networks and logistic/linear regression with the intent being to retain linearity and minimize excessive interactions among features while maximizing accuracy. The additive components can then be interpreted similarly to LR.~\cite{nazemi2022interpretable} similarly implements NAMs and Explainable Boosting Machine~\cite{nori2019interpretml} to identify financial drivers leading to creditor recovery rates. ~\cite{dumitrescu2022machine} proposes a hybrid approach of combining decision trees with logistic regression, capturing nonlinear effects while retaining the transparency of the model's behavior.~\cite{sudjianto2021designing} advocates for designing inherently transparent models in the pre-modeling/modeling stages and suggested a qualitative template, describing properties of model interpretability. The intention of the template is to allow researchers to ensure model interpretability while designing the model architecture. As a proof of concept, this work designs an interpretable ReLU network while conforming to the proposed template, and evaluates the network in a credit default classification task. The resulting network can be disentangled into a set of local linear models whose inherent transparency can be visualized by observing the local coefficients.\\

Transparent models have the advantage of being interpretable without requiring additional approaches to interpret the model or outcome. However, there exists a clear trade-off between desired performance and sufficient interpretability. Certain works have introduced approaches that combine different transparent models to achieve better performance while still retaining as much model transparency as possible. Transparent models remain a popular choice in the financial domain, as companies must undergo routine audits that require audited firms to provide accountability for their algorithmic services offered to end-users. Nonetheless, given the monotonic success of deep learning models, companies seeking to maintain their competitive edge must either improve on the existing transparent models or balance the performance-interpretability trade-off. A recommended approach would be to stick with transparent models if their performance proves sufficient and proceed with less interpretable models otherwise. One could also break down the task in a hierarchical manner, using interpretable models for lower-level tasks and better-performing models for more complicated tasks.

\section{FinXAI and ethical goals}
\label{section: FinXAI and ethical goals}

In Sections \ref{section: numerical}-\ref{section: transparent}, we have reviewed different FinXAI methods and summarized their technical strengths and weaknesses, based on representative papers over the past years. In this section, we analyze the contributions of these FinXAI techniques to the ethical goals that were set out in Section \ref{section: Reasons for FinXAI}. We also discuss some of the goals lacking sufficient study in current FinXAI techniques. In this section, the goal of accessibility also encompasses the fact that developers and domain experts can easily access the decision-making mechanisms of complicated black-box models due to improved interpretability. This is slightly different from the narrative that we introduced in Section~\ref{section: Reasons for FinXAI}, which mainly focuses on accessibility for non-expert users. Such an extension can better explain the technical contributions of the reviewed works to XAI.

\begin{table}[ht!]
\caption{The contributions of current FinXAI techniques to ethical goals. VE, ES, FR, EE, and TE denote visual explanation, explanation by simplification, feature relevance, explanation by example, and text explanation, respectively.}
\label{tab: FinXAI and ethical goals}
\begin{tabular}{|l|l|l|l|l|l|l|l|l|l|}
\hline
\multicolumn{2}{|c|}{Method}  & \multicolumn{1}{c|}{Trust.} & \multicolumn{1}{c|}{Fairness} & \multicolumn{1}{c|}{Informative.} & \multicolumn{1}{c|}{Accessibility} & \multicolumn{1}{c|}{Privacy} & \multicolumn{1}{c|}{Confidence} & \multicolumn{1}{c|}{Caus.} & \multicolumn{1}{c|}{Trans.}\\ \hline
\multirow{14}{*}{\begin{tabular}[c]{@{}l@{}}Num.\\ Feat.\end{tabular}}  & VE & \cite{zhang2022interpretable} &  & {\begin{tabular}[c]{@{}l@{}} \cite{achituve2019interpretable} \cite{biecek2021enabling} \cite{chen2020explainable} \\ \cite{crosato2021look} \cite{farzad2019determinants} \cite{kumar2017opening} \\ \cite{shi2021xpm} \cite{zhang2022explainable} \\ \cite{cazhang2022explainable} \end{tabular}} & {\begin{tabular}[c]{@{}l@{}} \cite{achituve2019interpretable} \cite{crosato2021look} \cite{zhang2022explainable} \end{tabular}} &  &  &  & \\ \cline{2-10} 
  & ES & {\begin{tabular}[c]{@{}l@{}}\cite{collaris2018instance} \cite{maree2022understanding} \\ \cite{yan2019new}\end{tabular}} & \cite{misheva2021explainable} \cite{serengil2022comparative} &  &  &  &  &  & \\ \cline{2-10} 
  & FR & {\begin{tabular}[c]{@{}l@{}}\cite{babaei2022explainable} \cite{cho2023feature}\\ \cite{grath2018interpretable}\end{tabular}} &  & \cite{bussmann2021explainable} & {\begin{tabular}[c]{@{}l@{}} \cite{babaei2022explainable} \cite{benhamou2021explainable} \cite{bracke2019machine} \\ \cite{bueff2022machine} \cite{bussmann2020explainable} \cite{carta2022explainable} \\ \cite{demajo2020explainable} \cite{dikmen2022effects} \cite{fior2022leveraging} \\ \cite{fritz2022financial} \cite{gramegna2020buy} \cite{islam2019infusing} \\ \cite{kumar2022explainable} \cite{lachuer2022explainable} \cite{muller2022reshape} \\ \cite{park2022interpretable}  \cite{rallis2022interpretation} \cite{rizinski2022ethically} \\ \cite{tran2022explainable} \cite{vivek2022explainable} \\ \cite{wand2022identifying} \cite{weng2022analysis} \\ \cite{yasodhara2021trustworthiness}  \end{tabular}} &  &  & \cite{vivek2022explainable} & \\ \cline{2-10} 
  & EE & \cite{davis2022explainable} &  &  &  &  & \cite{davis2022explainable} \cite{demajo2020explainable} &  & \\ \hline
\multirow{2}{*}{\begin{tabular}[c]{@{}l@{}}Text.\\ Info.\end{tabular}} & TE & \cite{yang2020generating} &  & \cite{srinivasan2019generating} \cite{yuan2020connecting} & \cite{srinivasan2019generating} \cite{yang2020generating} &  &  & \cite{yuan2020connecting} & \\ \cline{2-10} 
  & VE &  &  & \cite{lin2021xrr} \cite{luo2018beyond} \cite{yang2018explainable} & \cite{deng2019knowledge} \cite{ito2020ginn} \cite{lin2021xrr} &  &  &  & \\ \hline
\multicolumn{2}{|l|}{Hybr. Info.}  &  &  & {\begin{tabular}[c]{@{}l@{}} \cite{bandi2021integrated} \cite{cong2021alphaportfolio} \cite{ghosh2021introspecting} \\ \cite{gite2021explainable} \cite{liu2020predicting} \cite{maree2020towards} \\ \cite{newszhang2020explainable} \end{tabular}} & {\begin{tabular}[c]{@{}l@{}} \cite{bandi2021integrated} \cite{cong2021alphaportfolio} \cite{ghosh2021introspecting} \\ \cite{gite2021explainable} \cite{maree2020towards} \cite{ong2023finxabsaexplainable} \end{tabular}} &  &  &  & \\ \hline
\multicolumn{2}{|l|}{Tran. Mod.}  &  &  &  & {\begin{tabular}[c]{@{}l@{}} \cite{adams2020type} \cite{carta2021explainable} \cite{chen2022generalized} \\ \cite{dumitrescu2022machine} \cite{gramespacher2021employing} \cite{nazemi2022interpretable} \\ \cite{sudjianto2021designing} \end{tabular}} &  &  &  & \\ \hline
\end{tabular}
\end{table}

As seen in Table~\ref{tab: FinXAI and ethical goals}, different explainable methods contribute to the ethical goals from different aspects. XAI for numerical features has proposed several methods to approach ethical goals, regarding trustworthiness, fairness, informativeness, accessibility, confidence, and causality. For VE methods, \cite{zhang2022interpretable} is one example that advocates for trustworthiness by generating counterfactual explanations, while being informative by detecting important features that can alter the prediction. Counterfactual explanations reveal the slightest modifications that are necessary on the input data to achieve an alternative outcome. It helps to earn trustworthiness from target audiences because counterfactual explanations provide possible rescue measures for them to achieve their targets with minimum effort, e.g., proposing possible improvements to help borrowers pass the qualification review of credit agencies. Counterfactual explanations can also help to justify predictions besides factual explanations. Both merits of generating counterfactual explanations improve the trustworthiness of audiences. Many VE-based approaches improve informativeness by gaining insights into the decision-making mechanisms of models and revealing feature correlations \cite{achituve2019interpretable,biecek2021enabling,chen2020explainable,crosato2021look,farzad2019determinants,kumar2017opening,shi2021xpm,zhang2022explainable,cazhang2022explainable}, because visualization takes the advantages of demonstrating patterns and trends of data, e.g., model parameters and numerical features. VE can be also used to discover valuable features \cite{achituve2019interpretable,crosato2021look,zhang2022explainable}. It is easy to communicate with both experts and non-domain experts by using graphical representations or visual images whenever available. Thus, VE also improves accessibility for broader audiences. For ES methods, \cite{collaris2018instance,maree2022understanding,yan2019new} can explain why a certain prediction was made from outputs, which helps to improve the trustworthiness of AI predictions. \cite{misheva2021explainable,serengil2022comparative} use LIME to detect socially discriminative features to prevent social bias. ES is the only approach that was used for improving fairness in finance. For FR methods, \cite{babaei2022explainable} uses SHAP to improve the trustworthiness of algorithmic traders in crypto markets. \cite{cho2023feature,grath2018interpretable} generate contrastive explanations to explain required changes for certain predictions. \cite{bussmann2021explainable} visualizes similar outcomes that describe the risk of a company default with SHAP, while most SHAP-based methods \cite{babaei2022explainable,benhamou2021explainable,bracke2019machine,bueff2022machine,bussmann2020explainable,carta2022explainable,demajo2020explainable,dikmen2022effects,fior2022leveraging,fritz2022financial,gramegna2020buy,islam2019infusing,kumar2022explainable,lachuer2022explainable,park2022interpretable,muller2022reshape,rizinski2022ethically,tran2022explainable,vivek2022explainable,wand2022identifying,weng2022analysis,yasodhara2021trustworthiness} improve accessibility for technical audiences by discovering important features. \cite{fior2022leveraging} improves usability by constructing interactive graphical tools upon SHAP, which likewise promotes accessibility. \cite{vivek2022explainable} is one of the rare works that study causal inference based on generated counterfactuals. For EE methods, \cite{davis2022explainable}  generates counterfactuals to explain the required changes, based on representative instances. The selected representatives of similar instances by EE methods \cite{davis2022explainable,demajo2020explainable} can be used to select instances to represent a particular cluster in the output space. Similar instances aligned to such representatives can assure and improve the confidence of stakeholders. 

XAI for textual information targets to improve trustworthiness, informativeness, accessibility, and causality. For TE methods, \cite{yang2020generating,srinivasan2019generating} similarly improves trustworthiness with counterfactual texts, with the latter providing alignment according to the user's prompt. The alignment from educational to actionable information enhances information flow, especially for individuals not familiar with the service interface. For VE techniques operating on text, attention weights are widely used for interpretation purposes. Such techniques enhance informativeness and accessibility by using the attention weights to understand regions of focus by the underlying model~\cite{lin2021xrr,luo2018beyond,yang2018explainable}. On the other hand, \cite{deng2019knowledge,ito2020ginn} improve the interpretability of graph neural networks in the financial domain.

XAI for hybrid information leverages both textual and numerical features to improve informativeness and accessibility. These works interpret the black-box model's behavior  \cite{bandi2021integrated,cong2021alphaportfolio,ghosh2021introspecting,gite2021explainable,liu2020predicting,maree2020towards,newszhang2020explainable}, and provide textual evidence regarding predictions \cite{bandi2021integrated,cong2021alphaportfolio,ghosh2021introspecting,gite2021explainable,maree2020towards,ong2023finxabsaexplainable}. \cite{adams2020type,carta2021explainable,chen2022generalized,dumitrescu2022machine,nazemi2022interpretable,gramespacher2021employing,sudjianto2021designing} implements transparent models, mitigating the need for post-hoc analysis, and the simplicity of such models improves upon the accessibility for non-expert users.  Transparent models may be a suitable choice if the performance is satisfactory and the outcome has to be readily interpretable by non-technical stakeholders. Decision trees are one model which can be easily communicated to audiences without a technical background, given its easily understandable format.

From the above works, we can find that most of the XAI research lies in either studying the underlying model's behavior or identifying important features. Notably, the connotations of informativeness and accessibility goals are rich as seen in Table~\ref{tab: FinXAI and ethical goals}, no XAI technique can achieve all of the desired goals. It is therefore imperative that the XAI design process is tailored towards the desired goals of the target audience Fig.~\ref{fig:XAI goals}. Likewise, the format of presenting explanations is equally important. Non-technical audiences would very much prefer user-friendly visuals as compared to technical plots.
%Understanding model mechanisms and important features may be particularly useful for developers and domain experts but informativeness is also wanted by many other audiences, e.g., end-users, and internal and external regulatory entities. Understanding their information needs and developing explanation-informative models can achieve more diverse informativeness-driven FinXAI models. Similarly, many FinXAI models with accessibility advantages focus on the accessibility of important features, which may be also of interest to developers and domain experts. Simply delivering important features may not be the focus for non-expert users. In contrast, appropriate presentation methods, e.g., via text \cite{srinivasan2019generating,yuan2020connecting} or via graphical tools \cite{adams2020type,fior2022leveraging} can help them better understand the messages sent from those important features.

On the other hand, the ethical goal of preserving data-privacy has not been well studied in the works reviewed. Privacy-preserving techniques are a popular research direction, e.g., federated learning \cite{yang2019federated}. It is a decentralized learning method that allows parties to collaboratively train a model within a local environment without sharing their data with each other. Generating synthetic data in place of actual data for training models can be one such approach. One example is LIME which generates a local dataset given a target instance, without requiring access to other data instances. This can help to minimize the amount of information being accessed outside of the accountable circle. The understanding of how various features lead to a certain output creates the opportunity of generating more synthetic data. This can be seen as a form of self-supervised learning with the purpose of preserving privacy. However, XAI techniques can also become a double-edged sword, attributing to privacy leakage instead. Such concerns are especially prevalent in techniques manipulating decision boundaries including SVM, K-nearest neighbors, and counterfactual explanations~\cite{sokol2019counterfactual}. For example, a counterfactual explanation on reserving a loan application might reveal a suite of sensitive information (location, 10-year income, marital status) to be modified, even though such information is meant to be anonymized. The leaked information can be accessed by third-party providers who may be part of the product design or malicious hackers. A key challenge is managing the balance between the fidelity of the delivered explanation and the sensitive features altered. Data leakage goes against the privacy awareness goal of XAI and such events are not rare in the financial sector where there exists a constant supply of computerized bots looking to capitalize on these openings. The consequences often affect a large group of public stakeholders~\cite{insurancebreach}, and the affected firm has to pay large fines and incur a loss of trust from their clients. In addition, overly-expressive explanations may allow external competitors to reverse-engineer the models and potentially replicate and improve upon them, thereby compromising the competitive edge a company holds.

XAI techniques that improve transferability are another less frequently studied area. In the field of general AI, transferable knowledge is usually acquired through transfer learning~\cite{neyshabur2020being}, multi-task learning~\cite{mao2021bridging}, meta-learning~\cite{he2022metabased}, and domain adaptation~\cite{xie2022active}. However, the main carrier of these learning paradigms at present is usually deep neural networks. It is difficult to acquire explainability for a deep neural network by using these learning methods. In addition, knowledge forgetting also brings challenges to traditional neural network-based learning methods~\cite{he2022jcbie}. Thus, the old knowledge stored in the neural network is likely to be replaced by the learned new knowledge, if the old knowledge is not retained together with the new knowledge. In light of this, how can we leverage explainability and transferability, simultaneously? One possible direction is to utilize neural symbolic techniques. Neural symbolic AI has achieved significant impacts in natural language processing (NLP), e.g., sentiment analysis~\cite{cambria2022senticnet} and metaphor processing~\cite{mao2022metapro}. It takes the merits of both neural networks and symbolic representations. For example, neural networks have strong generalization ability in learning feature representations. Symbolic reasoning enables human-understandable explanations of the system's decision-making process through transparency and interpretability. Since symbolic knowledge can be readily stored in a knowledge base permanently, it avoids the problem of knowledge forgetting in neural networks. A comprehensive and accurate knowledge base can weaken the fitting ability of the neural network. As a result, a more lightweight and transparent neural network can be used in a neural symbolic system. However, developing domain-specific knowledge for finance is costly. Besides, developing symbolic representations for numerical data is also challenging.

Finally, improving fairness, confidence, and causality is also important for ethical concerns. Whereas, the FinXAI research in these areas is very limited. As noted in Table~\ref{tab: FinXAI and ethical goals}, there are not many explanation methods that approach these goals, e.g., ES for fairness with numerical features; EE for confidence with numerical features; and FR and TE for causality with numerical and textual features, respectively. However, it is difficult for a one-fit-all explanation. Hence, we highlight the importance of an audience-centric XAI technique as a more realistic expectation. 

%%%%%%%%%%%%%%%%%%%%%%%%%%%%%%%%%%%%%%%%%%%%%%%%
% SECTION 7
%%%%%%%%%%%%%%%%%%%%%%%%%%%%%%%%%%%%%%%%%%%%%%%%
\section{Challenges and Future directions}
\label{section:challenges}
%%%%%%%%%%%%%%%%%%%%%%%%%%%%%%%%%%%%%%%%%%%%%%%%

We exploit the knowledge and insights gained from the agglomeration of FinXAI research conducted thus far and put forward a list of challenges and directions we consider to be important for readers to consider. A few of these limitations have been similarly considered in previous works~\cite{chen2021seven}, which have presented seven major challenges encountered in the context of presenting explanations to stakeholders. Some of these limitations are evident from the reviewed XAI methodologies and we further elaborate on them and cater avenues for improvement.

\subsection{Over-reliance}
A means of interpreting the model can be helpful while transforming how users interact with data. However, it can cause users to over-rely on possibly inaccurate explanations. A survey was conducted to study how data scientists perceive explanations provided by different XAI tools and found out a large proportion tend to over-trust the explanations provided~\cite{kaur2020interpreting}, especially the ones which have received widespread usage. The visual explanations delivered by feature relevance techniques such as SHAP, tend to be absorbed at face value, which can cause researchers to not question their legitimacy. Concurrently, a data scientist who has spent an enormous amount of time designing the AI model may already have prior beliefs on the outcome or model and are more inclined to accept the explanation if aligned with their initial beliefs~\cite{hohman2019gamut}. Such an occurrence is commonly known as confirmation bias. Over-trusting these explanations can be especially damaging if conveyed to the layperson and can result in the spread of misinformation to a wider audience.

\subsection{Social aspects}
As mentioned by~\cite{miller2019explanation}, explanations are selective, the receiving users tend to only take a minor subset of the entire set of explanations, predominantly those that agree with their prior belief. This can sometimes cause the affected receiver to lose sight of the bigger picture and arrive at some misinterpreted conclusion.~\cite{kaur2020interpreting} notes this as a mismatch between the solution's conceptual purpose and the receiver's mental model. XAI tools that produce a feature ranking figure may overcloud the users with excessive information, thereby increasing their cognitive load and rendering the tool counterproductive. It is also observed that the amount of trust is correlated with the level of appreciation the receiver has in the explanation~\cite{mohseni2021machine}. Take for example the case of a rejected loan application, an under-appreciated explanation would just result in the applicant resubmitting the application to a different bank, without addressing the underlying root cause. Humans also tend to prefer contrastive explanations as opposed to visualizing a large number of probable causes, thus designing the explanation to be counterfactual can reduce under-appreciation and rejection of XAI tools. Future research on human-centric explanations can look to draw inspiration from social sciences and the study of human psychology to bridge the gap between the two ends of the explanation chain.

\subsection{Explanation evaluation}
It is evident from Table~\ref{tab: XAI_CE},~\ref{tab: XAI_FP},~\ref{tab: XAI_FM}, only a small subset of reviewed works attempt to provide some form of quantitative measurement of the proposed XAI technique. An even smaller number performs a comparison between multiple XAI techniques, possibly due to model incompatibility and differences between the explanatory structure of individual XAI techniques.~\cite{gurumoorthy2019efficient} uses a variety of evaluation approaches, grounded on both statistical and human knowledge and involves experts and non-technical users, while admitting the limitation of ambiguity and inconsistency in human judgment. In the case of surrogate model explanations, the fidelity of the surrogate model can be used to measure the accuracy of the approximation. However, such an approach cannot be used for feature relevance tools like SHAP~\cite{amparore2021trust}. A common basis for instilling interpretability in financial solutions goes beyond the social responsibilities of the provider firms but also concerns the need to comply with the rules and regulations laid out. Even so, the mismatch between each party's perception of explanation sufficiency extends beyond the model itself and includes commonly neglected variables such as accountable personnel, feedback process, and personnel training procedure~\cite{kuiper2022exploring}. \cite{hoffman2018metrics} highlights that explanations can be seen as a dynamic interaction between the conveyed message and the receiver's thought process. The effectiveness can be measured via goodness and satisfaction in the form of feedback upon receiving the message. This further exemplifies the fact that what makes an explanation good is largely subjective and coming to a consensus on a suitable set of metrics is no trivial task. Among the set of reviewed works in this paper, there exist two forms of evaluation, either through statistical approaches (F1-score, accuracy, and t-test) or opinions of a human expert. The latter is defined as plausibility and should be made distinct from faithfulness, which reflects how the AI model reasons about its behavior. \cite{jacovi2020towards} states that the assessment of faithfulness should be independent of human judgment and a common ground can be established by evaluating XAI techniques with respect to a pre-defined set of goals, rather than on the basis of achieving universal satisfaction. One criticized flaw in existing works on evaluating interpretation methods pertains to the distributional shift between training and test sets, as well as the infeasible requirement of retraining models. \cite{turbé_bjelogrlic_lovis_mengaldo_2023} creates a set of synthetic datasets with known discriminative features and additionally develops two new metrics which account for identifying top relevant time steps in terms of ranking and score. The proposed method takes into consideration the temporal elements in time-series analysis. We further note that it is imperative for future works on XAI evaluation criteria to precisely define the objective and target audience of the explanation.

\subsection{Trade-off between performance and interpretability}
It is often common to ponder \textit{``shouldn't we deploy more transparent models if no interpretability enhancement work is required?''}, such an initiative is often plagued by the limited representativeness of transparent models. The trade-off between performance and interpretability is quite commonly a major cause of the dilemma in selecting between black-box models and inherently transparent models. Though there exist studies that have shown that black-box models performing more complex operations do not necessarily lead to better performance~\cite{rudin2019stop,rudin2019we}, it is often the case for unstructured information and noisy environments such as the financial markets. XAI tools explaining through a surrogate model have to face the burden of ensuring both the fidelity of the surrogate model and the effectiveness of the underlying AI model, all the while matching the required goals toward the receiving audience. In light of such a challenge, it highlights the necessity for a consensus metric to serve as a quantitative assessment. In general, it happens more often than not that the selected model at hand is more complex than required, resulting in additional explainability engineering. Moving forward, an efficient way of handling the trade-off is to prioritize the usage of transparent models if the obtained performance is satisfactory and progress to a more complex model when necessary.

\subsection{Better transparent models}
We note that transparent models refer to models which exhibit inherent transparency without the need to apply post-hoc explainability techniques. However, caution should be taken in the assumption of such model~\cite{jacovi2020towards}. The inherent transparency is dependent on the achievable explanation goals and the explanation receiver, while there is much doubt surrounding truly inherently transparent models~\cite{serrano2019attention}. Nevertheless, there are numerous studies advocating for a greater need in adopting transparent models. A study by~\cite{lipton2018mythos} argues that transparent models are essential for promoting fairness in machine learning, as they allow for easier identification and mitigation of biases in the decision-making process. A team of researchers~\cite{rudin2019we} participated in an explainable machine learning challenge and concluded that transparent models do not only sidestep the common issues of trust and misinterpretation but also exhibit the potential to match complex models in terms of performance on specific tasks. It can therefore be for the well-being of society that researchers prioritize the development of more sophisticated and robust transparent machine learning models that are able to balance the trade-off between model accuracy and interpretability.

\subsection{Human-centric XAI}
We note that, in order to effectively support human decision-making and ensure the interpretability of AI models, there is a growing need for human-centric XAI tools that prioritize user understandability and usability. Explanations are interactive and should be viewed as a bidirectional form of communication~\cite{kaur2020interpreting}, with the XAI tool explaining to the user and the user reciprocating back for clarity. Incorporating Human-Computer Interaction (HCI) principles into the design of XAI systems is beneficial in translating interpretability, as HCI principles embrace user-friendly interfaces that enhance human comprehension and engagement. One important aspect of this is the development of interactive systems that enable users to actively engage with and explore XAI models, allowing them to gain a deeper understanding of how the models work and how they can be understood. Several studies have highlighted the importance of incorporating HCI principles into the development of interactive XAI systems~\cite{hohman2019gamut}. The results show that users had more trust when presented with virtual interactive explanations~\cite{weitz2019you}. Some popular examples of interactive XAI toolkits include Microsoft AI widgets~\cite{microsofttool} and What-if tool by Google~\cite{wexler2019if}. Besides being an easily approachable and interpretable tool, interactive systems improve system usability and entice users to frequent the financial services provided, thus adding to the benefits of financial firms prioritizing the development of human-centric, interactive XAI tools.

\subsection{Multimodal XAI}
A less discussed avenue for improvement is in the incorporation of multimodal information, particularly natural language. Among the works reviewed, a large subset of processed input data only involves numerical features, while the inclusion of textual information remains a minority. An underlying reason might be due to the redundancy of using textual data or the lack of substantial increase in performance while requiring additional pre-processing works due to the inclusion of such information. Nevertheless, there exist benefits from a transparency point-of-view in incorporating textual information.~\cite{danilevsky2020survey} highlighted that the separation of the underlying AI model from the explainability tool is less distinct since NLP models, particularly through the use of attention can produce both the prediction and explanation. Likewise, NLP-type explanations can be attractive for the layperson since it exhibits a natural feel which makes the whole process interactive and efficient~\cite{cambria2023survey}. Ultimately, the incorporation of multimodal information entails more flexibility in crafting a good explanation and is supported by the abundance of textual information available. We believe the inclusion of NLP in XAI presents an exciting opportunity to enhance our understanding of financial models and further promote better transparency and trustworthiness in today's AI models.

%%%%%%%%%%%%%%%%%%%%%%%%%%%%%%%%%%%%%%%%%%%%%%%%
% SECTION 8
%%%%%%%%%%%%%%%%%%%%%%%%%%%%%%%%%%%%%%%%%%%%%%%%
\section{Conclusion}
\label{section:conclusion}
%%%%%%%%%%%%%%%%%%%%%%%%%%%%%%%%%%%%%%%%%%%%%%%%

Overall, explainability will continue to be a critical area of focus in FinTech as companies seek to build trust and confidence with consumers and regulators alike. To conclude our work, we have provided a comprehensive review of XAI tools in the financial domain (FinXAI), highlighting the significant progress made in recent years toward developing explainable AI models for financial applications. This includes both inherently transparent models and post-hoc explainability techniques, the former of which we advocate for more improvements to be made. We provided a framework that establishes the selection of appropriate FinXAI tools as a sequential decision-making process, placing great emphasis on the audience and iterative assessment of produced explanation. The reviewed works are categorized according to their respective characteristics for ease of access by interested readers. We also examine the contributions of current FinXAI to several ethical goals, e.g., trustworthiness, fairness, informativeness, accessibility, privacy, confidence, causality, and transparency. 

Though there have been many great works done thus far, the review also reveals some limitations and challenges associated with FinXAI. This includes appropriate metrics to measure both the faithfulness and plausibility of explanations, as well as issues concerning the over-reliance on potentially misleading explanations. Future research should focus on addressing these challenges, as well as exploring new directions for FinXAI, including integrating NLP into explanation-generating techniques and a greater focus on inherently transparent models. Nevertheless, there is great potential for XAI techniques to enhance transparency, trust, and accountability in the financial domain. This underscores the importance of active research and development in this field.

\bibliographystyle{ACM-Reference-Format}
\bibliography{sample-base}


\begin{thebibliography}{139}

%%% the Web address.

\ifx \showCODEN    \undefined \def \showCODEN     #1{\unskip}     \fi
\ifx \showDOI      \undefined \def \showDOI       #1{#1}\fi
\ifx \showISBNx    \undefined \def \showISBNx     #1{\unskip}     \fi
\ifx \showISBNxiii \undefined \def \showISBNxiii  #1{\unskip}     \fi
\ifx \showISSN     \undefined \def \showISSN      #1{\unskip}     \fi
\ifx \showLCCN     \undefined \def \showLCCN      #1{\unskip}     \fi
\ifx \shownote     \undefined \def \shownote      #1{#1}          \fi
\ifx \showarticletitle \undefined \def \showarticletitle #1{#1}   \fi
\ifx \showURL      \undefined \def \showURL       {\relax}        \fi
% invisible to TeX
\providecommand\bibfield[2]{#2}
\providecommand\bibinfo[2]{#2}
\providecommand\natexlab[1]{#1}
\providecommand\showeprint[2][]{arXiv:#2}

\bibitem[mic(2021)]%
        {microsofttool}
 \bibinfo{year}{2021}\natexlab{}.
\newblock \bibinfo{booktitle}{\emph{Microsoft Responsible AI Toolbox}}.
\newblock
\urldef\tempurl%
\url{https://github.com/microsoft/responsible-ai-toolbox}
\showURL{%
Retrieved February 23, 2023 from \tempurl}


\bibitem[Achituve et~al\mbox{.}(2019)]%
        {achituve2019interpretable}
\bibfield{author}{\bibinfo{person}{Idan Achituve}, \bibinfo{person}{Sarit
  Kraus}, {and} \bibinfo{person}{Jacob Goldberger}.}
  \bibinfo{year}{2019}\natexlab{}.
\newblock \showarticletitle{Interpretable online banking fraud detection based
  on hierarchical attention mechanism}. In \bibinfo{booktitle}{\emph{2019 IEEE
  29th International Workshop on Machine Learning for Signal Processing
  (MLSP)}}. IEEE, \bibinfo{pages}{1--6}.
\newblock


\bibitem[Adams and Hagras(2020)]%
        {adams2020type}
\bibfield{author}{\bibinfo{person}{Janet Adams} {and} \bibinfo{person}{Hani
  Hagras}.} \bibinfo{year}{2020}\natexlab{}.
\newblock \showarticletitle{A type-2 fuzzy logic approach to explainable AI for
  regulatory compliance, fair customer outcomes and market stability in the
  global financial sector}. In \bibinfo{booktitle}{\emph{2020 IEEE
  International Conference on Fuzzy Systems (FUZZ-IEEE)}}. IEEE,
  \bibinfo{pages}{1--8}.
\newblock


\bibitem[Agarwal et~al\mbox{.}(2021)]%
        {agarwal2021neural}
\bibfield{author}{\bibinfo{person}{Rishabh Agarwal}, \bibinfo{person}{Levi
  Melnick}, \bibinfo{person}{Nicholas Frosst}, \bibinfo{person}{Xuezhou Zhang},
  \bibinfo{person}{Ben Lengerich}, \bibinfo{person}{Rich Caruana}, {and}
  \bibinfo{person}{Geoffrey~E Hinton}.} \bibinfo{year}{2021}\natexlab{}.
\newblock \showarticletitle{Neural additive models: Interpretable machine
  learning with neural nets}.
\newblock \bibinfo{journal}{\emph{Advances in Neural Information Processing
  Systems}}  \bibinfo{volume}{34} (\bibinfo{year}{2021}),
  \bibinfo{pages}{4699--4711}.
\newblock


\bibitem[Amparore et~al\mbox{.}(2021)]%
        {amparore2021trust}
\bibfield{author}{\bibinfo{person}{Elvio Amparore}, \bibinfo{person}{Alan
  Perotti}, {and} \bibinfo{person}{Paolo Bajardi}.}
  \bibinfo{year}{2021}\natexlab{}.
\newblock \showarticletitle{To trust or not to trust an explanation: using LEAF
  to evaluate local linear XAI methods}.
\newblock \bibinfo{journal}{\emph{PeerJ Computer Science}}  \bibinfo{volume}{7}
  (\bibinfo{year}{2021}), \bibinfo{pages}{e479}.
\newblock


\bibitem[Apley and Zhu(2016)]%
        {apley2016visualizing}
\bibfield{author}{\bibinfo{person}{Daniel~W Apley} {and}
  \bibinfo{person}{Jingyu Zhu}.} \bibinfo{year}{2016}\natexlab{}.
\newblock \showarticletitle{Visualizing the effects of predictor variables in
  black box supervised learning models}.
\newblock \bibinfo{journal}{\emph{arXiv preprint arXiv:1612.08468}}
  (\bibinfo{year}{2016}).
\newblock


\bibitem[Arrieta et~al\mbox{.}(2020)]%
        {arrieta2020explainable}
\bibfield{author}{\bibinfo{person}{Alejandro~Barredo Arrieta},
  \bibinfo{person}{Natalia D{\'\i}az-Rodr{\'\i}guez}, \bibinfo{person}{Javier
  Del~Ser}, \bibinfo{person}{Adrien Bennetot}, \bibinfo{person}{Siham Tabik},
  \bibinfo{person}{Alberto Barbado}, \bibinfo{person}{Salvador Garc{\'\i}a},
  \bibinfo{person}{Sergio Gil-L{\'o}pez}, \bibinfo{person}{Daniel Molina},
  \bibinfo{person}{Richard Benjamins}, {et~al\mbox{.}}}
  \bibinfo{year}{2020}\natexlab{}.
\newblock \showarticletitle{Explainable Artificial Intelligence (XAI):
  Concepts, taxonomies, opportunities and challenges toward responsible AI}.
\newblock \bibinfo{journal}{\emph{Information fusion}}  \bibinfo{volume}{58}
  (\bibinfo{year}{2020}), \bibinfo{pages}{82--115}.
\newblock


\bibitem[Babaei et~al\mbox{.}(2022)]%
        {babaei2022explainable}
\bibfield{author}{\bibinfo{person}{Golnoosh Babaei}, \bibinfo{person}{Paolo
  Giudici}, {and} \bibinfo{person}{Emanuela Raffinetti}.}
  \bibinfo{year}{2022}\natexlab{}.
\newblock \showarticletitle{Explainable artificial intelligence for crypto
  asset allocation}.
\newblock \bibinfo{journal}{\emph{Finance Research Letters}}
  \bibinfo{volume}{47} (\bibinfo{year}{2022}), \bibinfo{pages}{102941}.
\newblock


\bibitem[Bach et~al\mbox{.}(2015)]%
        {bach2015pixel}
\bibfield{author}{\bibinfo{person}{Sebastian Bach}, \bibinfo{person}{Alexander
  Binder}, \bibinfo{person}{Gr{\'e}goire Montavon}, \bibinfo{person}{Frederick
  Klauschen}, \bibinfo{person}{Klaus-Robert M{\"u}ller}, {and}
  \bibinfo{person}{Wojciech Samek}.} \bibinfo{year}{2015}\natexlab{}.
\newblock \showarticletitle{On pixel-wise explanations for non-linear
  classifier decisions by layer-wise relevance propagation}.
\newblock \bibinfo{journal}{\emph{PloS one}} \bibinfo{volume}{10},
  \bibinfo{number}{7} (\bibinfo{year}{2015}), \bibinfo{pages}{e0130140}.
\newblock


\bibitem[Bahrammirzaee(2010)]%
        {bahrammirzaee2010comparative}
\bibfield{author}{\bibinfo{person}{Arash Bahrammirzaee}.}
  \bibinfo{year}{2010}\natexlab{}.
\newblock \showarticletitle{A comparative survey of artificial intelligence
  applications in finance: artificial neural networks, expert system and hybrid
  intelligent systems}.
\newblock \bibinfo{journal}{\emph{Neural Computing and Applications}}
  \bibinfo{volume}{19}, \bibinfo{number}{8} (\bibinfo{year}{2010}),
  \bibinfo{pages}{1165--1195}.
\newblock


\bibitem[Bandi et~al\mbox{.}(2021)]%
        {bandi2021integrated}
\bibfield{author}{\bibinfo{person}{Harit Bandi}, \bibinfo{person}{Suyash
  Joshi}, \bibinfo{person}{Siddhant Bhagat}, {and} \bibinfo{person}{Dayanand
  Ambawade}.} \bibinfo{year}{2021}\natexlab{}.
\newblock \showarticletitle{Integrated Technical and Sentiment Analysis Tool
  for Market Index Movement Prediction, comprehensible using XAI}. In
  \bibinfo{booktitle}{\emph{2021 International Conference on Communication
  information and Computing Technology (ICCICT)}}. IEEE, \bibinfo{pages}{1--8}.
\newblock


\bibitem[Benhamou et~al\mbox{.}(2021)]%
        {benhamou2021explainable}
\bibfield{author}{\bibinfo{person}{Eric Benhamou},
  \bibinfo{person}{Jean-Jacques Ohana}, \bibinfo{person}{David Saltiel}, {and}
  \bibinfo{person}{Beatrice Guez}.} \bibinfo{year}{2021}\natexlab{}.
\newblock \showarticletitle{Explainable AI (XAI) models applied to planning in
  financial markets}.
\newblock  (\bibinfo{year}{2021}).
\newblock


\bibitem[Biecek et~al\mbox{.}(2021)]%
        {biecek2021enabling}
\bibfield{author}{\bibinfo{person}{Przemys{\l}aw Biecek},
  \bibinfo{person}{Marcin Chlebus}, \bibinfo{person}{Janusz Gajda},
  \bibinfo{person}{Alicja Gosiewska}, \bibinfo{person}{Anna Kozak},
  \bibinfo{person}{Dominik Ogonowski}, \bibinfo{person}{Jakub Sztachelski},
  {and} \bibinfo{person}{Piotr Wojewnik}.} \bibinfo{year}{2021}\natexlab{}.
\newblock \showarticletitle{Enabling machine learning algorithms for credit
  scoring--explainable artificial intelligence (XAI) methods for clear
  understanding complex predictive models}.
\newblock \bibinfo{journal}{\emph{arXiv preprint arXiv:2104.06735}}
  (\bibinfo{year}{2021}).
\newblock


\bibitem[Bracke et~al\mbox{.}(2019)]%
        {bracke2019machine}
\bibfield{author}{\bibinfo{person}{Philippe Bracke}, \bibinfo{person}{Anupam
  Datta}, \bibinfo{person}{Carsten Jung}, {and} \bibinfo{person}{Shayak Sen}.}
  \bibinfo{year}{2019}\natexlab{}.
\newblock \showarticletitle{Machine learning explainability in finance: an
  application to default risk analysis}.
\newblock  (\bibinfo{year}{2019}).
\newblock


\bibitem[Bueff et~al\mbox{.}(2022)]%
        {bueff2022machine}
\bibfield{author}{\bibinfo{person}{Andreas~C Bueff}, \bibinfo{person}{Mateusz
  Cytry{\'n}ski}, \bibinfo{person}{Raffaella Calabrese},
  \bibinfo{person}{Matthew Jones}, \bibinfo{person}{John Roberts},
  \bibinfo{person}{Jonathon Moore}, {and} \bibinfo{person}{Iain Brown}.}
  \bibinfo{year}{2022}\natexlab{}.
\newblock \showarticletitle{Machine learning interpretability for a stress
  scenario generation in credit scoring based on counterfactuals}.
\newblock \bibinfo{journal}{\emph{Expert Systems with Applications}}
  \bibinfo{volume}{202} (\bibinfo{year}{2022}), \bibinfo{pages}{117271}.
\newblock


\bibitem[Bussmann et~al\mbox{.}(2020)]%
        {bussmann2020explainable}
\bibfield{author}{\bibinfo{person}{Niklas Bussmann}, \bibinfo{person}{Paolo
  Giudici}, \bibinfo{person}{Dimitri Marinelli}, {and} \bibinfo{person}{Jochen
  Papenbrock}.} \bibinfo{year}{2020}\natexlab{}.
\newblock \showarticletitle{Explainable AI in fintech risk management}.
\newblock \bibinfo{journal}{\emph{Frontiers in Artificial Intelligence}}
  \bibinfo{volume}{3} (\bibinfo{year}{2020}), \bibinfo{pages}{26}.
\newblock


\bibitem[Bussmann et~al\mbox{.}(2021)]%
        {bussmann2021explainable}
\bibfield{author}{\bibinfo{person}{Niklas Bussmann}, \bibinfo{person}{Paolo
  Giudici}, \bibinfo{person}{Dimitri Marinelli}, {and} \bibinfo{person}{Jochen
  Papenbrock}.} \bibinfo{year}{2021}\natexlab{}.
\newblock \showarticletitle{Explainable machine learning in credit risk
  management}.
\newblock \bibinfo{journal}{\emph{Computational Economics}}
  \bibinfo{volume}{57} (\bibinfo{year}{2021}), \bibinfo{pages}{203--216}.
\newblock


\bibitem[Cambria et~al\mbox{.}(2022)]%
        {cambria2022senticnet}
\bibfield{author}{\bibinfo{person}{Erik Cambria}, \bibinfo{person}{Qian Liu},
  \bibinfo{person}{Sergio Decherchi}, \bibinfo{person}{Frank Xing}, {and}
  \bibinfo{person}{Kenneth Kwok}.} \bibinfo{year}{2022}\natexlab{}.
\newblock \showarticletitle{{SenticNet} 7: A commonsense-based neurosymbolic AI
  framework for explainable sentiment analysis}. In
  \bibinfo{booktitle}{\emph{Proceedings of the Thirteenth Language Resources
  and Evaluation Conference}}. \bibinfo{pages}{3829--3839}.
\newblock


\bibitem[Cambria et~al\mbox{.}(2023)]%
        {cambria2023survey}
\bibfield{author}{\bibinfo{person}{Erik Cambria}, \bibinfo{person}{Lorenzo
  Malandri}, \bibinfo{person}{Fabio Mercorio}, \bibinfo{person}{Mario
  Mezzanzanica}, {and} \bibinfo{person}{Navid Nobani}.}
  \bibinfo{year}{2023}\natexlab{}.
\newblock \showarticletitle{A survey on XAI and natural language explanations}.
\newblock \bibinfo{journal}{\emph{Information Processing \& Management}}
  \bibinfo{volume}{60}, \bibinfo{number}{1} (\bibinfo{year}{2023}),
  \bibinfo{pages}{103111}.
\newblock


\bibitem[Carta et~al\mbox{.}(2022)]%
        {carta2022explainable}
\bibfield{author}{\bibinfo{person}{Salvatore Carta},
  \bibinfo{person}{Alessandro~Sebastian Podda}, \bibinfo{person}{Diego
  Reforgiato~Recupero}, {and} \bibinfo{person}{Maria~Madalina Stanciu}.}
  \bibinfo{year}{2022}\natexlab{}.
\newblock \showarticletitle{Explainable AI for financial forecasting}. In
  \bibinfo{booktitle}{\emph{Machine Learning, Optimization, and Data Science:
  7th International Conference, LOD 2021, Grasmere, UK, October 4--8, 2021,
  Revised Selected Papers, Part II}}. Springer, \bibinfo{pages}{51--69}.
\newblock


\bibitem[Carta et~al\mbox{.}(2021)]%
        {carta2021explainable}
\bibfield{author}{\bibinfo{person}{Salvatore~M Carta}, \bibinfo{person}{Sergio
  Consoli}, \bibinfo{person}{Luca Piras}, \bibinfo{person}{Alessandro~Sebastian
  Podda}, {and} \bibinfo{person}{Diego~Reforgiato Recupero}.}
  \bibinfo{year}{2021}\natexlab{}.
\newblock \showarticletitle{Explainable machine learning exploiting news and
  domain-specific lexicon for stock market forecasting}.
\newblock \bibinfo{journal}{\emph{IEEE Access}}  \bibinfo{volume}{9}
  (\bibinfo{year}{2021}), \bibinfo{pages}{30193--30205}.
\newblock


\bibitem[Chen and Ye(2022)]%
        {chen2022generalized}
\bibfield{author}{\bibinfo{person}{Dangxing Chen} {and}
  \bibinfo{person}{Weicheng Ye}.} \bibinfo{year}{2022}\natexlab{}.
\newblock \showarticletitle{Generalized Gloves of Neural Additive Models:
  Pursuing transparent and accurate machine learning models in finance}.
\newblock \bibinfo{journal}{\emph{arXiv preprint arXiv:2209.10082}}
  (\bibinfo{year}{2022}).
\newblock


\bibitem[Chen and Storchan(2021)]%
        {chen2021seven}
\bibfield{author}{\bibinfo{person}{Jiahao Chen} {and} \bibinfo{person}{Victor
  Storchan}.} \bibinfo{year}{2021}\natexlab{}.
\newblock \showarticletitle{Seven challenges for harmonizing explainability
  requirements}.
\newblock \bibinfo{journal}{\emph{arXiv preprint arXiv:2108.05390}}
  (\bibinfo{year}{2021}).
\newblock


\bibitem[Chen et~al\mbox{.}(2020)]%
        {chen2020explainable}
\bibfield{author}{\bibinfo{person}{Jun-Hao Chen}, \bibinfo{person}{Samuel
  Yen-Chi Chen}, \bibinfo{person}{Yun-Cheng Tsai}, {and}
  \bibinfo{person}{Chih-Shiang Shur}.} \bibinfo{year}{2020}\natexlab{}.
\newblock \showarticletitle{Explainable deep convolutional candlestick
  learner}.
\newblock \bibinfo{journal}{\emph{arXiv preprint arXiv:2001.02767}}
  (\bibinfo{year}{2020}).
\newblock


\bibitem[Chen and Guestrin(2016)]%
        {chen2016xgboost}
\bibfield{author}{\bibinfo{person}{Tianqi Chen} {and} \bibinfo{person}{Carlos
  Guestrin}.} \bibinfo{year}{2016}\natexlab{}.
\newblock \showarticletitle{Xgboost: A scalable tree boosting system}. In
  \bibinfo{booktitle}{\emph{Proceedings of the 22nd acm sigkdd international
  conference on knowledge discovery and data mining}}.
  \bibinfo{pages}{785--794}.
\newblock


\bibitem[Cho and Shin(2023)]%
        {cho2023feature}
\bibfield{author}{\bibinfo{person}{Soo~Hyun Cho} {and}
  \bibinfo{person}{Kyung-shik Shin}.} \bibinfo{year}{2023}\natexlab{}.
\newblock \showarticletitle{Feature-Weighted Counterfactual-Based Explanation
  for Bankruptcy Prediction}.
\newblock \bibinfo{journal}{\emph{Expert Systems with Applications}}
  \bibinfo{volume}{216} (\bibinfo{year}{2023}), \bibinfo{pages}{119390}.
\newblock


\bibitem[Chung et~al\mbox{.}(2014)]%
        {chung2014empirical}
\bibfield{author}{\bibinfo{person}{Junyoung Chung}, \bibinfo{person}{Caglar
  Gulcehre}, \bibinfo{person}{KyungHyun Cho}, {and} \bibinfo{person}{Yoshua
  Bengio}.} \bibinfo{year}{2014}\natexlab{}.
\newblock \showarticletitle{Empirical evaluation of gated recurrent neural
  networks on sequence modeling}.
\newblock \bibinfo{journal}{\emph{arXiv preprint arXiv:1412.3555}}
  (\bibinfo{year}{2014}).
\newblock


\bibitem[Collaris et~al\mbox{.}(2018)]%
        {collaris2018instance}
\bibfield{author}{\bibinfo{person}{Dennis Collaris}, \bibinfo{person}{Leo~M
  Vink}, {and} \bibinfo{person}{Jarke~J van Wijk}.}
  \bibinfo{year}{2018}\natexlab{}.
\newblock \showarticletitle{Instance-level explanations for fraud detection: A
  case study}.
\newblock \bibinfo{journal}{\emph{arXiv preprint arXiv:1806.07129}}
  (\bibinfo{year}{2018}).
\newblock


\bibitem[Cong et~al\mbox{.}(2021)]%
        {cong2021alphaportfolio}
\bibfield{author}{\bibinfo{person}{Lin~William Cong}, \bibinfo{person}{Ke
  Tang}, \bibinfo{person}{Jingyuan Wang}, {and} \bibinfo{person}{Yang Zhang}.}
  \bibinfo{year}{2021}\natexlab{}.
\newblock \showarticletitle{AlphaPortfolio: Direct construction through deep
  reinforcement learning and interpretable AI}.
\newblock \bibinfo{journal}{\emph{Available at SSRN}}
  \bibinfo{volume}{3554486} (\bibinfo{year}{2021}).
\newblock


\bibitem[Crosato et~al\mbox{.}(2021)]%
        {crosato2021look}
\bibfield{author}{\bibinfo{person}{Lisa Crosato}, \bibinfo{person}{Caterina
  Liberati}, {and} \bibinfo{person}{Marco Repetto}.}
  \bibinfo{year}{2021}\natexlab{}.
\newblock \showarticletitle{Look Who's Talking: Interpretable Machine Learning
  for Assessing Italian SMEs Credit Default}.
\newblock \bibinfo{journal}{\emph{arXiv preprint arXiv:2108.13914}}
  (\bibinfo{year}{2021}).
\newblock


\bibitem[Danilevsky et~al\mbox{.}(2020)]%
        {danilevsky2020survey}
\bibfield{author}{\bibinfo{person}{Marina Danilevsky}, \bibinfo{person}{Kun
  Qian}, \bibinfo{person}{Ranit Aharonov}, \bibinfo{person}{Yannis Katsis},
  \bibinfo{person}{Ban Kawas}, {and} \bibinfo{person}{Prithviraj Sen}.}
  \bibinfo{year}{2020}\natexlab{}.
\newblock \showarticletitle{A survey of the state of explainable AI for natural
  language processing}.
\newblock \bibinfo{journal}{\emph{arXiv preprint arXiv:2010.00711}}
  (\bibinfo{year}{2020}).
\newblock


\bibitem[Datta et~al\mbox{.}(2016)]%
        {datta2016algorithmic}
\bibfield{author}{\bibinfo{person}{Anupam Datta}, \bibinfo{person}{Shayak Sen},
  {and} \bibinfo{person}{Yair Zick}.} \bibinfo{year}{2016}\natexlab{}.
\newblock \showarticletitle{Algorithmic transparency via quantitative input
  influence: Theory and experiments with learning systems}. In
  \bibinfo{booktitle}{\emph{2016 IEEE symposium on security and privacy (SP)}}.
  IEEE, \bibinfo{pages}{598--617}.
\newblock


\bibitem[Davis et~al\mbox{.}(2022)]%
        {davis2022explainable}
\bibfield{author}{\bibinfo{person}{Randall Davis}, \bibinfo{person}{Andrew~W
  Lo}, \bibinfo{person}{Sudhanshu Mishra}, \bibinfo{person}{Arash Nourian},
  \bibinfo{person}{Manish Singh}, \bibinfo{person}{Nicholas Wu}, {and}
  \bibinfo{person}{Ruixun Zhang}.} \bibinfo{year}{2022}\natexlab{}.
\newblock \showarticletitle{Explainable machine learning models of consumer
  credit risk}.
\newblock \bibinfo{journal}{\emph{Available at SSRN}} (\bibinfo{year}{2022}).
\newblock


\bibitem[De~Bruin et~al\mbox{.}(2009)]%
        {de2009ad}
\bibfield{author}{\bibinfo{person}{Kelly~C De~Bruin}, \bibinfo{person}{Rob~B
  Dellink}, {and} \bibinfo{person}{Richard~SJ Tol}.}
  \bibinfo{year}{2009}\natexlab{}.
\newblock \showarticletitle{AD-DICE: an implementation of adaptation in the
  DICE model}.
\newblock \bibinfo{journal}{\emph{Climatic Change}}  \bibinfo{volume}{95}
  (\bibinfo{year}{2009}), \bibinfo{pages}{63--81}.
\newblock


\bibitem[Dellinger(2019)]%
        {insurancebreach}
\bibfield{author}{\bibinfo{person}{AJ Dellinger}.}
  \bibinfo{year}{2019}\natexlab{}.
\newblock \bibinfo{booktitle}{\emph{Understanding The First American Financial
  Data Leak: How Did It Happen And What Does It Mean?}}
\newblock
\urldef\tempurl%
\url{https://www.forbes.com/sites/ajdellinger/2019/05/26/understanding-the-first-american-financial-data-leak-how-did-it-happen-and-what-does-it-mean/?sh=7716df86567f}
\showURL{%
Retrieved February 22, 2023 from \tempurl}


\bibitem[Demajo et~al\mbox{.}(2020)]%
        {demajo2020explainable}
\bibfield{author}{\bibinfo{person}{Lara~Marie Demajo}, \bibinfo{person}{Vince
  Vella}, {and} \bibinfo{person}{Alexiei Dingli}.}
  \bibinfo{year}{2020}\natexlab{}.
\newblock \showarticletitle{Explainable AI for Interpretable Credit Scoring}.
  In \bibinfo{booktitle}{\emph{CS \& IT Conference Proceedings}},
  Vol.~\bibinfo{volume}{10}. CS \& IT Conference Proceedings.
\newblock


\bibitem[Deng et~al\mbox{.}(2019)]%
        {deng2019knowledge}
\bibfield{author}{\bibinfo{person}{Shumin Deng}, \bibinfo{person}{Ningyu
  Zhang}, \bibinfo{person}{Wen Zhang}, \bibinfo{person}{Jiaoyan Chen},
  \bibinfo{person}{Jeff~Z Pan}, {and} \bibinfo{person}{Huajun Chen}.}
  \bibinfo{year}{2019}\natexlab{}.
\newblock \showarticletitle{Knowledge-driven stock trend prediction and
  explanation via temporal convolutional network}. In
  \bibinfo{booktitle}{\emph{Companion Proceedings of The 2019 World Wide Web
  Conference}}. \bibinfo{pages}{678--685}.
\newblock


\bibitem[Dikmen and Burns(2022)]%
        {dikmen2022effects}
\bibfield{author}{\bibinfo{person}{Murat Dikmen} {and}
  \bibinfo{person}{Catherine Burns}.} \bibinfo{year}{2022}\natexlab{}.
\newblock \showarticletitle{The effects of domain knowledge on trust in
  explainable AI and task performance: A case of peer-to-peer lending}.
\newblock \bibinfo{journal}{\emph{International Journal of Human-Computer
  Studies}}  \bibinfo{volume}{162} (\bibinfo{year}{2022}),
  \bibinfo{pages}{102792}.
\newblock


\bibitem[Dumitrescu et~al\mbox{.}(2022)]%
        {dumitrescu2022machine}
\bibfield{author}{\bibinfo{person}{Elena Dumitrescu}, \bibinfo{person}{Sullivan
  Hu{\'e}}, \bibinfo{person}{Christophe Hurlin}, {and} \bibinfo{person}{Sessi
  Tokpavi}.} \bibinfo{year}{2022}\natexlab{}.
\newblock \showarticletitle{Machine learning for credit scoring: Improving
  logistic regression with non-linear decision-tree effects}.
\newblock \bibinfo{journal}{\emph{European Journal of Operational Research}}
  \bibinfo{volume}{297}, \bibinfo{number}{3} (\bibinfo{year}{2022}),
  \bibinfo{pages}{1178--1192}.
\newblock


\bibitem[Farzad(2019)]%
        {farzad2019determinants}
\bibfield{author}{\bibinfo{person}{Taghi Farzad}.}
  \bibinfo{year}{2019}\natexlab{}.
\newblock \bibinfo{title}{Determinants of Mortgage Loan Delinquency:
  Application of Interpretable Machine Learning}.
\newblock
\newblock


\bibitem[Fior et~al\mbox{.}(2022)]%
        {fior2022leveraging}
\bibfield{author}{\bibinfo{person}{Jacopo Fior}, \bibinfo{person}{Luca
  Cagliero}, {and} \bibinfo{person}{Paolo Garza}.}
  \bibinfo{year}{2022}\natexlab{}.
\newblock \showarticletitle{Leveraging Explainable AI to Support Cryptocurrency
  Investors}.
\newblock \bibinfo{journal}{\emph{Future Internet}} \bibinfo{volume}{14},
  \bibinfo{number}{9} (\bibinfo{year}{2022}), \bibinfo{pages}{251}.
\newblock


\bibitem[Fritz-Morgenthal et~al\mbox{.}(2022)]%
        {fritz2022financial}
\bibfield{author}{\bibinfo{person}{Sebastian Fritz-Morgenthal},
  \bibinfo{person}{Bernhard Hein}, {and} \bibinfo{person}{Jochen Papenbrock}.}
  \bibinfo{year}{2022}\natexlab{}.
\newblock \showarticletitle{Financial risk management and explainable,
  trustworthy, responsible AI}.
\newblock \bibinfo{journal}{\emph{Frontiers in Artificial Intelligence}}
  \bibinfo{volume}{5} (\bibinfo{year}{2022}), \bibinfo{pages}{5}.
\newblock


\bibitem[Ghosh and Sanyal(2021)]%
        {ghosh2021introspecting}
\bibfield{author}{\bibinfo{person}{Indranil Ghosh} {and}
  \bibinfo{person}{Manas~K Sanyal}.} \bibinfo{year}{2021}\natexlab{}.
\newblock \showarticletitle{Introspecting predictability of market fear in
  Indian context during COVID-19 pandemic: An integrated approach of applied
  predictive modelling and explainable AI}.
\newblock \bibinfo{journal}{\emph{International Journal of Information
  Management Data Insights}} \bibinfo{volume}{1}, \bibinfo{number}{2}
  (\bibinfo{year}{2021}), \bibinfo{pages}{100039}.
\newblock


\bibitem[Gite et~al\mbox{.}(2021)]%
        {gite2021explainable}
\bibfield{author}{\bibinfo{person}{Shilpa Gite}, \bibinfo{person}{Hrituja
  Khatavkar}, \bibinfo{person}{Ketan Kotecha}, \bibinfo{person}{Shilpi
  Srivastava}, \bibinfo{person}{Priyam Maheshwari}, {and}
  \bibinfo{person}{Neerav Pandey}.} \bibinfo{year}{2021}\natexlab{}.
\newblock \showarticletitle{Explainable stock prices prediction from financial
  news articles using sentiment analysis}.
\newblock \bibinfo{journal}{\emph{PeerJ Computer Science}}  \bibinfo{volume}{7}
  (\bibinfo{year}{2021}), \bibinfo{pages}{e340}.
\newblock


\bibitem[Goodman and Flaxman(2017)]%
        {goodman2017european}
\bibfield{author}{\bibinfo{person}{Bryce Goodman} {and} \bibinfo{person}{Seth
  Flaxman}.} \bibinfo{year}{2017}\natexlab{}.
\newblock \showarticletitle{European Union regulations on algorithmic
  decision-making and a ``right to explanation''}.
\newblock \bibinfo{journal}{\emph{AI magazine}} \bibinfo{volume}{38},
  \bibinfo{number}{3} (\bibinfo{year}{2017}), \bibinfo{pages}{50--57}.
\newblock


\bibitem[Gramegna and Giudici(2020)]%
        {gramegna2020buy}
\bibfield{author}{\bibinfo{person}{Alex Gramegna} {and} \bibinfo{person}{Paolo
  Giudici}.} \bibinfo{year}{2020}\natexlab{}.
\newblock \showarticletitle{Why to buy insurance? An explainable artificial
  intelligence approach}.
\newblock \bibinfo{journal}{\emph{Risks}} \bibinfo{volume}{8},
  \bibinfo{number}{4} (\bibinfo{year}{2020}), \bibinfo{pages}{137}.
\newblock


\bibitem[Gramespacher and Posth(2021)]%
        {gramespacher2021employing}
\bibfield{author}{\bibinfo{person}{Thomas Gramespacher} {and}
  \bibinfo{person}{Jan-Alexander Posth}.} \bibinfo{year}{2021}\natexlab{}.
\newblock \showarticletitle{Employing explainable AI to optimize the return
  target function of a loan portfolio}.
\newblock \bibinfo{journal}{\emph{Frontiers in Artificial Intelligence}}
  \bibinfo{volume}{4} (\bibinfo{year}{2021}), \bibinfo{pages}{693022}.
\newblock


\bibitem[Grath et~al\mbox{.}(2018)]%
        {grath2018interpretable}
\bibfield{author}{\bibinfo{person}{Rory~Mc Grath}, \bibinfo{person}{Luca
  Costabello}, \bibinfo{person}{Chan~Le Van}, \bibinfo{person}{Paul Sweeney},
  \bibinfo{person}{Farbod Kamiab}, \bibinfo{person}{Zhao Shen}, {and}
  \bibinfo{person}{Freddy Lecue}.} \bibinfo{year}{2018}\natexlab{}.
\newblock \showarticletitle{Interpretable credit application predictions with
  counterfactual explanations}.
\newblock \bibinfo{journal}{\emph{arXiv preprint arXiv:1811.05245}}
  (\bibinfo{year}{2018}).
\newblock


\bibitem[Guidotti et~al\mbox{.}(2018)]%
        {guidotti2018survey}
\bibfield{author}{\bibinfo{person}{Riccardo Guidotti}, \bibinfo{person}{Anna
  Monreale}, \bibinfo{person}{Salvatore Ruggieri}, \bibinfo{person}{Franco
  Turini}, \bibinfo{person}{Fosca Giannotti}, {and} \bibinfo{person}{Dino
  Pedreschi}.} \bibinfo{year}{2018}\natexlab{}.
\newblock \showarticletitle{A survey of methods for explaining black box
  models}.
\newblock \bibinfo{journal}{\emph{ACM computing surveys (CSUR)}}
  \bibinfo{volume}{51}, \bibinfo{number}{5} (\bibinfo{year}{2018}),
  \bibinfo{pages}{1--42}.
\newblock


\bibitem[Gurumoorthy et~al\mbox{.}(2019)]%
        {gurumoorthy2019efficient}
\bibfield{author}{\bibinfo{person}{Karthik~S Gurumoorthy},
  \bibinfo{person}{Amit Dhurandhar}, \bibinfo{person}{Guillermo Cecchi}, {and}
  \bibinfo{person}{Charu Aggarwal}.} \bibinfo{year}{2019}\natexlab{}.
\newblock \showarticletitle{Efficient data representation by selecting
  prototypes with importance weights}. In \bibinfo{booktitle}{\emph{2019 IEEE
  International Conference on Data Mining (ICDM)}}. IEEE,
  \bibinfo{pages}{260--269}.
\newblock


\bibitem[Han et~al\mbox{.}(2022)]%
        {han2022hierarchical}
\bibfield{author}{\bibinfo{person}{Sooji Han}, \bibinfo{person}{Rui Mao}, {and}
  \bibinfo{person}{Erik Cambria}.} \bibinfo{year}{2022}\natexlab{}.
\newblock \showarticletitle{Hierarchical Attention Network for Explainable
  Depression Detection on {Twitter} Aided by Metaphor Concept Mappings}. In
  \bibinfo{booktitle}{\emph{Proceedings of the 29th International Conference on
  Computational Linguistics (COLING)}}. \bibinfo{publisher}{International
  Committee on Computational Linguistics}, \bibinfo{address}{Gyeongju, Republic
  of Korea}, \bibinfo{pages}{94--104}.
\newblock


\bibitem[He et~al\mbox{.}(2022a)]%
        {he2022jcbie}
\bibfield{author}{\bibinfo{person}{Kai He}, \bibinfo{person}{Rui Mao},
  \bibinfo{person}{Tieliang Gong}, \bibinfo{person}{Erik Cambria}, {and}
  \bibinfo{person}{Chen Li}.} \bibinfo{year}{2022}\natexlab{a}.
\newblock \showarticletitle{{JCBIE}: A Joint Continual Learning Neural Network
  for Biomedical Information Extraction}.
\newblock \bibinfo{journal}{\emph{BMC Bioinformatics}} \bibinfo{volume}{23},
  \bibinfo{number}{549} (\bibinfo{year}{2022}).
\newblock
\urldef\tempurl%
\url{https://doi.org/10.1186/s12859-022-05096-w}
\showDOI{\tempurl}


\bibitem[He et~al\mbox{.}(2022b)]%
        {he2022metabased}
\bibfield{author}{\bibinfo{person}{Kai He}, \bibinfo{person}{Rui Mao},
  \bibinfo{person}{Tieliang Gong}, \bibinfo{person}{Chen Li}, {and}
  \bibinfo{person}{Erik Cambria}.} \bibinfo{year}{2022}\natexlab{b}.
\newblock \showarticletitle{Meta-based Self-training and Re-weighting for
  Aspect-based Sentiment Analysis}.
\newblock \bibinfo{journal}{\emph{IEEE Transactions on Affective Computing}}
  (\bibinfo{year}{2022}).
\newblock
\urldef\tempurl%
\url{https://doi.org/10.1109/TAFFC.2022.3202831}
\showDOI{\tempurl}


\bibitem[HLEG(2019)]%
        {euhighlevel}
\bibfield{author}{\bibinfo{person}{AI HLEG}.} \bibinfo{year}{2019}\natexlab{}.
\newblock \bibinfo{booktitle}{\emph{Ethics guidelines for trustworthy AI}}.
\newblock
\urldef\tempurl%
\url{https://digital-strategy.ec.europa.eu/en/library/ethics-guidelines-trustworthy-ai}
\showURL{%
Retrieved February 8, 2023 from \tempurl}


\bibitem[Hoffman et~al\mbox{.}(2018)]%
        {hoffman2018metrics}
\bibfield{author}{\bibinfo{person}{Robert~R Hoffman}, \bibinfo{person}{Shane~T
  Mueller}, \bibinfo{person}{Gary Klein}, {and} \bibinfo{person}{Jordan
  Litman}.} \bibinfo{year}{2018}\natexlab{}.
\newblock \showarticletitle{Metrics for explainable AI: Challenges and
  prospects}.
\newblock \bibinfo{journal}{\emph{arXiv preprint arXiv:1812.04608}}
  (\bibinfo{year}{2018}).
\newblock


\bibitem[Hohman et~al\mbox{.}(2019)]%
        {hohman2019gamut}
\bibfield{author}{\bibinfo{person}{Fred Hohman}, \bibinfo{person}{Andrew Head},
  \bibinfo{person}{Rich Caruana}, \bibinfo{person}{Robert DeLine}, {and}
  \bibinfo{person}{Steven~M Drucker}.} \bibinfo{year}{2019}\natexlab{}.
\newblock \showarticletitle{Gamut: A design probe to understand how data
  scientists understand machine learning models}. In
  \bibinfo{booktitle}{\emph{Proceedings of the 2019 CHI conference on human
  factors in computing systems}}. \bibinfo{pages}{1--13}.
\newblock


\bibitem[Islam et~al\mbox{.}(2019)]%
        {islam2019infusing}
\bibfield{author}{\bibinfo{person}{Sheikh~Rabiul Islam},
  \bibinfo{person}{William Eberle}, \bibinfo{person}{Sid Bundy}, {and}
  \bibinfo{person}{Sheikh~Khaled Ghafoor}.} \bibinfo{year}{2019}\natexlab{}.
\newblock \showarticletitle{Infusing domain knowledge in AI-based `` black
  box'' models for better explainability with application in bankruptcy
  prediction}.
\newblock \bibinfo{journal}{\emph{arXiv preprint arXiv:1905.11474}}
  (\bibinfo{year}{2019}).
\newblock


\bibitem[Ito et~al\mbox{.}(2020)]%
        {ito2020ginn}
\bibfield{author}{\bibinfo{person}{Tomoki Ito}, \bibinfo{person}{Hiroki
  Sakaji}, \bibinfo{person}{Kiyoshi Izumi}, \bibinfo{person}{Kota Tsubouchi},
  {and} \bibinfo{person}{Tatsuo Yamashita}.} \bibinfo{year}{2020}\natexlab{}.
\newblock \showarticletitle{GINN: gradient interpretable neural networks for
  visualizing financial texts}.
\newblock \bibinfo{journal}{\emph{International Journal of Data Science and
  Analytics}}  \bibinfo{volume}{9} (\bibinfo{year}{2020}),
  \bibinfo{pages}{431--445}.
\newblock


\bibitem[Jacovi and Goldberg(2020)]%
        {jacovi2020towards}
\bibfield{author}{\bibinfo{person}{Alon Jacovi} {and} \bibinfo{person}{Yoav
  Goldberg}.} \bibinfo{year}{2020}\natexlab{}.
\newblock \showarticletitle{Towards faithfully interpretable NLP systems: How
  should we define and evaluate faithfulness?}
\newblock \bibinfo{journal}{\emph{arXiv preprint arXiv:2004.03685}}
  (\bibinfo{year}{2020}).
\newblock


\bibitem[Jin et~al\mbox{.}(2019)]%
        {jin2019towards}
\bibfield{author}{\bibinfo{person}{Xisen Jin}, \bibinfo{person}{Zhongyu Wei},
  \bibinfo{person}{Junyi Du}, \bibinfo{person}{Xiangyang Xue}, {and}
  \bibinfo{person}{Xiang Ren}.} \bibinfo{year}{2019}\natexlab{}.
\newblock \showarticletitle{Towards hierarchical importance attribution:
  Explaining compositional semantics for neural sequence models}.
\newblock \bibinfo{journal}{\emph{arXiv preprint arXiv:1911.06194}}
  (\bibinfo{year}{2019}).
\newblock


\bibitem[Kaur et~al\mbox{.}(2020)]%
        {kaur2020interpreting}
\bibfield{author}{\bibinfo{person}{Harmanpreet Kaur}, \bibinfo{person}{Harsha
  Nori}, \bibinfo{person}{Samuel Jenkins}, \bibinfo{person}{Rich Caruana},
  \bibinfo{person}{Hanna Wallach}, {and} \bibinfo{person}{Jennifer
  Wortman~Vaughan}.} \bibinfo{year}{2020}\natexlab{}.
\newblock \showarticletitle{Interpreting interpretability: understanding data
  scientists' use of interpretability tools for machine learning}. In
  \bibinfo{booktitle}{\emph{Proceedings of the 2020 CHI conference on human
  factors in computing systems}}. \bibinfo{pages}{1--14}.
\newblock


\bibitem[Kolanovic and Krishnamachari(2017)]%
        {markok}
\bibfield{author}{\bibinfo{person}{Marko Kolanovic} {and}
  \bibinfo{person}{Rajesh~T. Krishnamachari}.} \bibinfo{year}{2017}\natexlab{}.
\newblock \bibinfo{title}{Big Data and AI Strategies, Machine Learning and
  Alternative Data Approach to Investing}.
\newblock
\newblock


\bibitem[Kuiper et~al\mbox{.}(2022)]%
        {kuiper2022exploring}
\bibfield{author}{\bibinfo{person}{Ouren Kuiper}, \bibinfo{person}{Martin
  van~den Berg}, \bibinfo{person}{Joost van~der Burgt}, {and}
  \bibinfo{person}{Stefan Leijnen}.} \bibinfo{year}{2022}\natexlab{}.
\newblock \showarticletitle{Exploring explainable ai in the financial sector:
  Perspectives of banks and supervisory authorities}. In
  \bibinfo{booktitle}{\emph{Artificial Intelligence and Machine Learning: 33rd
  Benelux Conference on Artificial Intelligence, BNAIC/Benelearn 2021,
  Esch-sur-Alzette, Luxembourg, November 10--12, 2021, Revised Selected Papers
  33}}. Springer, \bibinfo{pages}{105--119}.
\newblock


\bibitem[Kumar et~al\mbox{.}(2017)]%
        {kumar2017opening}
\bibfield{author}{\bibinfo{person}{Devinder Kumar}, \bibinfo{person}{Graham~W
  Taylor}, {and} \bibinfo{person}{Alexander Wong}.}
  \bibinfo{year}{2017}\natexlab{}.
\newblock \showarticletitle{Opening the black box of financial ai with
  clear-trade: A class-enhanced attentive response approach for explaining and
  visualizing deep learning-driven stock market prediction}.
\newblock \bibinfo{journal}{\emph{arXiv preprint arXiv:1709.01574}}
  (\bibinfo{year}{2017}).
\newblock


\bibitem[Kumar et~al\mbox{.}(2022)]%
        {kumar2022explainable}
\bibfield{author}{\bibinfo{person}{Satyam Kumar}, \bibinfo{person}{Mendhikar
  Vishal}, {and} \bibinfo{person}{Vadlamani Ravi}.}
  \bibinfo{year}{2022}\natexlab{}.
\newblock \showarticletitle{Explainable Reinforcement Learning on Financial
  Stock Trading using SHAP}.
\newblock \bibinfo{journal}{\emph{arXiv preprint arXiv:2208.08790}}
  (\bibinfo{year}{2022}).
\newblock


\bibitem[Lachuer and Jabeur(2022)]%
        {lachuer2022explainable}
\bibfield{author}{\bibinfo{person}{Julien Lachuer} {and}
  \bibinfo{person}{Sami~Ben Jabeur}.} \bibinfo{year}{2022}\natexlab{}.
\newblock \showarticletitle{Explainable artificial intelligence modeling for
  corporate social responsibility and financial performance}.
\newblock \bibinfo{journal}{\emph{Journal of Asset Management}}
  \bibinfo{volume}{23}, \bibinfo{number}{7} (\bibinfo{year}{2022}),
  \bibinfo{pages}{619--630}.
\newblock


\bibitem[Levine(2014)]%
        {levine2014truth}
\bibfield{author}{\bibinfo{person}{Timothy~R Levine}.}
  \bibinfo{year}{2014}\natexlab{}.
\newblock \showarticletitle{Truth-default theory (TDT) a theory of human
  deception and deception detection}.
\newblock \bibinfo{journal}{\emph{Journal of Language and Social Psychology}}
  \bibinfo{volume}{33}, \bibinfo{number}{4} (\bibinfo{year}{2014}),
  \bibinfo{pages}{378--392}.
\newblock


\bibitem[Liang et~al\mbox{.}(2022)]%
        {liang2022aspect}
\bibfield{author}{\bibinfo{person}{Bin Liang}, \bibinfo{person}{Hang Su},
  \bibinfo{person}{Lin Gui}, \bibinfo{person}{Erik Cambria}, {and}
  \bibinfo{person}{Ruifeng Xu}.} \bibinfo{year}{2022}\natexlab{}.
\newblock \showarticletitle{Aspect-based sentiment analysis via affective
  knowledge enhanced graph convolutional networks}.
\newblock \bibinfo{journal}{\emph{Knowledge-Based Systems}}
  \bibinfo{volume}{235} (\bibinfo{year}{2022}), \bibinfo{pages}{107643}.
\newblock


\bibitem[Lin et~al\mbox{.}(2021)]%
        {lin2021xrr}
\bibfield{author}{\bibinfo{person}{Ting-Wei Lin}, \bibinfo{person}{Ruei-Yao
  Sun}, \bibinfo{person}{Hsuan-Ling Chang}, \bibinfo{person}{Chuan-Ju Wang},
  {and} \bibinfo{person}{Ming-Feng Tsai}.} \bibinfo{year}{2021}\natexlab{}.
\newblock \showarticletitle{XRR: Explainable Risk Ranking for Financial
  Reports}. In \bibinfo{booktitle}{\emph{Machine Learning and Knowledge
  Discovery in Databases. Applied Data Science Track: European Conference, ECML
  PKDD 2021, Bilbao, Spain, September 13--17, 2021, Proceedings, Part IV 21}}.
  Springer, \bibinfo{pages}{253--268}.
\newblock


\bibitem[Lipton(1990)]%
        {lipton1990contrastive}
\bibfield{author}{\bibinfo{person}{Peter Lipton}.}
  \bibinfo{year}{1990}\natexlab{}.
\newblock \showarticletitle{Contrastive explanation}.
\newblock \bibinfo{journal}{\emph{Royal Institute of Philosophy Supplements}}
  \bibinfo{volume}{27} (\bibinfo{year}{1990}), \bibinfo{pages}{247--266}.
\newblock


\bibitem[Lipton(2018)]%
        {lipton2018mythos}
\bibfield{author}{\bibinfo{person}{Zachary~C Lipton}.}
  \bibinfo{year}{2018}\natexlab{}.
\newblock \showarticletitle{The mythos of model interpretability: In machine
  learning, the concept of interpretability is both important and slippery.}
\newblock \bibinfo{journal}{\emph{Queue}} \bibinfo{volume}{16},
  \bibinfo{number}{3} (\bibinfo{year}{2018}), \bibinfo{pages}{31--57}.
\newblock


\bibitem[Liu et~al\mbox{.}(2020)]%
        {liu2020predicting}
\bibfield{author}{\bibinfo{person}{Rong Liu}, \bibinfo{person}{Feng Mai},
  \bibinfo{person}{Zhe Shan}, {and} \bibinfo{person}{Ying Wu}.}
  \bibinfo{year}{2020}\natexlab{}.
\newblock \showarticletitle{Predicting shareholder litigation on insider
  trading from financial text: An interpretable deep learning approach}.
\newblock \bibinfo{journal}{\emph{Information \& Management}}
  \bibinfo{volume}{57}, \bibinfo{number}{8} (\bibinfo{year}{2020}),
  \bibinfo{pages}{103387}.
\newblock


\bibitem[Lundberg et~al\mbox{.}(2018)]%
        {lundberg2018consistent}
\bibfield{author}{\bibinfo{person}{Scott~M Lundberg},
  \bibinfo{person}{Gabriel~G Erion}, {and} \bibinfo{person}{Su-In Lee}.}
  \bibinfo{year}{2018}\natexlab{}.
\newblock \showarticletitle{Consistent individualized feature attribution for
  tree ensembles}.
\newblock \bibinfo{journal}{\emph{arXiv preprint arXiv:1802.03888}}
  (\bibinfo{year}{2018}).
\newblock


\bibitem[Lundberg and Lee(2017)]%
        {lundberg2017unified}
\bibfield{author}{\bibinfo{person}{Scott~M Lundberg} {and}
  \bibinfo{person}{Su-In Lee}.} \bibinfo{year}{2017}\natexlab{}.
\newblock \showarticletitle{A unified approach to interpreting model
  predictions}.
\newblock \bibinfo{journal}{\emph{Advances in neural information processing
  systems}}  \bibinfo{volume}{30} (\bibinfo{year}{2017}).
\newblock


\bibitem[Luo et~al\mbox{.}(2018)]%
        {luo2018beyond}
\bibfield{author}{\bibinfo{person}{Ling Luo}, \bibinfo{person}{Xiang Ao},
  \bibinfo{person}{Feiyang Pan}, \bibinfo{person}{Jin Wang},
  \bibinfo{person}{Tong Zhao}, \bibinfo{person}{Ningzi Yu}, {and}
  \bibinfo{person}{Qing He}.} \bibinfo{year}{2018}\natexlab{}.
\newblock \showarticletitle{Beyond Polarity: Interpretable Financial Sentiment
  Analysis with Hierarchical Query-driven Attention}. In
  \bibinfo{booktitle}{\emph{IJCAI}}. \bibinfo{pages}{4244--4250}.
\newblock


\bibitem[Ma et~al\mbox{.}(2023)]%
        {ma2023multisource}
\bibfield{author}{\bibinfo{person}{Yu Ma}, \bibinfo{person}{Rui Mao},
  \bibinfo{person}{Qika Lin}, \bibinfo{person}{Peng Wu}, {and}
  \bibinfo{person}{Erik Cambria}.} \bibinfo{year}{2023}\natexlab{}.
\newblock \showarticletitle{Multi-source Aggregated Classification for Stock
  Price Movement Prediction}.
\newblock \bibinfo{journal}{\emph{Information Fusion}}  \bibinfo{volume}{91}
  (\bibinfo{year}{2023}), \bibinfo{pages}{515--528}.
\newblock
\urldef\tempurl%
\url{https://doi.org/10.1016/j.inffus.2022.10.025}
\showDOI{\tempurl}


\bibitem[Mao and Li(2021)]%
        {mao2021bridging}
\bibfield{author}{\bibinfo{person}{Rui Mao} {and} \bibinfo{person}{Xiao Li}.}
  \bibinfo{year}{2021}\natexlab{}.
\newblock \showarticletitle{Bridging Towers of Multi-task Learning with a
  Gating Mechanism for Aspect-based Sentiment Analysis and Sequential Metaphor
  Identification}.
\newblock \bibinfo{journal}{\emph{Proceedings of the AAAI Conference on
  Artificial Intelligence}} \bibinfo{volume}{35}, \bibinfo{number}{15}
  (\bibinfo{year}{2021}), \bibinfo{pages}{13534--13542}.
\newblock
\urldef\tempurl%
\url{https://doi.org/10.1609/aaai.v35i15.17596}
\showDOI{\tempurl}


\bibitem[Mao et~al\mbox{.}(2022)]%
        {mao2022metapro}
\bibfield{author}{\bibinfo{person}{Rui Mao}, \bibinfo{person}{Xiao Li},
  \bibinfo{person}{Mengshi Ge}, {and} \bibinfo{person}{Erik Cambria}.}
  \bibinfo{year}{2022}\natexlab{}.
\newblock \showarticletitle{{MetaPro}: A computational metaphor processing
  model for text pre-processing}.
\newblock \bibinfo{journal}{\emph{Information Fusion}}  \bibinfo{volume}{86-87}
  (\bibinfo{year}{2022}), \bibinfo{pages}{30--43}.
\newblock
\showISSN{1566-2535}
\urldef\tempurl%
\url{https://doi.org/10.1016/j.inffus.2022.06.002}
\showDOI{\tempurl}


\bibitem[Maree et~al\mbox{.}(2020)]%
        {maree2020towards}
\bibfield{author}{\bibinfo{person}{Charl Maree}, \bibinfo{person}{Jan~Erik
  Modal}, {and} \bibinfo{person}{Christian~W Omlin}.}
  \bibinfo{year}{2020}\natexlab{}.
\newblock \showarticletitle{Towards responsible AI for financial transactions}.
  In \bibinfo{booktitle}{\emph{2020 IEEE Symposium Series on Computational
  Intelligence (SSCI)}}. IEEE, \bibinfo{pages}{16--21}.
\newblock


\bibitem[Maree and Omlin(2022a)]%
        {maree2022can}
\bibfield{author}{\bibinfo{person}{Charl Maree} {and}
  \bibinfo{person}{Christian~W Omlin}.} \bibinfo{year}{2022}\natexlab{a}.
\newblock \showarticletitle{Can interpretable reinforcement learning manage
  prosperity your way?}
\newblock \bibinfo{journal}{\emph{AI}} \bibinfo{volume}{3}, \bibinfo{number}{2}
  (\bibinfo{year}{2022}), \bibinfo{pages}{526--537}.
\newblock


\bibitem[Maree and Omlin(2022b)]%
        {maree2022understanding}
\bibfield{author}{\bibinfo{person}{Charl Maree} {and}
  \bibinfo{person}{Christian~W Omlin}.} \bibinfo{year}{2022}\natexlab{b}.
\newblock \showarticletitle{Understanding Spending Behavior: Recurrent Neural
  Network Explanation and Interpretation}. In \bibinfo{booktitle}{\emph{2022
  IEEE Symposium on Computational Intelligence for Financial Engineering and
  Economics (CIFEr)}}. IEEE, \bibinfo{pages}{1--7}.
\newblock


\bibitem[Markowitz(1952)]%
        {10.2307/2975974}
\bibfield{author}{\bibinfo{person}{Harry Markowitz}.}
  \bibinfo{year}{1952}\natexlab{}.
\newblock \showarticletitle{Portfolio Selection}.
\newblock \bibinfo{journal}{\emph{The Journal of Finance}} \bibinfo{volume}{7},
  \bibinfo{number}{1} (\bibinfo{year}{1952}), \bibinfo{pages}{77--91}.
\newblock
\showISSN{00221082, 15406261}
\urldef\tempurl%
\url{http://www.jstor.org/stable/2975974}
\showURL{%
\tempurl}


\bibitem[Mellon(2021)]%
        {xaimodelling}
\bibfield{author}{\bibinfo{person}{BNY Mellon}.}
  \bibinfo{year}{2021}\natexlab{}.
\newblock \bibinfo{booktitle}{\emph{Why Every Financial Institution Should
  Consider Explainable AI}}.
\newblock
\urldef\tempurl%
\url{https://www.bnymellon.com/us/en/insights/all-insights/why-every-financial-institution-should-consider-explainable-ai.html}
\showURL{%
Retrieved February 8, 2023 from \tempurl}


\bibitem[Miller(2019)]%
        {miller2019explanation}
\bibfield{author}{\bibinfo{person}{Tim Miller}.}
  \bibinfo{year}{2019}\natexlab{}.
\newblock \showarticletitle{Explanation in artificial intelligence: Insights
  from the social sciences}.
\newblock \bibinfo{journal}{\emph{Artificial intelligence}}
  \bibinfo{volume}{267} (\bibinfo{year}{2019}), \bibinfo{pages}{1--38}.
\newblock


\bibitem[Misheva et~al\mbox{.}(2021)]%
        {misheva2021explainable}
\bibfield{author}{\bibinfo{person}{Branka~Hadji Misheva},
  \bibinfo{person}{Joerg Osterrieder}, \bibinfo{person}{Ali Hirsa},
  \bibinfo{person}{Onkar Kulkarni}, {and} \bibinfo{person}{Stephen~Fung Lin}.}
  \bibinfo{year}{2021}\natexlab{}.
\newblock \showarticletitle{Explainable AI in credit risk management}.
\newblock \bibinfo{journal}{\emph{arXiv preprint arXiv:2103.00949}}
  (\bibinfo{year}{2021}).
\newblock


\bibitem[Mohseni et~al\mbox{.}(2021a)]%
        {mohseni2021machine}
\bibfield{author}{\bibinfo{person}{Sina Mohseni}, \bibinfo{person}{Fan Yang},
  \bibinfo{person}{Shiva Pentyala}, \bibinfo{person}{Mengnan Du},
  \bibinfo{person}{Yi Liu}, \bibinfo{person}{Nic Lupfer}, \bibinfo{person}{Xia
  Hu}, \bibinfo{person}{Shuiwang Ji}, {and} \bibinfo{person}{Eric Ragan}.}
  \bibinfo{year}{2021}\natexlab{a}.
\newblock \showarticletitle{Machine learning explanations to prevent overtrust
  in fake news detection}. In \bibinfo{booktitle}{\emph{Proceedings of the
  International AAAI Conference on Web and Social Media}},
  Vol.~\bibinfo{volume}{15}. \bibinfo{pages}{421--431}.
\newblock


\bibitem[Mohseni et~al\mbox{.}(2021b)]%
        {mohseni2021multidisciplinary}
\bibfield{author}{\bibinfo{person}{Sina Mohseni}, \bibinfo{person}{Niloofar
  Zarei}, {and} \bibinfo{person}{Eric~D Ragan}.}
  \bibinfo{year}{2021}\natexlab{b}.
\newblock \showarticletitle{A multidisciplinary survey and framework for design
  and evaluation of explainable AI systems}.
\newblock \bibinfo{journal}{\emph{ACM Transactions on Interactive Intelligent
  Systems (TiiS)}} \bibinfo{volume}{11}, \bibinfo{number}{3-4}
  (\bibinfo{year}{2021}), \bibinfo{pages}{1--45}.
\newblock


\bibitem[Molnar(2020)]%
        {molnar2020interpretable}
\bibfield{author}{\bibinfo{person}{Christoph Molnar}.}
  \bibinfo{year}{2020}\natexlab{}.
\newblock \bibinfo{booktitle}{\emph{Interpretable machine learning}}.
\newblock \bibinfo{publisher}{Lulu. com}.
\newblock


\bibitem[Mroczkowska(2020)]%
        {Fintech}
\bibfield{author}{\bibinfo{person}{Agnieszka Mroczkowska}.}
  \bibinfo{year}{2020}\natexlab{}.
\newblock \bibinfo{booktitle}{\emph{What is a Fintech Application?, Definition
  and Insights for Business Owners}}.
\newblock
\urldef\tempurl%
\url{https://www.thedroidsonroids.com/blog/what-is-a-fintech-application-definition-and-insights-for-business-owners/}
\showURL{%
Retrieved February 7, 2023 from \tempurl}


\bibitem[Mueller et~al\mbox{.}(2019)]%
        {mueller2019explanation}
\bibfield{author}{\bibinfo{person}{Shane~T Mueller}, \bibinfo{person}{Robert~R
  Hoffman}, \bibinfo{person}{William Clancey}, \bibinfo{person}{Abigail Emrey},
  {and} \bibinfo{person}{Gary Klein}.} \bibinfo{year}{2019}\natexlab{}.
\newblock \showarticletitle{Explanation in human-AI systems: A literature
  meta-review, synopsis of key ideas and publications, and bibliography for
  explainable AI}.
\newblock \bibinfo{journal}{\emph{arXiv preprint arXiv:1902.01876}}
  (\bibinfo{year}{2019}).
\newblock


\bibitem[M{\"u}ller et~al\mbox{.}(2022)]%
        {muller2022reshape}
\bibfield{author}{\bibinfo{person}{Ricardo M{\"u}ller}, \bibinfo{person}{Marco
  Schreyer}, \bibinfo{person}{Timur Sattarov}, {and} \bibinfo{person}{Damian
  Borth}.} \bibinfo{year}{2022}\natexlab{}.
\newblock \showarticletitle{RESHAPE: Explaining Accounting Anomalies in
  Financial Statement Audits by enhancing SHapley Additive exPlanations}. In
  \bibinfo{booktitle}{\emph{Proceedings of the Third ACM International
  Conference on AI in Finance}}. \bibinfo{pages}{174--182}.
\newblock


\bibitem[Nazemi et~al\mbox{.}(2022)]%
        {nazemi2022interpretable}
\bibfield{author}{\bibinfo{person}{Abdolreza Nazemi}, \bibinfo{person}{Jonas
  Rauch}, {and} \bibinfo{person}{Frank~J Fabozzi}.}
  \bibinfo{year}{2022}\natexlab{}.
\newblock \showarticletitle{Interpretable Machine Learning for Creditor
  Recovery Rates}.
\newblock \bibinfo{journal}{\emph{Available at SSRN 4190345}}
  (\bibinfo{year}{2022}).
\newblock


\bibitem[Neyshabur et~al\mbox{.}(2020)]%
        {neyshabur2020being}
\bibfield{author}{\bibinfo{person}{Behnam Neyshabur}, \bibinfo{person}{Hanie
  Sedghi}, {and} \bibinfo{person}{Chiyuan Zhang}.}
  \bibinfo{year}{2020}\natexlab{}.
\newblock \showarticletitle{What is being transferred in transfer learning?}
\newblock \bibinfo{journal}{\emph{Advances in neural information processing
  systems}}  \bibinfo{volume}{33} (\bibinfo{year}{2020}),
  \bibinfo{pages}{512--523}.
\newblock


\bibitem[Nori et~al\mbox{.}(2019)]%
        {nori2019interpretml}
\bibfield{author}{\bibinfo{person}{Harsha Nori}, \bibinfo{person}{Samuel
  Jenkins}, \bibinfo{person}{Paul Koch}, {and} \bibinfo{person}{Rich Caruana}.}
  \bibinfo{year}{2019}\natexlab{}.
\newblock \showarticletitle{Interpretml: A unified framework for machine
  learning interpretability}.
\newblock \bibinfo{journal}{\emph{arXiv preprint arXiv:1909.09223}}
  (\bibinfo{year}{2019}).
\newblock


\bibitem[of~Singapore(2021)]%
        {mfa}
\bibfield{author}{\bibinfo{person}{Monetary~Authority of Singapore}.}
  \bibinfo{year}{2021}\natexlab{}.
\newblock \bibinfo{booktitle}{\emph{Veritas Initiative Addresses Implementation
  Challenges in the Responsible Use of Artificial Intelligence and Data
  Analytics}}.
\newblock
\urldef\tempurl%
\url{https://www.mas.gov.sg/news/media-releases/2021/veritas-initiative-addresses-implementation-challenges}
\showURL{%
Retrieved February 8, 2023 from \tempurl}


\bibitem[Ong et~al\mbox{.}(2023)]%
        {ong2023finxabsaexplainable}
\bibfield{author}{\bibinfo{person}{Keane Ong}, \bibinfo{person}{Wihan van~der
  Heever}, \bibinfo{person}{Ranjan Satapathy}, \bibinfo{person}{Gianmarco
  Mengaldo}, {and} \bibinfo{person}{Erik Cambria}.}
  \bibinfo{year}{2023}\natexlab{}.
\newblock \bibinfo{title}{FinXABSA: Explainable Finance through Aspect-Based
  Sentiment Analysis}.
\newblock
\newblock
\showeprint[arxiv]{2303.02563}~[cs.CL]


\bibitem[Park and Yang(2022)]%
        {park2022interpretable}
\bibfield{author}{\bibinfo{person}{Sangjin Park} {and} \bibinfo{person}{Jae-Suk
  Yang}.} \bibinfo{year}{2022}\natexlab{}.
\newblock \showarticletitle{Interpretable deep learning LSTM model for
  intelligent economic decision-making}.
\newblock \bibinfo{journal}{\emph{Knowledge-Based Systems}}
  \bibinfo{volume}{248} (\bibinfo{year}{2022}), \bibinfo{pages}{108907}.
\newblock


\bibitem[Radford et~al\mbox{.}(2019)]%
        {radford2019language}
\bibfield{author}{\bibinfo{person}{Alec Radford}, \bibinfo{person}{Jeffrey Wu},
  \bibinfo{person}{Rewon Child}, \bibinfo{person}{David Luan},
  \bibinfo{person}{Dario Amodei}, \bibinfo{person}{Ilya Sutskever},
  {et~al\mbox{.}}} \bibinfo{year}{2019}\natexlab{}.
\newblock \showarticletitle{Language models are unsupervised multitask
  learners}.
\newblock \bibinfo{journal}{\emph{OpenAI blog}} \bibinfo{volume}{1},
  \bibinfo{number}{8} (\bibinfo{year}{2019}), \bibinfo{pages}{9}.
\newblock


\bibitem[Rallis et~al\mbox{.}(2022)]%
        {rallis2022interpretation}
\bibfield{author}{\bibinfo{person}{Ioannis Rallis}, \bibinfo{person}{Yannis
  Markoulidakis}, \bibinfo{person}{Ioannis Georgoulas}, \bibinfo{person}{George
  Kopsiaftis}, \bibinfo{person}{Maria Kaselimi}, \bibinfo{person}{Nikolaos
  Doulamis}, {and} \bibinfo{person}{Anastasios Doulamis}.}
  \bibinfo{year}{2022}\natexlab{}.
\newblock \showarticletitle{Interpretation of net promoter score attributes
  using explainable AI}. In \bibinfo{booktitle}{\emph{Proceedings of the 15th
  International Conference on PErvasive Technologies Related to Assistive
  Environments}}. \bibinfo{pages}{113--117}.
\newblock


\bibitem[Ribeiro et~al\mbox{.}(2016)]%
        {ribeiro2016should}
\bibfield{author}{\bibinfo{person}{Marco~Tulio Ribeiro},
  \bibinfo{person}{Sameer Singh}, {and} \bibinfo{person}{Carlos Guestrin}.}
  \bibinfo{year}{2016}\natexlab{}.
\newblock \showarticletitle{``Why should i trust you?'' Explaining the
  predictions of any classifier}. In \bibinfo{booktitle}{\emph{Proceedings of
  the 22nd ACM SIGKDD international conference on knowledge discovery and data
  mining}}. \bibinfo{pages}{1135--1144}.
\newblock


\bibitem[Rizinski et~al\mbox{.}(2022)]%
        {rizinski2022ethically}
\bibfield{author}{\bibinfo{person}{Maryan Rizinski}, \bibinfo{person}{Hristijan
  Peshov}, \bibinfo{person}{Kostadin Mishev}, \bibinfo{person}{Lubomir~T
  Chitkushev}, \bibinfo{person}{Irena Vodenska}, {and} \bibinfo{person}{Dimitar
  Trajanov}.} \bibinfo{year}{2022}\natexlab{}.
\newblock \showarticletitle{Ethically Responsible Machine Learning in Fintech}.
\newblock \bibinfo{journal}{\emph{IEEE Access}}  \bibinfo{volume}{10}
  (\bibinfo{year}{2022}), \bibinfo{pages}{97531--97554}.
\newblock


\bibitem[Rojat et~al\mbox{.}(2021)]%
        {rojat2021explainable}
\bibfield{author}{\bibinfo{person}{Thomas Rojat}, \bibinfo{person}{Rapha{\"e}l
  Puget}, \bibinfo{person}{David Filliat}, \bibinfo{person}{Javier Del~Ser},
  \bibinfo{person}{Rodolphe Gelin}, {and} \bibinfo{person}{Natalia
  D{\'\i}az-Rodr{\'\i}guez}.} \bibinfo{year}{2021}\natexlab{}.
\newblock \showarticletitle{Explainable artificial intelligence (xai) on
  timeseries data: A survey}.
\newblock \bibinfo{journal}{\emph{arXiv preprint arXiv:2104.00950}}
  (\bibinfo{year}{2021}).
\newblock


\bibitem[Rudin(2019)]%
        {rudin2019stop}
\bibfield{author}{\bibinfo{person}{Cynthia Rudin}.}
  \bibinfo{year}{2019}\natexlab{}.
\newblock \showarticletitle{Stop explaining black box machine learning models
  for high stakes decisions and use interpretable models instead}.
\newblock \bibinfo{journal}{\emph{Nature machine intelligence}}
  \bibinfo{volume}{1}, \bibinfo{number}{5} (\bibinfo{year}{2019}),
  \bibinfo{pages}{206--215}.
\newblock


\bibitem[Rudin and Radin(2019)]%
        {rudin2019we}
\bibfield{author}{\bibinfo{person}{Cynthia Rudin} {and} \bibinfo{person}{Joanna
  Radin}.} \bibinfo{year}{2019}\natexlab{}.
\newblock \showarticletitle{Why are we using black box models in AI when we
  don't need to? A lesson from an explainable AI competition}.
\newblock \bibinfo{journal}{\emph{Harvard Data Science Review}}
  \bibinfo{volume}{1}, \bibinfo{number}{2} (\bibinfo{year}{2019}),
  \bibinfo{pages}{10--1162}.
\newblock


\bibitem[Sahakyan et~al\mbox{.}(2021)]%
        {sahakyan2021explainable}
\bibfield{author}{\bibinfo{person}{Maria Sahakyan}, \bibinfo{person}{Zeyar
  Aung}, {and} \bibinfo{person}{Talal Rahwan}.}
  \bibinfo{year}{2021}\natexlab{}.
\newblock \showarticletitle{Explainable artificial intelligence for tabular
  data: A survey}.
\newblock \bibinfo{journal}{\emph{IEEE Access}}  \bibinfo{volume}{9}
  (\bibinfo{year}{2021}), \bibinfo{pages}{135392--135422}.
\newblock


\bibitem[Selvaraju et~al\mbox{.}(2017)]%
        {selvaraju2017grad}
\bibfield{author}{\bibinfo{person}{Ramprasaath~R Selvaraju},
  \bibinfo{person}{Michael Cogswell}, \bibinfo{person}{Abhishek Das},
  \bibinfo{person}{Ramakrishna Vedantam}, \bibinfo{person}{Devi Parikh}, {and}
  \bibinfo{person}{Dhruv Batra}.} \bibinfo{year}{2017}\natexlab{}.
\newblock \showarticletitle{Grad-cam: Visual explanations from deep networks
  via gradient-based localization}. In \bibinfo{booktitle}{\emph{Proceedings of
  the IEEE international conference on computer vision}}.
  \bibinfo{pages}{618--626}.
\newblock


\bibitem[Serengil et~al\mbox{.}(2022)]%
        {serengil2022comparative}
\bibfield{author}{\bibinfo{person}{Sefik~Ilkin Serengil},
  \bibinfo{person}{Salih Imece}, \bibinfo{person}{Ugur~Gurkan Tosun},
  \bibinfo{person}{Ege~Berk Buyukbas}, {and} \bibinfo{person}{Bilge Koroglu}.}
  \bibinfo{year}{2022}\natexlab{}.
\newblock \showarticletitle{A Comparative Study of Machine Learning Approaches
  for Non Performing Loan Prediction with Explainability}.
\newblock \bibinfo{journal}{\emph{International Journal of Machine Learning and
  Computing}} \bibinfo{volume}{12}, \bibinfo{number}{5} (\bibinfo{year}{2022}).
\newblock


\bibitem[Serrano and Smith(2019)]%
        {serrano2019attention}
\bibfield{author}{\bibinfo{person}{Sofia Serrano} {and} \bibinfo{person}{Noah~A
  Smith}.} \bibinfo{year}{2019}\natexlab{}.
\newblock \showarticletitle{Is attention interpretable?}
\newblock \bibinfo{journal}{\emph{arXiv preprint arXiv:1906.03731}}
  (\bibinfo{year}{2019}).
\newblock


\bibitem[Shapley et~al\mbox{.}(1953)]%
        {shapley1953value}
\bibfield{author}{\bibinfo{person}{Lloyd~S Shapley} {et~al\mbox{.}}}
  \bibinfo{year}{1953}\natexlab{}.
\newblock \showarticletitle{A value for n-person games}.
\newblock  (\bibinfo{year}{1953}).
\newblock


\bibitem[Shi et~al\mbox{.}(2021)]%
        {shi2021xpm}
\bibfield{author}{\bibinfo{person}{Si Shi}, \bibinfo{person}{Jianjun Li},
  \bibinfo{person}{Guohui Li}, \bibinfo{person}{Peng Pan}, {and}
  \bibinfo{person}{Ke Liu}.} \bibinfo{year}{2021}\natexlab{}.
\newblock \showarticletitle{XPM: An explainable deep reinforcement learning
  framework for portfolio management}. In \bibinfo{booktitle}{\emph{Proceedings
  of the 30th ACM international conference on information \& knowledge
  management}}. \bibinfo{pages}{1661--1670}.
\newblock


\bibitem[Sokol and Flach(2019)]%
        {sokol2019counterfactual}
\bibfield{author}{\bibinfo{person}{Kacper Sokol} {and} \bibinfo{person}{Peter~A
  Flach}.} \bibinfo{year}{2019}\natexlab{}.
\newblock \showarticletitle{Counterfactual Explanations of Machine Learning
  Predictions: Opportunities and Challenges for AI Safety.}
\newblock \bibinfo{journal}{\emph{SafeAI@ AAAI}} (\bibinfo{year}{2019}).
\newblock


\bibitem[Srinivasan et~al\mbox{.}(2019)]%
        {srinivasan2019generating}
\bibfield{author}{\bibinfo{person}{Ramya Srinivasan}, \bibinfo{person}{Ajay
  Chander}, {and} \bibinfo{person}{Pouya Pezeshkpour}.}
  \bibinfo{year}{2019}\natexlab{}.
\newblock \showarticletitle{Generating user-friendly explanations for loan
  denials using GANs}.
\newblock \bibinfo{journal}{\emph{arXiv preprint arXiv:1906.10244}}
  (\bibinfo{year}{2019}).
\newblock


\bibitem[Sudjianto and Zhang(2021)]%
        {sudjianto2021designing}
\bibfield{author}{\bibinfo{person}{Agus Sudjianto} {and} \bibinfo{person}{Aijun
  Zhang}.} \bibinfo{year}{2021}\natexlab{}.
\newblock \showarticletitle{Designing Inherently Interpretable Machine Learning
  Models}.
\newblock \bibinfo{journal}{\emph{arXiv preprint arXiv:2111.01743}}
  (\bibinfo{year}{2021}).
\newblock


\bibitem[Team(2022)]%
        {FinOverview}
\bibfield{author}{\bibinfo{person}{CFI Team}.} \bibinfo{year}{2022}\natexlab{}.
\newblock \bibinfo{booktitle}{\emph{Finance Overview: Personal,business and
  government}}.
\newblock
\urldef\tempurl%
\url{https://corporatefinanceinstitute.com/resources/wealth-management/finance-industry-overview/}
\showURL{%
Retrieved February 7, 2023 from \tempurl}


\bibitem[Tomsett et~al\mbox{.}(2018)]%
        {tomsett2018interpretable}
\bibfield{author}{\bibinfo{person}{Richard Tomsett}, \bibinfo{person}{Dave
  Braines}, \bibinfo{person}{Dan Harborne}, \bibinfo{person}{Alun Preece},
  {and} \bibinfo{person}{Supriyo Chakraborty}.}
  \bibinfo{year}{2018}\natexlab{}.
\newblock \showarticletitle{Interpretable to whom? A role-based model for
  analyzing interpretable machine learning systems}.
\newblock \bibinfo{journal}{\emph{arXiv preprint arXiv:1806.07552}}
  (\bibinfo{year}{2018}).
\newblock


\bibitem[Tran et~al\mbox{.}(2022)]%
        {tran2022explainable}
\bibfield{author}{\bibinfo{person}{Kim~Long Tran}, \bibinfo{person}{Hoang~Anh
  Le}, \bibinfo{person}{Thanh~Hien Nguyen}, {and} \bibinfo{person}{Duc~Trung
  Nguyen}.} \bibinfo{year}{2022}\natexlab{}.
\newblock \showarticletitle{Explainable Machine Learning for Financial Distress
  Prediction: Evidence from Vietnam}.
\newblock \bibinfo{journal}{\emph{Data}} \bibinfo{volume}{7},
  \bibinfo{number}{11} (\bibinfo{year}{2022}), \bibinfo{pages}{160}.
\newblock


\bibitem[Turbé et~al\mbox{.}(2023)]%
        {turbé_bjelogrlic_lovis_mengaldo_2023}
\bibfield{author}{\bibinfo{person}{Hugues Turbé}, \bibinfo{person}{Mina
  Bjelogrlic}, \bibinfo{person}{Christian Lovis}, {and}
  \bibinfo{person}{Gianmarco Mengaldo}.} \bibinfo{year}{2023}\natexlab{}.
\newblock \showarticletitle{Evaluation of post-hoc interpretability methods in
  time-series classification}.
\newblock \bibinfo{journal}{\emph{Nature Machine Intelligence}}
  (\bibinfo{year}{2023}).
\newblock
\urldef\tempurl%
\url{https://doi.org/10.1038/s42256-023-00620-w}
\showDOI{\tempurl}


\bibitem[van~den Berg and Kuiper(2020)]%
        {van2020xai}
\bibfield{author}{\bibinfo{person}{Martin van~den Berg} {and}
  \bibinfo{person}{Ouren Kuiper}.} \bibinfo{year}{2020}\natexlab{}.
\newblock \showarticletitle{XAI in the financial sector: a conceptual framework
  for explainable AI (XAI)}.
\newblock \bibinfo{journal}{\emph{https://www. hu.
  nl/-/media/hu/documenten/onderzoek/projecten/}} (\bibinfo{year}{2020}).
\newblock


\bibitem[Vaswani et~al\mbox{.}(2017)]%
        {vaswani2017attention}
\bibfield{author}{\bibinfo{person}{Ashish Vaswani}, \bibinfo{person}{Noam
  Shazeer}, \bibinfo{person}{Niki Parmar}, \bibinfo{person}{Jakob Uszkoreit},
  \bibinfo{person}{Llion Jones}, \bibinfo{person}{Aidan~N Gomez},
  \bibinfo{person}{{\L}ukasz Kaiser}, {and} \bibinfo{person}{Illia
  Polosukhin}.} \bibinfo{year}{2017}\natexlab{}.
\newblock \showarticletitle{Attention is all you need}.
\newblock \bibinfo{journal}{\emph{Advances in neural information processing
  systems}}  \bibinfo{volume}{30} (\bibinfo{year}{2017}).
\newblock


\bibitem[VINCENT(2018)]%
        {googlecase}
\bibfield{author}{\bibinfo{person}{JAMES VINCENT}.}
  \bibinfo{year}{2018}\natexlab{}.
\newblock \bibinfo{booktitle}{\emph{Google `fixed' its racist algorithm by
  removing gorillas from its image-labeling tech}}.
\newblock
\urldef\tempurl%
\url{https://www.theverge.com/2018/1/12/16882408/google-racist-gorillas-photo-recognition-algorithm-ai}
\showURL{%
Retrieved February 8, 2023 from \tempurl}


\bibitem[Vivek et~al\mbox{.}(2022)]%
        {vivek2022explainable}
\bibfield{author}{\bibinfo{person}{Yelleti Vivek}, \bibinfo{person}{Vadlamani
  Ravi}, \bibinfo{person}{Abhay~Anand Mane}, {and}
  \bibinfo{person}{Laveti~Ramesh Naidu}.} \bibinfo{year}{2022}\natexlab{}.
\newblock \showarticletitle{Explainable Artificial Intelligence and Causal
  Inference based ATM Fraud Detection}.
\newblock \bibinfo{journal}{\emph{arXiv preprint arXiv:2211.10595}}
  (\bibinfo{year}{2022}).
\newblock


\bibitem[Wand et~al\mbox{.}(2022)]%
        {wand2022identifying}
\bibfield{author}{\bibinfo{person}{Tobias Wand}, \bibinfo{person}{Martin
  He{\ss}ler}, {and} \bibinfo{person}{Oliver Kamps}.}
  \bibinfo{year}{2022}\natexlab{}.
\newblock \showarticletitle{Identifying Dominant Industrial Sectors in Market
  States of the S\&P 500 Financial Data}.
\newblock \bibinfo{journal}{\emph{arXiv preprint arXiv:2208.14106}}
  (\bibinfo{year}{2022}).
\newblock


\bibitem[Weitz et~al\mbox{.}(2019)]%
        {weitz2019you}
\bibfield{author}{\bibinfo{person}{Katharina Weitz}, \bibinfo{person}{Dominik
  Schiller}, \bibinfo{person}{Ruben Schlagowski}, \bibinfo{person}{Tobias
  Huber}, {and} \bibinfo{person}{Elisabeth Andr{\'e}}.}
  \bibinfo{year}{2019}\natexlab{}.
\newblock \showarticletitle{" Do you trust me?" Increasing user-trust by
  integrating virtual agents in explainable AI interaction design}. In
  \bibinfo{booktitle}{\emph{Proceedings of the 19th ACM international
  conference on intelligent virtual agents}}. \bibinfo{pages}{7--9}.
\newblock


\bibitem[Weng et~al\mbox{.}(2022)]%
        {weng2022analysis}
\bibfield{author}{\bibinfo{person}{Futian Weng}, \bibinfo{person}{Jianping
  Zhu}, \bibinfo{person}{Cai Yang}, \bibinfo{person}{Wang Gao}, {and}
  \bibinfo{person}{Hongwei Zhang}.} \bibinfo{year}{2022}\natexlab{}.
\newblock \showarticletitle{Analysis of financial pressure impacts on the
  health care industry with an explainable machine learning method: China
  versus the USA}.
\newblock \bibinfo{journal}{\emph{Expert Systems with Applications}}
  \bibinfo{volume}{210} (\bibinfo{year}{2022}), \bibinfo{pages}{118482}.
\newblock


\bibitem[Wexler et~al\mbox{.}(2019)]%
        {wexler2019if}
\bibfield{author}{\bibinfo{person}{James Wexler}, \bibinfo{person}{Mahima
  Pushkarna}, \bibinfo{person}{Tolga Bolukbasi}, \bibinfo{person}{Martin
  Wattenberg}, \bibinfo{person}{Fernanda Vi{\'e}gas}, {and}
  \bibinfo{person}{Jimbo Wilson}.} \bibinfo{year}{2019}\natexlab{}.
\newblock \showarticletitle{The what-if tool: Interactive probing of machine
  learning models}.
\newblock \bibinfo{journal}{\emph{IEEE transactions on visualization and
  computer graphics}} \bibinfo{volume}{26}, \bibinfo{number}{1}
  (\bibinfo{year}{2019}), \bibinfo{pages}{56--65}.
\newblock


\bibitem[Xie et~al\mbox{.}(2022)]%
        {xie2022active}
\bibfield{author}{\bibinfo{person}{Binhui Xie}, \bibinfo{person}{Longhui Yuan},
  \bibinfo{person}{Shuang Li}, \bibinfo{person}{Chi~Harold Liu},
  \bibinfo{person}{Xinjing Cheng}, {and} \bibinfo{person}{Guoren Wang}.}
  \bibinfo{year}{2022}\natexlab{}.
\newblock \showarticletitle{Active learning for domain adaptation: An
  energy-based approach}. In \bibinfo{booktitle}{\emph{Proceedings of the AAAI
  Conference on Artificial Intelligence}}, Vol.~\bibinfo{volume}{36}.
  \bibinfo{pages}{8708--8716}.
\newblock


\bibitem[Yan et~al\mbox{.}(2019)]%
        {yan2019new}
\bibfield{author}{\bibinfo{person}{Han Yan}, \bibinfo{person}{Sheng Lin},
  {et~al\mbox{.}}} \bibinfo{year}{2019}\natexlab{}.
\newblock \showarticletitle{New Trend in FinTech: Research on Artificial
  Intelligence Model Interpretability in Financial Fields}.
\newblock \bibinfo{journal}{\emph{Open Journal of Applied Sciences}}
  \bibinfo{volume}{9}, \bibinfo{number}{10} (\bibinfo{year}{2019}),
  \bibinfo{pages}{761}.
\newblock


\bibitem[Yang et~al\mbox{.}(2020)]%
        {yang2020generating}
\bibfield{author}{\bibinfo{person}{Linyi Yang}, \bibinfo{person}{Eoin Kenny},
  \bibinfo{person}{Tin Lok~James Ng}, \bibinfo{person}{Yi Yang},
  \bibinfo{person}{Barry Smyth}, {and} \bibinfo{person}{Ruihai Dong}.}
  \bibinfo{year}{2020}\natexlab{}.
\newblock \showarticletitle{Generating Plausible Counterfactual Explanations
  for Deep Transformers in Financial Text Classification}. In
  \bibinfo{booktitle}{\emph{Proceedings of the 28th International Conference on
  Computational Linguistics}}. \bibinfo{pages}{6150--6160}.
\newblock


\bibitem[Yang et~al\mbox{.}(2018)]%
        {yang2018explainable}
\bibfield{author}{\bibinfo{person}{Linyi Yang}, \bibinfo{person}{Zheng Zhang},
  \bibinfo{person}{Su Xiong}, \bibinfo{person}{Lirui Wei},
  \bibinfo{person}{James Ng}, \bibinfo{person}{Lina Xu}, {and}
  \bibinfo{person}{Ruihai Dong}.} \bibinfo{year}{2018}\natexlab{}.
\newblock \showarticletitle{Explainable text-driven neural network for stock
  prediction}. In \bibinfo{booktitle}{\emph{2018 5th IEEE International
  Conference on Cloud Computing and Intelligence Systems (CCIS)}}. IEEE,
  \bibinfo{pages}{441--445}.
\newblock


\bibitem[Yang et~al\mbox{.}(2019)]%
        {yang2019federated}
\bibfield{author}{\bibinfo{person}{Qiang Yang}, \bibinfo{person}{Yang Liu},
  \bibinfo{person}{Tianjian Chen}, {and} \bibinfo{person}{Yongxin Tong}.}
  \bibinfo{year}{2019}\natexlab{}.
\newblock \showarticletitle{Federated machine learning: Concept and
  applications}.
\newblock \bibinfo{journal}{\emph{ACM Transactions on Intelligent Systems and
  Technology (TIST)}} \bibinfo{volume}{10}, \bibinfo{number}{2}
  (\bibinfo{year}{2019}), \bibinfo{pages}{1--19}.
\newblock


\bibitem[Yasodhara et~al\mbox{.}(2021)]%
        {yasodhara2021trustworthiness}
\bibfield{author}{\bibinfo{person}{Angeline Yasodhara}, \bibinfo{person}{Azin
  Asgarian}, \bibinfo{person}{Diego Huang}, {and} \bibinfo{person}{Parinaz
  Sobhani}.} \bibinfo{year}{2021}\natexlab{}.
\newblock \showarticletitle{On the trustworthiness of tree ensemble
  explainability methods}. In \bibinfo{booktitle}{\emph{Machine Learning and
  Knowledge Extraction: 5th IFIP TC 5, TC 12, WG 8.4, WG 8.9, WG 12.9
  International Cross-Domain Conference, CD-MAKE 2021, Virtual Event, August
  17--20, 2021, Proceedings 5}}. Springer, \bibinfo{pages}{293--308}.
\newblock


\bibitem[Yeong Zee~Kin(2023)]%
        {davos}
\bibfield{author}{\bibinfo{person}{Tan Wen~Rui Yeong Zee~Kin, Lee Wan~Sie}.}
  \bibinfo{year}{2023}\natexlab{}.
\newblock \bibinfo{booktitle}{\emph{How Singapore is developing trustworthy
  AI}}.
\newblock
\urldef\tempurl%
\url{https://www.weforum.org/agenda/2023/01/how-singapore-is-demonstrating-trustworthy-ai-davos2023/}
\showURL{%
Retrieved February 8, 2023 from \tempurl}


\bibitem[Yuan and Zhang(2020)]%
        {yuan2020connecting}
\bibfield{author}{\bibinfo{person}{Jie Yuan} {and} \bibinfo{person}{Zhu
  Zhang}.} \bibinfo{year}{2020}\natexlab{}.
\newblock \showarticletitle{Connecting the dots: forecasting and explaining
  short-term market volatility}. In \bibinfo{booktitle}{\emph{Proceedings of
  the First ACM International Conference on AI in Finance}}.
  \bibinfo{pages}{1--8}.
\newblock


\bibitem[Zhang et~al\mbox{.}(2022c)]%
        {zhang2022explainable}
\bibfield{author}{\bibinfo{person}{Chanyuan~Abigail Zhang},
  \bibinfo{person}{Soohyun Cho}, {and} \bibinfo{person}{Miklos Vasarhelyi}.}
  \bibinfo{year}{2022}\natexlab{c}.
\newblock \showarticletitle{Explainable Artificial Intelligence (XAI) in
  auditing}.
\newblock \bibinfo{journal}{\emph{International Journal of Accounting
  Information Systems}}  \bibinfo{volume}{46} (\bibinfo{year}{2022}),
  \bibinfo{pages}{100572}.
\newblock


\bibitem[Zhang et~al\mbox{.}(2020b)]%
        {zhang2020explainable}
\bibfield{author}{\bibinfo{person}{Ruoyun Zhang}, \bibinfo{person}{Chao Yi},
  {and} \bibinfo{person}{Yixin Chen}.} \bibinfo{year}{2020}\natexlab{b}.
\newblock \showarticletitle{Explainable machine learning for regime-based asset
  allocation}. In \bibinfo{booktitle}{\emph{2020 IEEE International Conference
  on Big Data (Big Data)}}. IEEE, \bibinfo{pages}{5480--5485}.
\newblock


\bibitem[Zhang et~al\mbox{.}(2022a)]%
        {zhang2022interpretable}
\bibfield{author}{\bibinfo{person}{Wei Zhang}, \bibinfo{person}{Brian Barr},
  {and} \bibinfo{person}{John Paisley}.} \bibinfo{year}{2022}\natexlab{a}.
\newblock \showarticletitle{An Interpretable Deep Classifier for Counterfactual
  Generation}. In \bibinfo{booktitle}{\emph{Proceedings of the Third ACM
  International Conference on AI in Finance}}. \bibinfo{pages}{36--43}.
\newblock


\bibitem[Zhang et~al\mbox{.}(2022b)]%
        {zhang2022understanding}
\bibfield{author}{\bibinfo{person}{Wei Zhang}, \bibinfo{person}{Brian Barr},
  {and} \bibinfo{person}{John Paisley}.} \bibinfo{year}{2022}\natexlab{b}.
\newblock \showarticletitle{Understanding Counterfactual Generation using
  Maximum Mean Discrepancy}. In \bibinfo{booktitle}{\emph{Proceedings of the
  Third ACM International Conference on AI in Finance}}.
  \bibinfo{pages}{44--52}.
\newblock


\bibitem[Zhang et~al\mbox{.}(2020a)]%
        {newszhang2020explainable}
\bibfield{author}{\bibinfo{person}{Xiaohui Zhang}, \bibinfo{person}{Qianzhou
  Du}, {and} \bibinfo{person}{Zhongju Zhang}.}
  \bibinfo{year}{2020}\natexlab{a}.
\newblock \showarticletitle{An explainable machine learning framework for fake
  financial news detection}. In \bibinfo{booktitle}{\emph{2020 International
  Conference on Information Systems-Making Digital Inclusive: Blending the
  Local and the Global, ICIS 2020}}. Association for Information Systems.
\newblock


\bibitem[Zhang et~al\mbox{.}(2022d)]%
        {cazhang2022explainable}
\bibfield{author}{\bibinfo{person}{Zijiao Zhang}, \bibinfo{person}{Chong Wu},
  \bibinfo{person}{Shiyou Qu}, {and} \bibinfo{person}{Xiaofang Chen}.}
  \bibinfo{year}{2022}\natexlab{d}.
\newblock \showarticletitle{An explainable artificial intelligence approach for
  financial distress prediction}.
\newblock \bibinfo{journal}{\emph{Information Processing \& Management}}
  \bibinfo{volume}{59}, \bibinfo{number}{4} (\bibinfo{year}{2022}),
  \bibinfo{pages}{102988}.
\newblock


\end{thebibliography}

\appendix

\end{document}